\documentclass[twoside]{article}

\usepackage[accepted]{aistats2023}

\usepackage{booktabs}       
\usepackage{amsfonts}       
\usepackage{nicefrac}       
\usepackage{microtype}      
\usepackage{mathtools}
\usepackage{multirow}
\usepackage{bm}
\usepackage{epsfig}
\usepackage{graphicx}
\usepackage{caption}
\usepackage{subcaption}
\usepackage{amsthm}
\usepackage{amsmath}
\usepackage{amssymb}
\usepackage{enumitem}
\usepackage{makecell}
\usepackage{wrapfig}
\usepackage{indentfirst}
\usepackage{verbatim}
\usepackage{color}
\usepackage{setspace}
\usepackage{array}
\usepackage{booktabs}
\usepackage{stackengine}
\usepackage{algorithm}
\usepackage[algo2e,algoruled,boxed,lined]{algorithm2e}

\newtheorem{theorem}{Theorem}

\newcommand{\norm}[1]{\left\lVert#1\right\rVert}

\newcommand{\eg}{\emph{e.g.}}
\newcommand{\ie}{\emph{i.e.}}

\usepackage{tocloft}

\usepackage[toc,page,header]{appendix}

\usepackage{adjustbox}

\usepackage{minitoc}

\renewcommand \thepart{}
\renewcommand \partname{}

\usepackage{xcolor}
\usepackage[pagebackref=true,breaklinks=true,colorlinks,bookmarks=false]{hyperref}
\definecolor{deepred}{HTML}{940000}
\hypersetup{linkcolor=deepred}
\hypersetup{citecolor=[rgb]{0.4,0.15,0.95}}
\usepackage{url}            

\usepackage{colortbl}
\definecolor{Gray}{gray}{0.94}

\newcolumntype{a}{>{\columncolor{Gray}}c}

\newlength\savewidth\newcommand\shline{\noalign{\global\savewidth\arrayrulewidth
  \global\arrayrulewidth 1pt}\hline\noalign{\global\arrayrulewidth\savewidth}}

\DeclareMathOperator*{\argmin}{argmin}

\begin{document}

\twocolumn[

\aistatstitle{Iterative Teaching by Data Hallucination}

\vspace{-6.3mm}

\aistatsauthor{\small Zeju Qiu\textsuperscript{1,3,*},~~~~~Weiyang Liu\textsuperscript{1,2,*},~~~~~Tim Z. Xiao\textsuperscript{4},~~~~~Zhen Liu\textsuperscript{5},~~~~~Umang Bhatt\textsuperscript{2,6}\\
\bf \small Yucen Luo\textsuperscript{1},~~~~~Adrian Weller\textsuperscript{2,6},~~~~~Bernhard Schölkopf\textsuperscript{1}}

\vspace{0.2mm}
\aistatsaddress{\small\textsuperscript{1}Max Planck Institute for Intelligent Systems, Tübingen,~~~\textsuperscript{2}University of Cambridge,~~~\textsuperscript{3}Technical University of Munich\\
\small\textsuperscript{4}University of Tübingen,~~~\textsuperscript{5}Mila, Université de Montréal,~~~\textsuperscript{6}The Alan Turing Institute
} 
\vspace{-0.5mm}
]

\doparttoc 
\faketableofcontents

\begin{abstract}
\vspace{-2mm}
  We consider the problem of iterative machine teaching,  where a teacher sequentially provides examples based on the status of a learner under a discrete input space (\ie, a pool of finite samples), which greatly limits the teacher's capability. To address this issue, we study iterative teaching under a continuous input space where the input example (\ie, image) can be either generated by solving an optimization problem or drawn directly from a continuous distribution. Specifically, we propose data hallucination teaching~(DHT) where the teacher can generate input data intelligently based on labels, the learner's status and the target concept. We study a number of challenging teaching setups (\eg, linear/neural learners in omniscient and black-box settings). Extensive empirical results verify the effectiveness of DHT. The code is made publicly available on \href{https://github.com/Zeju1997/data_halucination_teaching}{Github}.
\end{abstract}

\vspace{-2.5mm}
\section{Introduction}
\vspace{-1mm}

Machine teaching~\cite{zhu2015machine,zhu2018overview} seeks a training dataset of minimal size such that a learner can learn a target concept based on this minimal dataset. Compared to machine learning where a learner is provided with a dataset to find the optimal parameters, machine teaching studies the inverse problem where the goal is to find a minimal dataset with which the learner can converge to the given target parameters. A deeper understanding towards machine teaching is essential in many applications, such as crowd sourcing~\cite{singla2014near,singla2013actively,zhou2018unlearn,zhou2020crowd}, optimal education~\cite{zhu2015machine}, model robustness~\cite{alfeld2016data,alfeld2017explicit,rakhsha2020policy,ma2019policy}, curriculum learning~\cite{bengio2009curriculum} and dataset distillation~\cite{wang2018dataset}.

\begin{figure}[t]
  \centering  
  \includegraphics[width=0.44\textwidth]{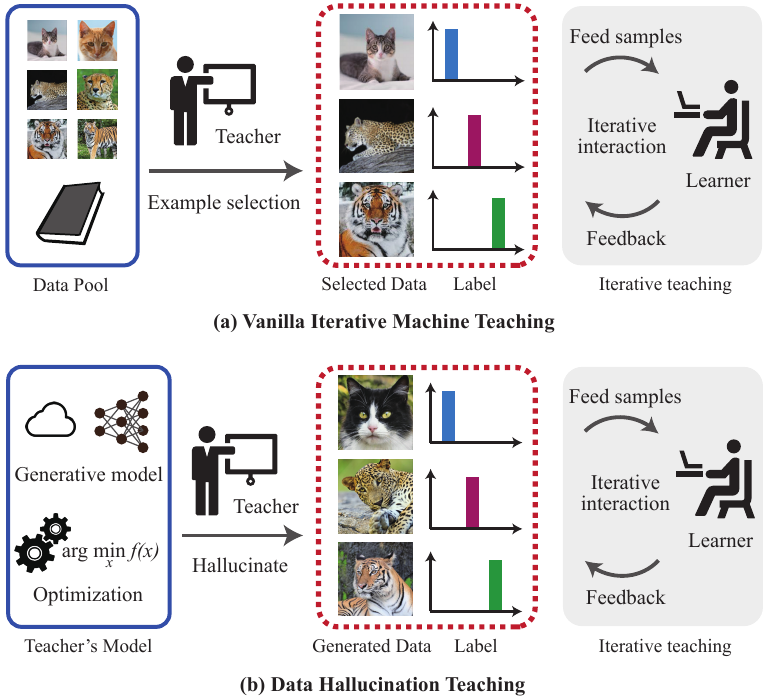}
  \vspace{-1mm}
  \caption{\footnotesize  Comparison of vanilla iterative machine
teaching and the proposed data hallucination teaching.}
  \vspace{-2.2mm}
  \label{fig:comparison}
\end{figure}

Depending on the type of learner, machine teaching can be carried out batch-wise (\ie, the teacher provides the dataset to the learner in one shot) or iteratively (\ie, the teacher provides data to the learner iteratively and adaptively). Motivated by the dominance of iterative learners (\eg, almost all types of neural networks), we study the problem of iterative machine teaching~(IMT)~\cite{liu2017iterative} where the teacher feeds data intelligently based on the learner's status in every iteration such that the learner can converge to the target concept within minimal iterations. The minimal number of such iterations is defined as \emph{iterative teaching dimension}. Vanilla IMT~\cite{liu2017iterative} iteratively selects examples from a fixed pool (\ie, dataset), which, however, is inherently a difficult combinatorial problem computationally prohibitive to solve. \cite{liu2021iterative} addresses this problem by finding a continuous teaching signal -- the label space. Despite its simplicity, label synthesis teaching still imposes a strong constraint on the teaching space, limiting its capability of faster convergence. To avoid the combinatorial problem of example selection while enjoying the flexibility of a continuous teaching space, we propose data hallucination teaching~(DHT), where the teacher generates from a continuous space the input data by conditioning on the learner's status. An intuitive comparison between IMT and DHT is given in Figure~\ref{fig:comparison}.

Another motivation behind DHT comes from the promising results of approximating a dataset with synthetic prototypes, such as dataset distillation~\cite{wang2018dataset,cazenavette2022dataset} and dataset condensation~\cite{zhao2021dataset}. DHT shares the same spirit as dataset approximation in the sense that both aim to guide the learner to some target concept with synthetic samples. Different from dataset approximation, DHT takes one step further by taking the specific iterative optimization algorithm into account and seeks to generate a sequence of examples (with ordering information) rather than a synthetic dataset.

\vspace{-0.14mm}

DHT can also be viewed as a natural generalization of IMT, extending the original discrete teaching space to a continuous one. Such a generalization introduces more modeling flexibility but meanwhile makes the teaching process more challenging. To tackle this challenge, we study both greedy teaching policy and parameterized teaching policy. Particularly for the parameterized one, we propose multiple teacher formulations (\eg, generative models) and multiple teacher's action spaces (\eg, Mixup sample space~\cite{zhang2017mixup}). We emphasize that DHT is quite different from standard generative models which usually capture static data distributions. In contrast, DHT models a \emph{dynamically changing} data distribution, which depends on the learner's status and is used for fast convergence rather than reconstruction.

\vspace{-0.135mm}

The intuition behind the benefits of the continuous teaching space in DHT comes from the empirical success of Mixup~\cite{zhang2017mixup,verma2019manifold} and data augmentation~\cite{krizhevsky2017imagenet,perez2017effectiveness,dao2019kernel,chen2020simple}. Mixup uses a linear interpolation between two samples for training neural networks. Data augmentation perturbs the inputs, \eg, images, in a small neighborhood around the original input. Both methods can be viewed as a continuous perturbation in the high-dimensional input space and special cases of DHT. Even if it is only a small subset of the continuous input space being considered, the empirical generalization performance can be significantly improved. Therefore, the original discrete input space can be sub-optimal for teaching, for which we propose to explore the continuous input space in order to improve the learner's convergence.

\vspace{-0.135mm}

Specifically, we study DHT under both the omniscient scenario, where the teacher knows everything (particularly the optimal learner parameters) about the learner, and the black-box scenario, where the teacher has no prior knowledge of the optimal learner parameters. We theoretically prove that DHT can achieve exponential teachability~(ET)~\cite{liu2017iterative}, and empirically show that DHT achieves much faster convergence than a random teacher (\ie, SGD) and IMT~\cite{liu2017iterative}.

\vspace{-0.135mm}

Most significantly, we formulate the problem of teaching black-box neural learners as \emph{performative teaching}, which is a novel application of performativity~\cite{perdomo2020performative} to iterative teaching. Specifically, performative teaching assumes that the teaching target will shift based on the teacher's action. This is exactly the scenario when we perform iterative teaching in the representation space of neural networks. We show that under this formulation, DHT is able to teach deep neural learners in a fully black-box manner on realistic datasets such as CIFAR-10 and CIFAR-100. We believe that it is the very first time that iterative teaching can be used to teach black-box nonlinear learners in realistic settings while achieving significant empirical performance gain. 

Our contribution can be briefly summarized as follows:

\vspace{-3.1mm}
\begin{itemize}[leftmargin=*]
\setlength\itemsep{0.04em}
    \item We propose a novel teaching framework -- \textbf{data hallucination teaching}, where the teacher iteratively generates synthetic training data depending on the learner's status. DHT yields a highly flexible teaching space.
    \item In the DHT framework, we comprehensively study the greedy and parameterized policies under both the omniscient and black-box scenarios.
    \item We propose a novel performative formulation for iterative teaching, which assumes a dynamically changing teaching target. The formulation is shown to be a natural fit for teaching black-box neural learners.
    \item For the first time, we are able to apply iterative teaching to black-box neural learners on realistic datasets. Significant performance gain is observed empirically.
    \item We demonstrate faster convergence of DHT versus SGD and other baselines, both theoretically and empirically.
\end{itemize}

\vspace{-3.5mm}
\section{Related Work}
\vspace{-1.5mm}

\textbf{Machine teaching}. The study of machine teaching begins with the batch setting~\cite{zhu2013machine,zhu2015machine,liu2016teaching,mansouri2019preference}, where the teacher simply prepares a dataset of minimal size to the learner towards some target concept. Efforts have been made on the teaching behavior of different types of learner, such as version space learners~\cite{chen2018understanding,tabibian2019enhancing}, linear learners~\cite{liu2016teaching}, kernel learners~\cite{kumar2020teaching}, reinforcement learner~\cite{zhang2020teaching}, active learners~\cite{wang2021teaching,peltola2019machine}, teacher-aware learners~\cite{yuan2021iterative} and forgetful learners~\cite{hunziker2019teaching,liu2018blackbox}. Iterative (or sequential) machine teaching~\cite{liu2017iterative,liu2018blackbox,lessard2019optimal,mansouri2019preference,Xu2021LST,liu2021iterative} studies iterative learners by considering the specific optimization algorithm that the learner uses. The teaching performance is measured by the learner's convergence. Machine teaching has diverse applications in reinforcement learning~\cite{tschiatschek2019learner,kamalaruban2019interactive,rakhsha2020policy,haug2018teaching}, human-in-the-loop learning~\cite{johns2015becoming,chen2018near,mac2018teaching}, crowd sourcing~\cite{singla2014near,zhou2018unlearn,zhou2020crowd} and cyber security~\cite{mei2015using,alfeld2016data,zhang2018training,zhang2020adaptive,zhang2020online}. Sharing similar spirits, cooperative communication~\cite{wang2020mathematical,shafto2021cooperative,wang2020sequential} also studies the interaction between a teacher and a learner as well as how information can be transmitted efficiently.

\vspace{-0.24mm}

\textbf{Data augmentation}. In deep learning, data augmentation is ubiquitous~\cite{krizhevsky2017imagenet,wei2019eda,park2019specaugment,chen2020simple} and plays a crucial role in regularizing neural networks and improving generalization. Without it, the training set can be easily fitted and training loss will be minimized to zero even with random labels~\cite{zhang2021understanding}. Data augmentation is the \emph{de facto} choice in image recognition~\cite{krizhevsky2017imagenet,simonyan2014very,he2015deep} and also one of the key ingredients to the success of contrastive learning~\cite{chen2020simple,he2020momentum}.

\vspace{-0.24mm}

\textbf{Dataset approximation}. How to approximate a dataset with a few representative prototypes that can be used for training remains an open problem and is actively studied in coreset~\cite{tsang2005core,campbell2018bayesian,sener2018active,mirzasoleiman2020coresets}, dataset pruning~\cite{angelova2005pruning,lapedriza2013all}, dataset distillation~\cite{wang2018dataset,cazenavette2022dataset} and dataset condensation~\cite{zhao2021dataset}. However, dataset approximation typically considers one batch of data, while DHT constructs a sequence of data samples.

\section{Data Hallucination Teaching}
\vspace{-1.75mm}
\subsection{Problem Settings}
\vspace{-1.5mm}

\textbf{Teaching protocol}. We generally follow the teaching protocol in \cite{liu2017iterative,liu2021iterative}. This section mostly considers the omniscient scenario. That is, both the teacher and the learner observe the same sample $\mathcal{A}$ and share the same feature space, which represents $\mathcal{A}$ as $\bm{x}$ with the label $\bm{y}$. The teacher knows all the information about the learner, including the model parameters $\bm{w}^i$ at the $i$-th iteration, the learning rate $\eta_i$, the loss function $\ell$ and the optimization algorithm (usually we consider SGD). The teacher can only feed examples $(\bm{x}^i,\bm{y}^i)$ to the learner at the $i$-th iteration.

\vspace{-0.25mm}

\textbf{Teacher's objective}. In the omniscient scenario, the teacher aims to provide examples to the learner in every iteration such that the learner parameters $\bm{w}$ converge to the desired parameters $\bm{w}^*$ as quickly as possible. We typically use $\bm{w}^*=\arg\min_{\bm{w}}\mathbb{E}_{(\bm{x},\bm{y})}\{\ell(\bm{x},\bm{y}|\bm{w})\}$. The teacher seeks to optimize the following objective for gradient decent learner:
\begin{equation}
    \footnotesize
    \begin{aligned}
    &\min_{\{(\bm{x}^1,\bm{y}^1),\cdots,(\bm{x}^T,\bm{y}^T)\}} T\\
    &\text{s.t.}~\left\{
{\begin{array}{*{20}{l}}
{d(\bm{w}^T,\bm{w}^*)\leq\epsilon}\\
{\bm{w}^{t+1}=\bm{w}^t-\eta_t\frac{\partial\ell(\bm{x}^{t+1},\bm{y}^{t+1}|\bm{w}^t)}{\partial \bm{w}^t} }
\end{array}} \right.
    \end{aligned}
\end{equation}
where $d(\cdot,\cdot)$ denotes some discrepancy measure (\eg, Euclidean distance or cosine similarity). The above optimization is in general intractable and $\epsilon$ is hard to set in practice, so we usually resort to a simpler teacher's objective:
\begin{equation}\label{eq:teacher_obj}
\footnotesize
    \min_{\{(\bm{x}^1,\bm{y}^1),\cdots,(\bm{x}^T,\bm{y}^T)\}} d( \bm{w}^T,\bm{w}^*)
\end{equation}
where $T$ is the prescribed termination iteration. This minimization aims to find a teaching trajectory of length $T$ such that the distance between $\bm{w}^T$ and $\bm{w}^*$ reaches the minimum.

\vspace{-0.25mm}

\textbf{Learner's objective}. The learner minimizes its loss function $\ell$ with examples given by the teacher. If the teacher feeds one example at a time, gradient descent learners use
\begin{equation}
\footnotesize
\bm{w}^{t+1}=\bm{w}^t-\eta_t\frac{\partial\ell(\bm{x}^{t+1},\bm{y}^{t+1}|\bm{w}^t)}{\partial \bm{w}^t}
\end{equation}
where $\ell$ can be any regression or classification loss.

\vspace{-1.75mm}
\subsection{Greedy Teaching Policy}
\vspace{-1.5mm}

We start with the simplest greedy teaching policy which uses Euclidean distance as $d$ and approximates Equation~\ref{eq:teacher_obj} with $T$-times one-step minimization. DHT aims to generate the teaching example $(\bm{x},\bm{y})$. To simplify the problem, we uniformly sample a label $\bm{y}$ and synthesize the corresponding data $\bm{x}$ for teaching. This leads to the one-step optimization:
\begin{equation}\label{eq:greedy_linear}
    \footnotesize
    \begin{aligned}
        \min_{\bm{x}^{t+1}\in\mathcal{X},\bm{y}^{t+1}\sim \mathbb{U}}  &\eta^2_{t} \norm{ \frac{\partial \ell(\bm{x}^{t+1},\bm{y}^{t+1}|\bm{w}^t) }{\partial \bm{w}^t} }^2_2  \\&- 2 \eta_{t} \langle \bm{w}^t - \bm{w}^*, \frac{\partial \ell(\bm{x}^{t+1},\bm{y}^{t+1}|\bm{w}^t) }{\partial \bm{w}^t} \rangle
    \end{aligned}
\end{equation}
where $\bm{x}^{t+1}$ is optimized within $\mathcal{X}$ (\eg, pixel space $[0,255]$) and $\bm{y}$ is sampled uniformly from the discrete label space.
\newpage

Equation~\ref{eq:greedy_linear} can be directly used to teach any linear learner such as least square regression and logistic regression. Despite its simplicity, the greedy policy is computationally expensive if $\bm{x}$ is high-dimensional (\eg, images).

\vspace{-2.2mm}
\subsection{Parameterized Teaching Policy}
\vspace{-1.5mm}

The greedy policy considers only a one-step update for the learner, hence is inevitably sub-optimal. However, considering all the possible combinations of $\bm{x}$ in multiple steps is computationally infeasible, especially when the teaching space $\mathcal{X}$ is continuous. To address this, here we study a parameterized teaching policy. 
The central idea is to parameterize the teacher by a neural network $\bm{\theta}$, and the teaching policy is denoted as $\pi_{\bm{\theta}}$. With a parameterized policy, we can easily consider multiple-step updates for the learner. For example, when taking $v$-step updates for the learner into account, the teacher optimizes $\min_{\pi_{\bm{\theta}}}\|\bm{w}^{t+v}-\bm{w}^*\|^2$.

\vspace{-2mm}
\subsubsection{Data Transformation}\label{sec:data_trans}
\vspace{-1.5mm}

To simplify the problem, we start with a data transformation policy (at the $t$-th iteration) $\tilde{\bm{x}}=\pi_{\bm{\theta}}(\bm{x},\bm{y},\bm{w}^t,\bm{w}^*)$ that transforms a randomly sampled data point $(\bm{x},\bm{y})$ to a teaching example $(\tilde{\bm{x}},\bm{y})$. To learn $\bm{\theta}$, we have that

\vspace{-8mm}
\begin{equation}
\footnotesize
\begin{aligned}
& \!\min_{\bm{\theta}}  \norm{ \bm{w}^{v}(\bm{\theta}) - \bm{w}^* }^2_2 + \alpha\sum_{i=1}^{v} \ell ({\pi}_{\bm{\theta}}(\bm{x}^i, \bm{y}^i, \bm{w}^i_{\text{SG}}, \bm{w^*}), \bm{y}^i | \bm{w}^i_{\text{SG}}) \\[-0.325mm]
& ~~~~\text{s.t.}~\bm{w}^{v} (\bm{\theta}) = \arg\min_{\bm{w}} \mathbb{E}_{ (\bm{x}, \bm{y} )} \big{\{} \ell \big({\pi}_{\bm{\theta}} (\bm{x}, \bm{y}, \bm{w}, \bm{w}^*), \bm{y} | \bm{w}\big)\big{\}}\nonumber
\end{aligned}
\end{equation}
\vspace{-5mm}

where $\alpha$ is a hyperparameter and the learner is initialized at $\bm{w}^0$. The policy $\pi_{\bm{\theta}}$ keeps transforming the randomly sampled data based on both the current learner parameters and the target parameters in order to improve the learner's convergence to $\bm{w}^*$. To solve this bi-level optimization, we can simply unroll the inner optimization with $v$ steps of stochastic gradient descent, which enables the gradient to flow back to $\bm{\theta}$ when solving the outer minimization~\cite{liu2021iterative,liu2021orthogonal}. This shares the same spirit as meta-learning~\cite{finn2017model} and back-propagation through time in recurrent networks~\cite{werbos1988generalization}. We note that the greedy policy is the special case of $v=1$. In order to amplify the learning signal, we introduce an auxiliary intermediate loss minimization into the teacher's objective. With this auxiliary term, the teacher will favor the teaching trajectory that not only quickly guides the learner to $\bm{w}^*$ but also well minimizes the learner's loss function. In the implementation, we simplify the gradient in the auxiliary term by replacing $\bm{w}^i$ with $\bm{w}^i_{\text{SG}}=\text{StopGradient}(\bm{w}^i)$.

\vspace{-2mm}
\subsubsection{Generative Modeling}
\vspace{-1.5mm}

The data transformation policy builds a learner-conditioned deterministic mapping from existing data to teaching examples, and it does not model the underlying distribution of teaching examples. Moreover, the data transformation policy is likely to generate unrealistic samples that do not match the underlying data distribution $p(\bm{x})$. To this end, we study the generative teaching policy. The central idea is to parameterize the teaching policy with a generative model and impose a distribution divergence constraint, \ie, $\text{Div}(p(\pi),p(\bm{x}))\leq \epsilon$, which is used to force teaching examples to be similar to the empirical data distribution.

\textbf{GAN-based teacher}. One of the simplest ways to perform generative modeling is to use generative adversarial networks~(GANs)~\cite{goodfellow2020generative}. Therefore, we parameterize the teacher as a generator and introduce an additional discriminator $D(\cdot)$ to close the gap between synthetic teaching examples and real data. Specifically, the teacher model optimizes
\begin{equation}
\footnotesize
\begin{aligned}
& \min_{\bm{\theta}} \max_{D} \mathbb{E}_{\tilde{\bm{x}}\sim p_{\pi}(\bm{z})}\log\big(1\!-\!D(\pi_{\bm{\theta}}(\bm{z}))\big)+\mathbb{E}_{\bm{x}\sim p(\bm{x}) } \log \big( D(\bm{x})\big)  \\[-0.35mm]
&+\!\norm{ \bm{w}^{v}(\bm{\theta}) \!-\! \bm{w}^* }^2_2 + \alpha\sum_{i=1}^{v} \ell \big({\pi}_{\bm{\theta}} (\bm{z},\bm{x}^i, \bm{y}^i, \bm{w}^i_{\text{SG}}, \bm{w}^*), \bm{y}^i | \bm{w}^i_{\text{SG}}\big) \\[-0.3mm]
& ~~~\text{s.t.}~\bm{w}^{v} (\bm{\theta}) = \arg\min_{\bm{w}} \mathbb{E}_{ (\bm{x}, \bm{y} )} \big{\{} \ell \big({\pi}_{\bm{\theta}} (\bm{z},\bm{x}, \bm{y}, \bm{w}, \bm{w}^*), \bm{y} | \bm{w}\big)\big{\}}\nonumber
\end{aligned}
\end{equation}
where $\bm{z}$ is a noise vector following a normal distribution and we sometimes omit the input arguments ($\bm{x}, \bm{y}, \bm{w}, \bm{w}^*$) for $\pi_{\bm{\theta}}$ for notational simplicity. Similar to learning the data transformation policy, we unroll $v$ steps of SGD for the inner optimization and put $\bm{w}^{v} (\bm{\theta})$ into the outer min-max optimization for end-to-end training. This outer min-max problem can be solved following standard GAN training.

\textbf{VAE-based teacher}. We take advantage of a pretrained variational autoencoder~(VAE)~\cite{kingma2013auto} to realize the distribution divergence constraint. Specifically, we first pretrain a VAE on the full dataset to capture the joint distribution $p(\bm{x},\bm{y})$, and then let the teaching policy to generate data in the latent space of the VAE. The objective for the teacher is

\vspace{-6.9mm}
\begin{equation}
\footnotesize
\begin{aligned}
& \min_{\bm{\theta}}  \norm{ \bm{w}^{v} \!-\! \bm{w}^* }^2_2 + \alpha\sum_{i=1}^{v} \ell \big(p_{\bm{\psi}}({\pi}_{\bm{\theta}}),\bm{y} | \bm{w}^i_{\text{SG}}\big) +\text{KL} \big({\pi}_{\bm{\theta}} || p(\bm{u})\big) \\[-0.125mm]
& ~~~~~~~~~~~~\text{s.t.}~\bm{w}^{v} (\bm{\theta}) = \arg\min_{\bm{w}} \mathbb{E}_{ (\bm{x}, \bm{y} )} \big{\{} \ell \big(p_{\bm{\psi}}({\pi}_{\bm{\theta}}), \bm{y} | \bm{w}\big)\big{\}}\nonumber
\end{aligned}
\end{equation}
\vspace{-4.3mm}

where the fixed Gaussian prior for VAE's latent $\bm{u}$ is $p(\bm{u})$ and the decoder is $p_{\bm{\psi}}({\pi}_{\bm{\theta}})=p_{\bm{\psi}}(\tilde{\bm{x}}|{\pi}_{\bm{\theta}} (\bm{u},\bm{x}, \bm{y}, \bm{w}, \bm{w}^*),\bm{y})$. $\text{KL}(\cdot||\cdot)$ is the Kullback–Leibler divergence. VAE essentially serves as a bridge between the latent space and the input data space, and the teacher generates the latent code which is then mapped to raw data by the decoder. Compared to GAN-based teacher, VAE-based teacher enjoys stronger training stability and also avoids the problem of mode collapse. Since the GAN-based teacher is jointly trained with the learner, it may produce examples that achieve better teaching performance but yield weaker semantic meaning.

\vspace{-1.7mm}
\subsection{Theoretical Insights and Discussions}
\vspace{-1.4mm}

Similar to sample selection~\cite{liu2017iterative,liu2018blackbox} and label synthesis~\cite{liu2021iterative}, we now show that the greedy DHT can provably achieve ET. We consider two types of linear learners here: $\ell_{\textnormal{LSR}}(\bm{x},y|\bm{w})=\frac{1}{2}(\langle· \bm{w},\bm{x}\rangle-y)^2$ for the least square regression~(LSR) learner and $ \ell_{\textnormal{LR}}(\bm{x},y|\bm{w})=\log(1+\exp\{-y\langle \bm{w},\bm{x}\rangle\})$ for the logistic regression~(LR) learner. For simplicity, we consider the case where the label is a scalar. For LSR, the gradient \emph{w.r.t.} $\bm{w}$ of a single sample $ (\bm{x},{y})$ is $ \nabla_{\bm{w}}\ell=(\langle\bm{w},\bm{x}\rangle-{y})\bm{x}$. For LR learners, the gradient is $\nabla_{\bm{w}}\ell=\frac{-{y}\bm{x}}{1+\exp({y}\langle\bm{w},\bm{x}\rangle)}$. Then we define $g(\tilde{\bm{x}})$ as the teaching gradient ratio that is used to quantify the scale difference between the gradient of a normal sample and that of a teaching example. $\mathcal{T}_{\bm{x}\rightarrow\tilde{\bm{x}}}$ is defined as the transformation operator that maps $\bm{x}$ to $\tilde{\bm{x}}$, \ie, $\tilde{\bm{x}}=\mathcal{T}_{\bm{x}\rightarrow\tilde{\bm{x}}}\circ\bm{x}$. We have
\begin{equation}
\footnotesize
    \nabla_{\bm{w}}\ell(\tilde{\bm{x}},y|\bm{w}) = g(\tilde{\bm{x}})\cdot\mathcal{T}_{\bm{x}\rightarrow\tilde{\bm{x}}}\circ\nabla_{\bm{w}}\ell(\bm{x},y|\bm{w}).
\end{equation}
For LSR, we have that $g(\tilde{\bm{x}})=\frac{\langle\bm{w},\bm{x}\rangle -y}{\langle\bm{w},\tilde{\bm{x}}\rangle -y}$. For LR, we have that $g(\tilde{\bm{x}})=\frac{1+\exp({y}\langle\bm{w},\tilde{\bm{x}}\rangle)}{1+\exp(y\langle\bm{w},\bm{x}\rangle)}$. We note that $g(\tilde{\bm{x}})$ is important for the convergence of the learner and also largely determines the teacher's ability to achieve ET. Following prior work~\cite{liu2017iterative,liu2018blackbox}, ET is defined as the ability for the teacher to guide the learner to converge to $\bm{w}^*$ at an exponential rate.

\vspace{1.25mm}
\begin{theorem}[Exponential teachability of DHT]\label{thm:ET_DHT}
    Assume that the learner loss $\ell_i$ has the property of interpolation, $L_i$-Lipschitz, and convexity. $f$ is order-1 $\mu$ strongly convex. Then DHT can achieve ET if $\mathcal{T}_{\bm{x}\rightarrow\tilde{\bm{x}}}$ is an scaling mapping, i.e., $\mathcal{T}_{\bm{x}\rightarrow\tilde{\bm{x}}}\circ\bm{x}=\beta \bm{x}$ and $\beta$ is adjusted such that $g(\tilde{\bm{x}})=c_1\|\bm{w}^t-\bm{w}^*\|$. Specifically, we have that
    \begin{equation}
    \footnotesize
       \! \mathbb{E}\{\|\bm{w}^T-\bm{w}^*\|^2\}\leq(1-c_1\eta_t\bar{\mu}+c_1^2\eta_t^2 L_{\textnormal{max}})^T\norm{\bm{w}^0-\bm{w}^*}^2
    \end{equation}
    in which $L_{\textnormal{max}}=\max_i L_i$ and $\bar{\mu}=\sum_i\mu_i/n$. It implies that $\mathcal{O}((\log \frac{1}{c_0})^{-1} \log(\frac{1}{\epsilon}))$ samples are needed to achieve $\mathbb{E}\{\|\bm{w}^T-\bm{w}^*\|^2\}\leq \epsilon$. We let $c_0=1-c_1\eta_t\bar{\mu}+c_1^2\eta_t^2 L_{\textnormal{max}}$ and $c_1$ is adjusted such that $0<c_1\eta_t<\bar{\mu}/L_{\textnormal{max}}$ holds. 
\end{theorem}

Theorem~\ref{thm:ET_DHT} 
validates the importance of $g(\tilde{\bm{x}})$ in achieving ET. When $\mathcal{T}_{\bm{x}\rightarrow\tilde{\bm{x}}}$ is a scaling mapping, DHT recovers the case of label synthesis~\cite{liu2021iterative}. Thus, DHT can always achieve a better convergence rate than label synthesis. Further, DHT enjoys all the theoretical guarantees for label synthesis. When $\mathcal{T}_{\bm{x}\rightarrow\tilde{\bm{x}}}$ is a nonlinear mapping, DHT will become very flexible and potentially a better convergence rate can be derived.

\vspace{-1.75mm}
\section{Black-box DHT for Neural Learners}
\vspace{-1.6mm}

In this section, we discuss how DHT can be used to teach neural learners in a black-box setting. Black-box teaching for neural teachers has long been an open challenge in iterative machine teaching. We start by studying how parameterized DHT can be extended to the black-box setting and then introduce a novel alternative -- performative teaching which can naturally be used to teach neural learners.

Black-box teaching is generally difficult due to two aspects. First, the optimal learner parameters $\bm{w}^*$ are no longer given and how to find a good surrogate to measure the distance to $\bm{w}^*$ is crucial (essentially when the learner is nonlinear and non-convex). In general, the goal of black-box teaching is to improve the learner's generalizability instead of its convergence to some $\bm{w}^*$. Therefore, we usually seek to find a surrogate for $\bm{w}^*=\arg\min_{\bm{w}}\mathbb{E}_{(\bm{x},\bm{y})\sim \mathbb{P}_{\text{real}}}\{\ell(\bm{x},\bm{y}|\bm{w})\}$ where $\mathbb{P}_{\text{real}}$ denotes the underlying joint data and label distribution. Second, the teaching space of DHT is huge and how to properly reduce the teaching space to a reasonably small yet sufficiently effective one is important. Black-box teaching shares a similar ultimate goal to knowledge distillation~\cite{hinton2015distilling}.

\subsection{Mixup-based Teaching}
\vspace{-1.5mm}

We propose a black-box DHT based on the data augmentation space in Mixup~\cite{zhang2017mixup}. The basic idea is to learn a teacher that produces the learner-conditioned Mixup coefficients.

\textbf{Surrogate target}. We use a simple surrogate to measure the distance to $\bm{w}^*$: the validation accuracy on a held-out validation data set that is not used for training the learner. Recent studies in neural architecture search~\cite{zoph2016neural,liu2018progressive,liu2018darts}, meta-learning~\cite{andrychowicz2016learning} and automated machine learning~\cite{cubuk2019autoaugment} validate the effectiveness of such a surrogate to approximate the distance to a generalizable $\bm{w}^*$.

\textbf{Teaching space}. We restrict the teacher's action space to the data augmentation space in Mixup. Specifically, the teaching policy outputs the interpolation between two randomly selected samples $(\bm{x}_1,\bm{y}_1)$ and $(\bm{x}_2,\bm{y}_2)$:
\begin{equation}
\footnotesize
    \pi_{\bm{\theta}}\big((\bm{x}_1,\bm{y}_1),(\bm{x}_2,\bm{y}_2),\bm{w}^t\big) = \lambda_{\bm{\theta}}\bm{x}_1 + (1-\lambda_{\bm{\theta}}) \bm{x}_2
\end{equation}
where $\lambda_{\bm{\theta}}=h_{\bm{\theta}}((\bm{x}_1,\bm{y}_1),(\bm{x}_2,\bm{y}_2),\bm{w}^t)$. $h_{\bm{\theta}}(\cdot)$ is a neural network parameterized by $\bm{\theta}$ and outputs the Mixup coefficient $\lambda$ for mixing $\bm{x}_1$ and $\bm{x}_2$. The teaching example is $\tilde{\bm{x}}=\lambda_{\bm{\theta}}\bm{x}_1+(1-\lambda_{\bm{\theta}})\bm{x}_2$, and its label is $\tilde{\bm{y}}=\lambda_{\bm{\theta}}\bm{y}_1+(1-\lambda_{\bm{\theta}})\bm{y}_2$. Specifically, we can learn a teacher network $h(\cdot)$ that either outputs a continuous value within $[0,1]$ or outputs a discrete value (\eg, $\{0,0.25,0.5,0.75,1\}$). For the latter case, the teacher is a classifier for the discrete Mixup coefficients.

\textbf{Unrolling}. We first formulate the learning of the teacher as a bi-level optimization similar to the one in Section~\ref{sec:data_trans}. The outer optimization is to minimize the empirical risk on the validation set $\mathcal{D}_a$ and the inner optimization is to minimize the empirical risk on the training set $\mathcal{D}_r$. We have
\begin{equation}\label{eq:pertteac_alg}
    \footnotesize
    \begin{aligned}
    &~~~~~~~~~~~~~~~~~\min_{\bm{\theta}}\mathbb{E}_{(\bm{x}_a,\bm{y}_a)\sim\mathcal{D}_a}\big{\{}\ell\big(\bm{x}_a, 
 \bm{y}_a|\bm{w}^v(\bm{\theta})\big)\big{\}}\\
        &\textnormal{s.t.}~\bm{w}^v(\bm{\theta}) =\arg\min_{\bm{w}}\mathbb{E}_{(\bm{x}_r,{\bm{y}_r})\sim\mathcal{D}_r}\big{\{}\ell\big(\pi_{\bm{\theta}}(\bm{x}_r,\bm{y}_r,\bm{w}),\tilde{\bm{y}}_r|\bm{w}\big)\big{\}}\nonumber
    \end{aligned}
\end{equation}
which can be solved by unrolling a few gradient descent steps of the inner minimization to the outer minimization, similar to works~\cite{finn2017model,liu2018darts}. $\pi_{\bm{\theta}}$ consists of a network $h_{\bm{\theta}}(\cdot)$ that outputs the Mixup coefficient. Here the empirical risk on the validation set serves as a proxy to the distance to $\bm{w}^*$.

\textbf{Policy gradient}. Alternatively, we can also resort to the policy gradient approach~\cite{williams1992simple}. We can use the accuracy on the validation set as the reward signal $R$. Thus 
simply maximizing this terminal reward: $\min_{\bm{\theta}} J(\bm{\theta}):=\mathbb{E}_{\pi_{\bm{\theta}}}\{R\}$ leads to the update rule for the teacher: $\bm{\theta}\leftarrow \bm{\theta}+\eta\cdot\nabla_{\bm{\theta}}J(\bm{\theta})$ where $\nabla_{\bm{\theta}}J(\bm{\theta})=\sum_t\nabla_{\bm{\theta}}\log\pi_{\bm{\theta}}(a_t|s_t)R$ and $(a_t,s_t)$ is the state-action pair at the $t$-th iteration. For the state features, we use the current learner's predictions of representative samples. For the action space, we use a discrete Mixup coefficient space $\{0,0.5,1\}$ to reduce the search space. The overall training is conceptually similar to prior work~\cite{zoph2016neural}.

\begin{figure}
    \centering
    \includegraphics[width=0.9\linewidth]{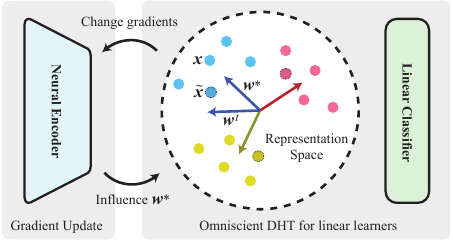}
    \vspace{-0.5mm}
    \caption{\footnotesize Performative teaching for black-box neural learners.}
    \label{fig:performative}
    \vspace{-2mm}
\end{figure}

\vspace{-1.75mm}
\subsection{Performative Teaching}
\vspace{-1.5mm}

The concept of performativity has been studied primarily in economics~\cite{healy2015performativity,mackenzie2007economists} and recently in machine learning~\cite{perdomo2020performative,hardt2022performative}. When supporting consequential decision-making, predictive models can produce actions that influence the outcome they aim to predict at the beginning. These predictions are called \emph{performative}. 

As shown in Figure~\ref{fig:performative}, we decompose a neural network into two components: a neural encoder $g_1(\cdot)$, which is used to extract features, and a linear classifier $g_2(\cdot)$, which is used to obtain class labels. Suppose we teach the last-layer classifier of a neural network with omniscient DHT and the teaching example will thus change the gradients for updating the neural encoder. Then after the neural encoder gets updated, the teaching target $\bm{w}^*$ will also be shifted because the data representation changes. The entire process is iteratively executed. Inspired by the striking connection between performativity and iterative teaching in the representation space, we introduce performative teaching where the teaching target $\bm{w}^*$ will change dynamically according to the teacher's action. In the context of teaching black-box neural learners, performative teaching is formulated as
\begin{equation}\label{eq:perfteach}
\footnotesize
    \min_{(\bm{x}_t,\bm{y}_t)} d\big( \bm{w}^t,\bm{w}^*(t)\big),~~~~
    \text{s.t.}~\bm{w}^*(t)\sim\mathcal{M}(\bm{x}_{t-1},\bm{y}_{t-1})
\end{equation}
where $\bm{w}^*(t)$ is the target parameters at the $t$-the iteration and $\mathcal{M}(\bm{x}_{i},\bm{y}_{i})$ denotes the distribution of the target learner parameters that is dynamically dependent on the teaching example $(\bm{x}_{i},\bm{y}_{i})$. In training, Equation~\ref{eq:perfteach} is solved alternately with the gradient update for the neural encoder.

If we want to use performative teaching to train neural learners in practice, we still need to consider a few unresolved problems. First, we have to estimate $\bm{w}^*$ in each iteration. Because the teaching is performed for linear classifiers, estimating $\bm{w}^*$ is relatively easy. We simply run a few more gradient descent steps to update the last-layer linear classifiers with the neural encoder fixed, and the resulting classifier weights are viewed as an approximate $\bm{w}^*$. Second, we need to develop a concrete algorithm to minimize $d( \bm{w}^t,\bm{w}^*(t))$ even if $\bm{w}^*(t)$ can be estimated. We resort to the simplest greedy teaching algorithm. In order to preserve the semantic meaning of the original feature $\bm{x}$ (given its ground truth label $\bm{y}$) and to remove potential degenerate solutions, we optimize the teaching example in an $\epsilon$-neighborhood of $\bm{x}$, \eg, $\|\tilde{\bm{x}}- \bm{x}\|\leq\epsilon$. Combining all the pieces, we summarize our performative teaching algorithm for training a black-box neural learners in Algorithm~\ref{alg:perform_teach}. After we obtain the teaching examples $\tilde{\bm{x}}_j^i$ with greedy DHT, we replace the original $\bm{x}^i_j$ with $\tilde{\bm{x}}_j^i$ during training. This is essentially to add a perturbation to the original feature and the backward gradients to update the network will be affected. In practice, the computational overhead is reasonably small as long as we use a small number of steps to estimate $\bm{w}^*$ in each iteration. Instead of using a $\epsilon$-neighborhood of the original feature in the Euclidean space, we use a $\epsilon$-neighborhood on the hypersphere with the radius being the norm of the original feature in Equation~\ref{eq:pertteac_alg}. This is inspired by the observation in \cite{liu2017sphereface,wang2018cosface,deng2019arcface,liu2018learning,liu2018decoupled,chen2020angular} that angular distance in the representation space tends to model semantic difference.

\begin{algorithm}[t]
\footnotesize
\caption{\footnotesize Performative teaching for neural learners}\label{alg:perform_teach}

\vspace{0.25mm}
\underline{\textbf{1}}. Randomly initialize the neural network. We denote the neural weights of $g_1$ as $\bm{v}$, and the neural weights of $g_2$ as $\bm{w}$\.;

\For{$i=1,2,\cdots,T_1$}{

\underline{\textbf{2}}. Form a mini-batch of $m$ samples and perform inference to extract features, denoted as $(\bm{x}_1^i,\bm{y}_1^i),\cdots,(\bm{x}_m^i,\bm{y}_m^i)$.\;

\underline{\textbf{3}}. $\bm{w}_{\text{buffer}}\leftarrow\bm{w}$.\;

\underline{\textbf{4}}. Fix $\bm{v}$ and update $\bm{w}$ by minimizing the empirical risk on the training set (\eg,  a few SGD steps).\;

\underline{\textbf{5}}. $\bm{w}^*\leftarrow\bm{w}$ and then $\bm{w}\leftarrow\bm{w}_{\text{buffer}}$.\;

\For{$j=1,2,\cdots,m$}{

\underline{\textbf{6}}. Solve the greedy teaching problem for the $j$-th sample: 
\vspace{-1.5mm}
\begin{equation}\label{eq:greedy_linear_algo}
    \scriptsize
    \begin{aligned}
        &\tilde{\bm{x}}^{i}_j=\arg\min_{\bm{x}}  ~\eta^2_{t} \norm{ \frac{\partial \ell(\bm{x},\bm{y}^i_j|\bm{w}) }{\partial \bm{w}} }^2_2  \\&~~~~~~~~~~~~~~~- 2 \eta_{t} \langle \bm{w} - \bm{w}^*, \frac{\partial \ell(\bm{x},\bm{y}^i_j|\bm{w}) }{\partial \bm{w}} \rangle\\
        &\text{s.t.}~\norm{\frac{\bm{x}}{\|{\bm{x}}\|}-\frac{\bm{x}_j^i}{\|{\bm{x}_j^i}\|}}\leq \epsilon,~~\|\bm{x}\|=\|\bm{x}_j^i\|
        \end{aligned}
\end{equation}
\vspace{-3.6mm}
\;

}

    \underline{\textbf{7}}. Use SGD to update the neural network ($\bm{w}$ and $\bm{v}$) by replacing $(\bm{x}_1^i,\bm{y}_1^i),\cdots,(\bm{x}_m^i,\bm{y}_m^i)$ with $(\tilde{\bm{x}}_1^i,\bm{y}_1^i),\cdots,(\tilde{\bm{x}}_m^i,\bm{y}_m^i)$.\;
    \vspace{-0.25mm}
    
}
\vspace{-0.6mm}
\end{algorithm}

\vspace{-0.3mm}

Performing DHT iteratively in the representation space can provide additional information for the last-layer classifiers, leading to better convergence of the last-layer classifiers. This will in turn improve the backward gradients of the loss \emph{w.r.t.} the representation which is responsible for updating the network encoder. From an optimization perspective, performative teaching shares similar spirits to Lookahead optimizer~\cite{zhang2019lookahead} in the sense that both methods use information about future steps, which can be viewed as an approximate form of $\bm{w}^*$. With the surrogate knowledge of $\bm{w}^*$ for the last-layer classifier, DHT implicitly encodes more information about the loss landscape and may help the neural encoder to avoid some poor local minima.

\vspace{-2.2mm}
\section{Experiments and Results}
\vspace{-1.8mm}

We evaluate DHT on several widely used image classification datasets: for white-box and black-box teaching in the logistic regression, we test our policies on synthetic half-moon and MNIST; for black-box teaching in deep neural networks, we test our methods on MNIST, CIFAR-10, and CIFAR-100. Full experiment details and additional experimental results can be found in Appendix.

\vspace{-2.1mm}
\subsection{Omniscient Teaching}
\vspace{-1.75mm}

In the omniscient teaching scenario, we seek to optimize the convergence speed to a target classifier $w^*$, and because the target classifier exhibits good classification performance, we also measure the convergence of the testing accuracy. We initialize all the models with the same architecture and model weights. All experiments are repeated ten times with different seeds. We compare random teacher (\ie, SGD), samples selected by IMT, and samples generated by DHT. We update the student model with a standard SGD optimizer, a learning rate of 0.001, and compare the convergence behavior in the first 300 iterations. 

\vspace{-0.2mm}

\begin{figure}[t!]
    \centering
    \begin{subfigure}{.48\linewidth}
      \centering
      \includegraphics[width=.99\linewidth]{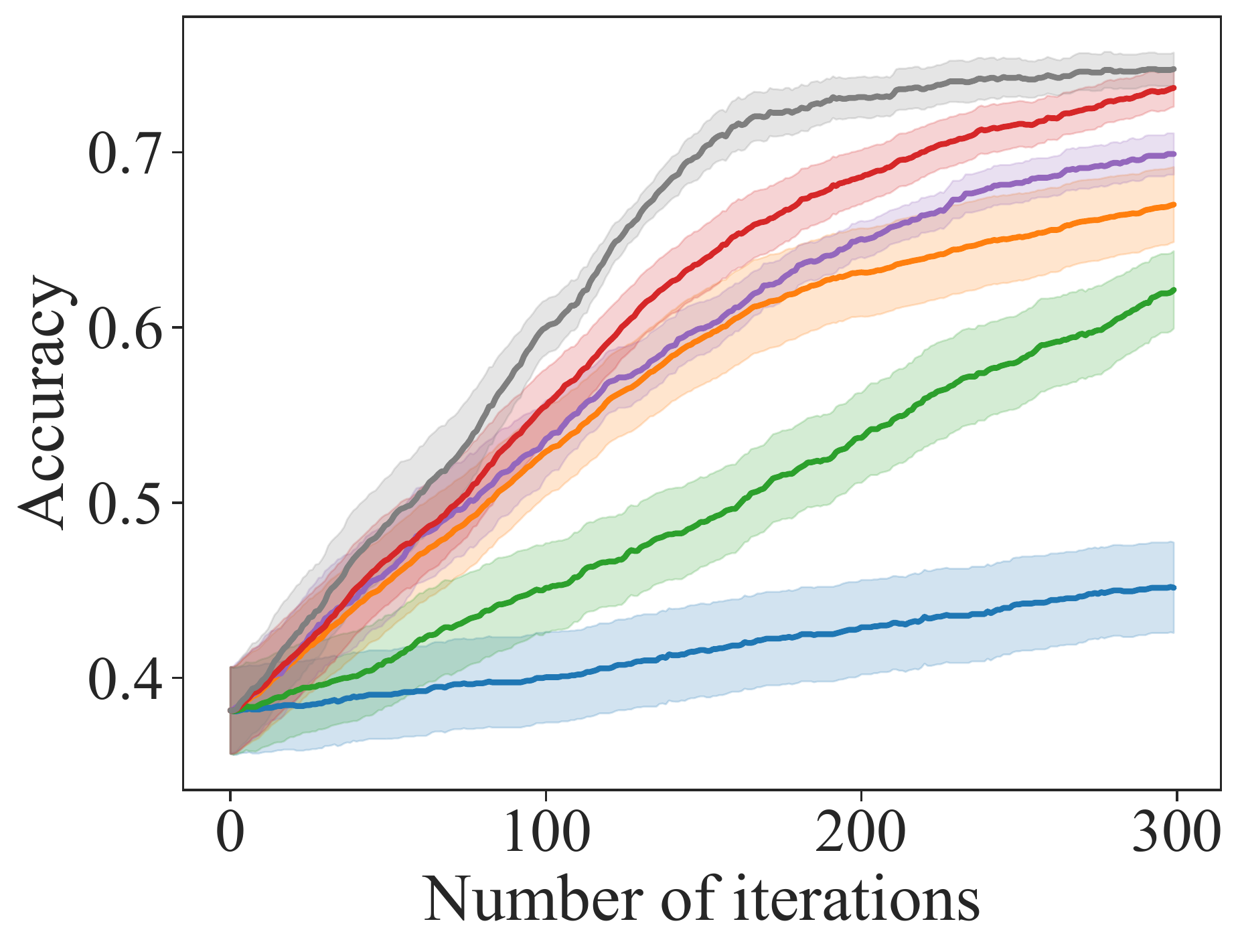}
    \end{subfigure}%
    \begin{subfigure}{.48\linewidth}
      \centering
      \includegraphics[width=.99\linewidth]{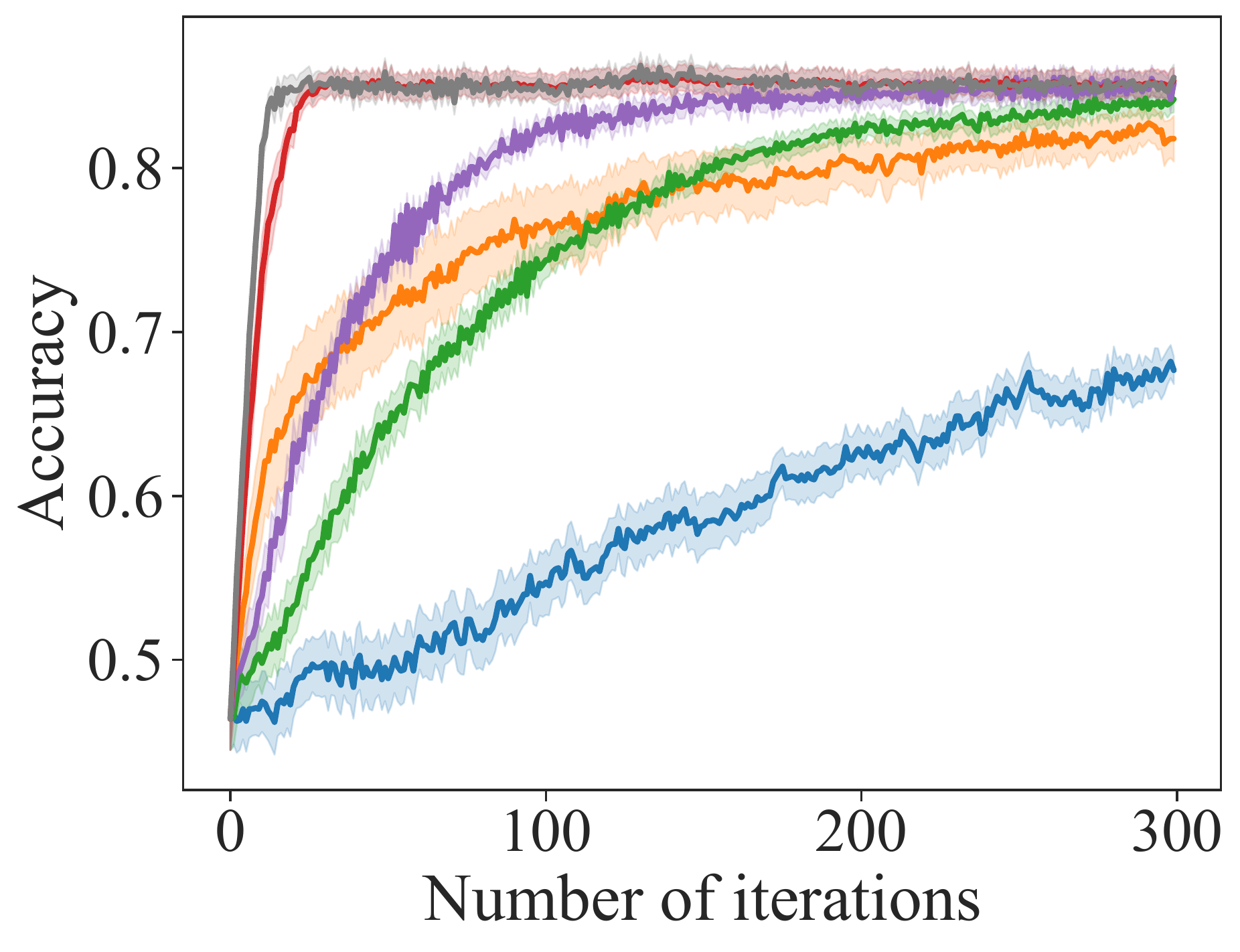}
    \end{subfigure}
    \begin{subfigure}{.48\linewidth}
      \centering
      \includegraphics[width=.99\linewidth]{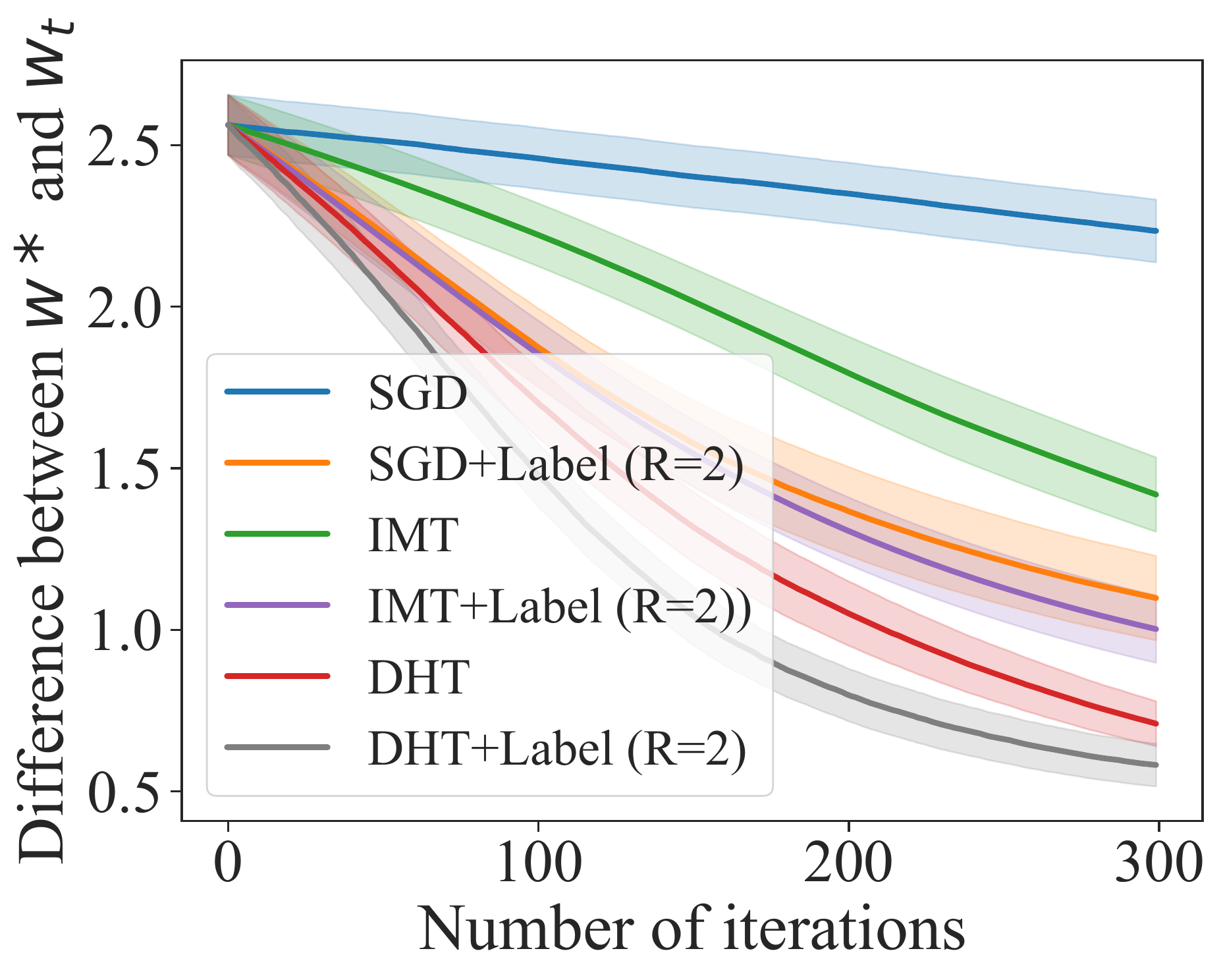}
    \end{subfigure}%
    \begin{subfigure}{.48\linewidth}
      \centering
      \includegraphics[width=.99\linewidth]{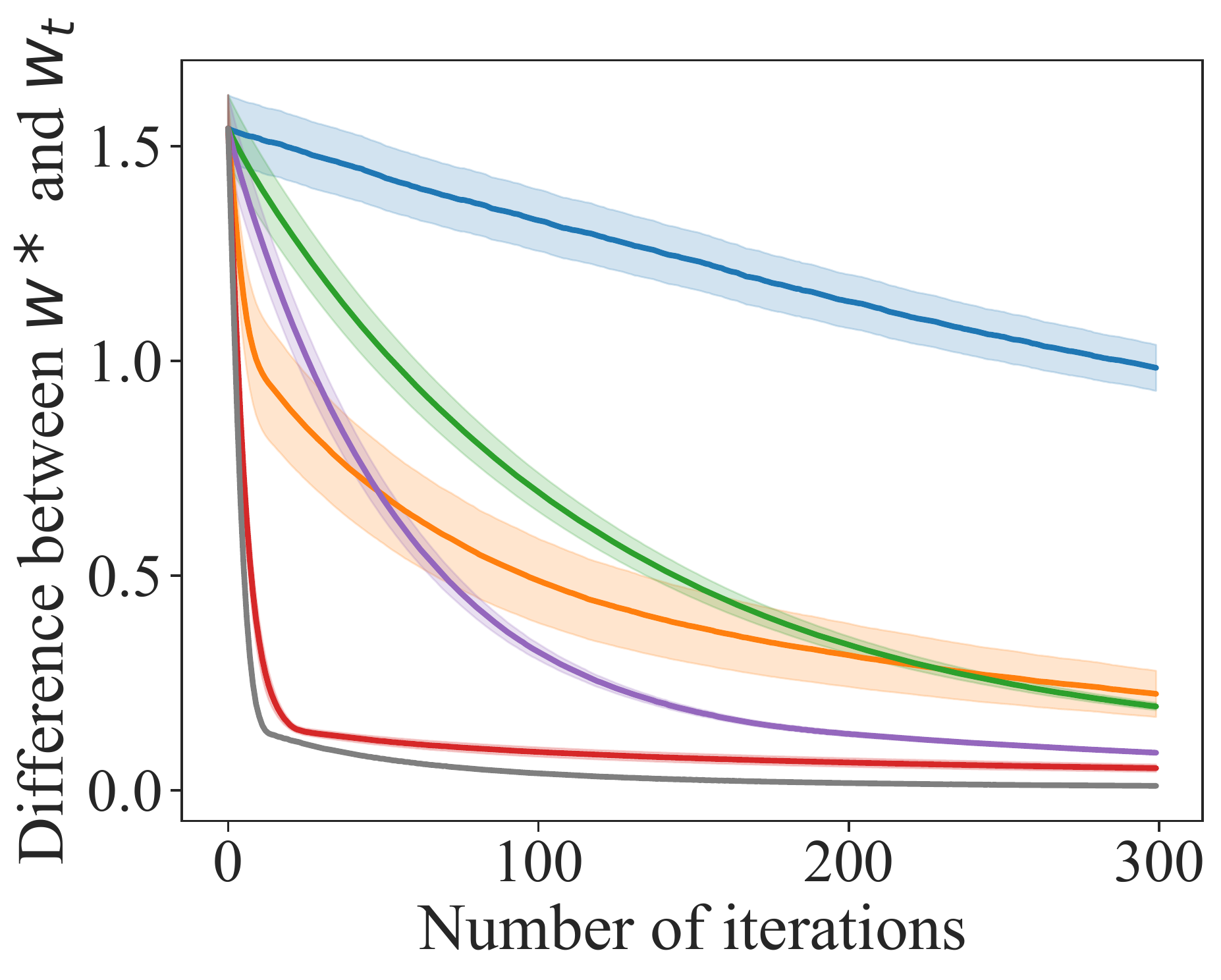}
    \end{subfigure}
    \vspace{-1.7mm}
    \caption{\footnotesize Omniscient teaching with or without label synthesis~\cite{liu2021iterative}. Convergence comparison between our greedy teaching policy with several other baseline methods. Left: half-moon. Right: MNIST.}
    \vspace{-2mm}
    \label{fig:qualitative_results_greedy}
\end{figure}

\textbf{Greedy teaching policy.} For greedy teaching, we additionally optimize the labels with \cite{liu2021iterative}, \ie, after one sample has been selected or generated, we further optimize the corresponding label by fixing the sample. The results are summarized in Figure~\ref{fig:qualitative_results_greedy}. The baseline methods SGD+Label and IMT+Label correspond to variants of the LAST framework~\cite{liu2021iterative}. For samples generated by DHT, we constrain the value of each dimension to be within the minimum and maximum of the original dataset; for the optimized one-hot labels, we constraint its values to be positive $y_i > 0$ and the magnitude to satisfy $\| \bm{\tilde{y}} \|_2 \leq 2$. Constraining the magnitude to 2 gives the teacher more flexibility when generating labels than the original one-hot label space. We observe that with or without label synthesis, greedy DHT achieves the fastest convergence and outperforms both SGD and IMT.

\vspace{-0.2mm}

\begin{figure}[t!]
    \centering
    \vspace{-1.5mm}
    \begin{subfigure}{.48\linewidth}
      \centering
      \includegraphics[width=.99\linewidth]{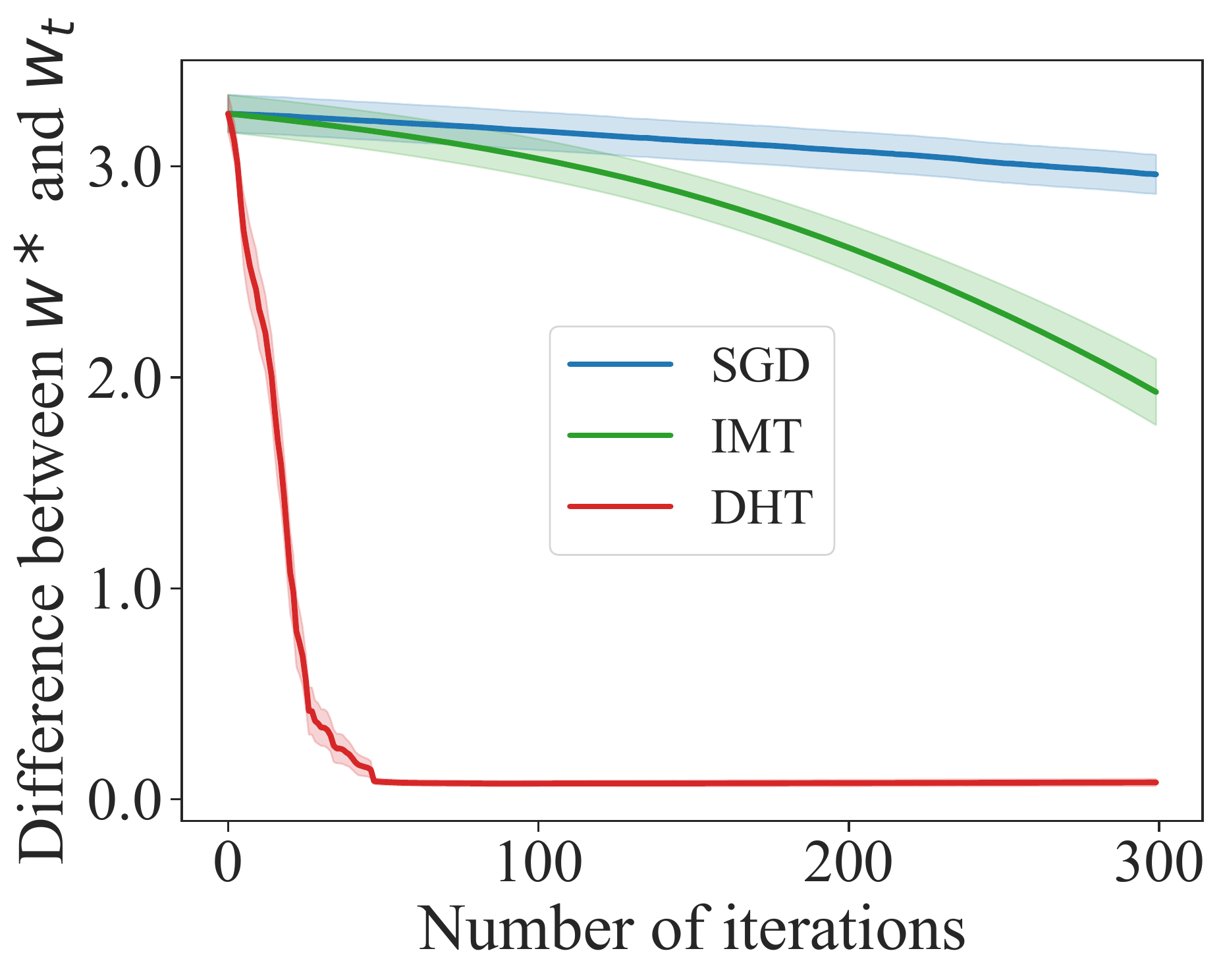}
    \end{subfigure}%
    \begin{subfigure}{.48\linewidth}
      \centering
      \includegraphics[width=.99\linewidth]{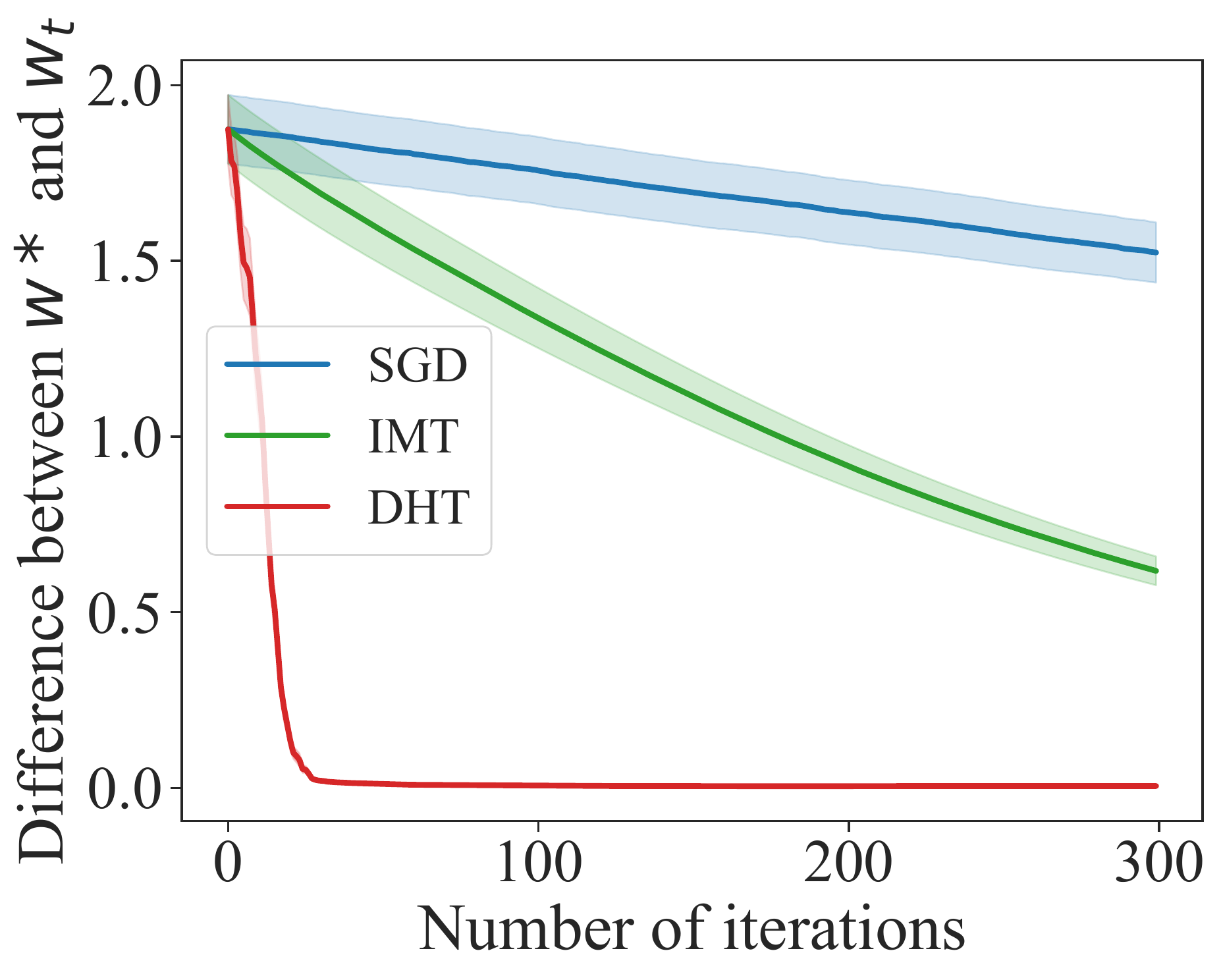}
    \end{subfigure}
    \vspace{-2mm}
    \caption{\footnotesize Convergence of data transformation policy. Left: binary classification on half-moon. Right: 3/5 classification on MNIST.}
    \vspace{-1.2mm}
    \label{fig:qualitative_unrolled}
\end{figure}

\textbf{Parameterized teaching policy}. The results of the data transformation policy are given in Figure~\ref{fig:qualitative_unrolled}. We generally observe that parameterized policy exhibits faster convergence than greedy policy, and DHT again outperforms both SGD and IMT by a significant margin. It is worth mentioning that we do not impose any constraint on the output space when synthesizing samples, resulting in out-of-distribution samples that are not semantically interpretable for humans (\eg, on MNIST it appears to be random noise images).

Next, we discuss the difference between GAN-based and VAE-based teaching policies. The weight convergence is given in Figure~\ref{fig:generative_modeling} and some generated samples are exemplified below in Figure~\ref{fig:vae_samples_moon} and Figure~\ref{fig:samples_mnist}. The GAN-based teacher is able to outperform the VAE-based teacher in both experiments. In general, we observe that by synthesizing samples directly in the image space, the teacher has more freedom in conveying information, causing the student to learn faster. However, this also means that the data synthesized by a GAN-based teacher might appear to be visually very different from the original data set. A VAE-based teacher can synthesize samples that appear much closer to the original data distribution at the cost of teaching performance. We observe in Figure~\ref{fig:vae_samples_moon} and Figure~\ref{fig:samples_mnist} that the synthesized samples are nearly differentiable from the original data set. One interesting observation is that the GAN-based teacher implicitly learns a meaningful teaching policy. First, samples are synthesized that have certain stylistic characteristics of the original MNIST dataset but do not contain any meaningful semantic information. During this stage, the student can efficiently converge to $w^*$; afterward, the GAN-based teacher generates samples that look more similar to the original dataset, and we can interpret it as a fine-tuning process. The evolution of the generated samples during the teaching process can be seen in Figure~\ref{fig:samples_mnist}.

\begin{figure}[t!]
    \centering
    \begin{subfigure}{.48\linewidth}
      \centering
      \includegraphics[width=.99\linewidth]{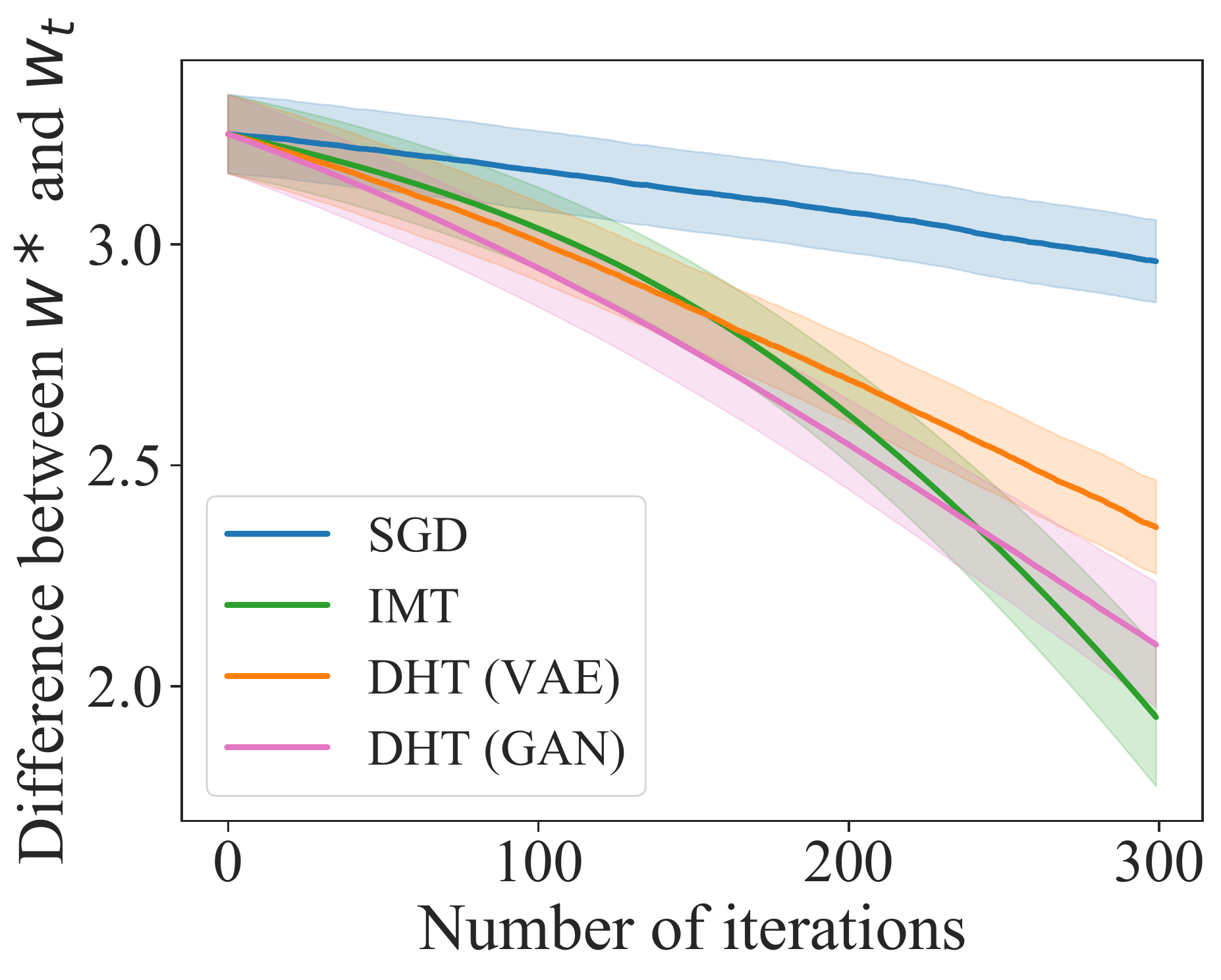}
    \end{subfigure}%
    \begin{subfigure}{.48\linewidth}
      \centering
      \includegraphics[width=.99\linewidth]{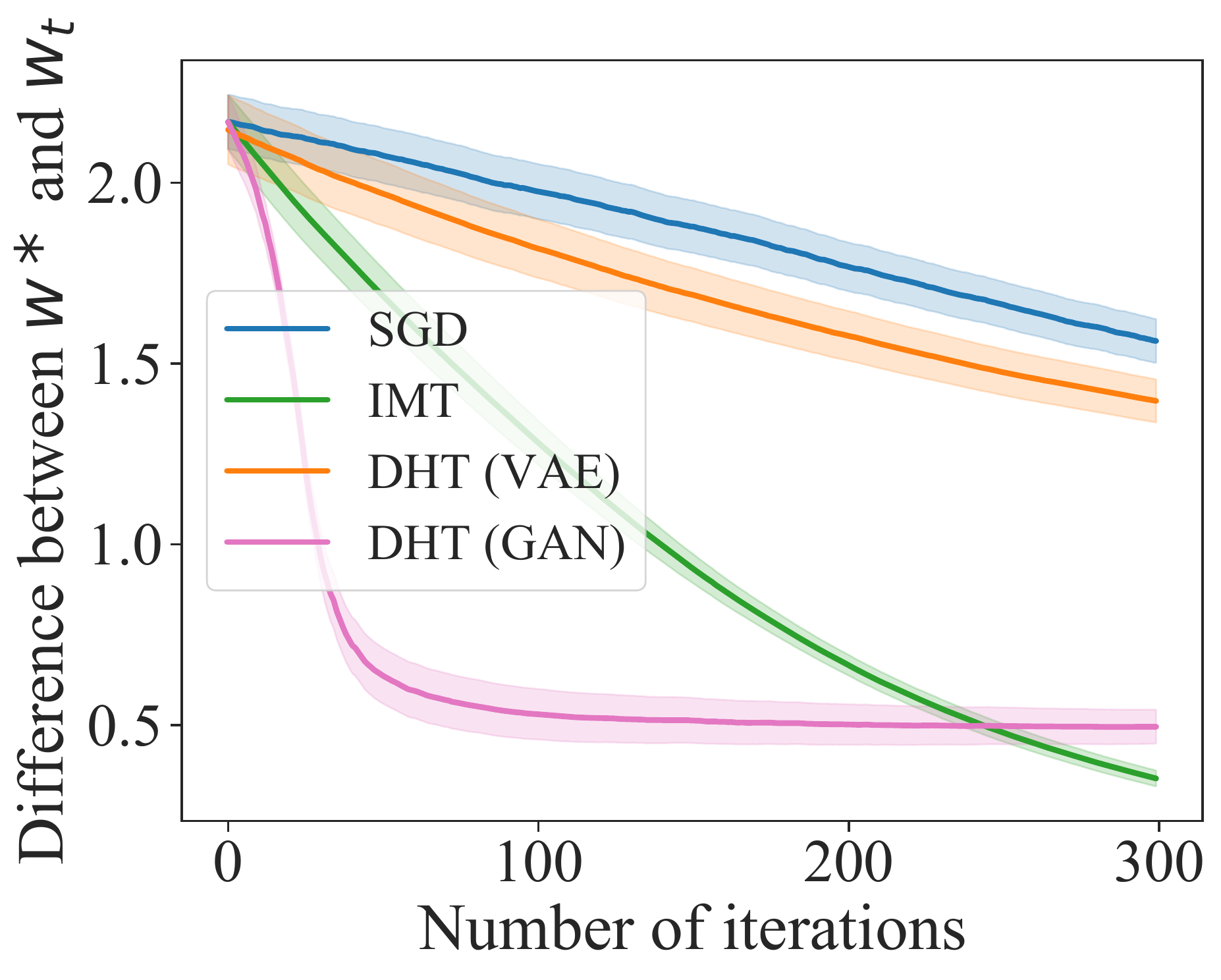}
    \end{subfigure}
    \vspace{-2mm}
    \caption{\footnotesize Convergence of GAN-based and VAE-based generative modeling policy. Left: half-moon. Right: MNIST.}
    \vspace{-2.7mm}
    \label{fig:generative_modeling}
\end{figure}

\begin{figure}
    \centering
    \begin{subfigure}{.33\linewidth}
        \centering
        \includegraphics[width=0.99\linewidth]{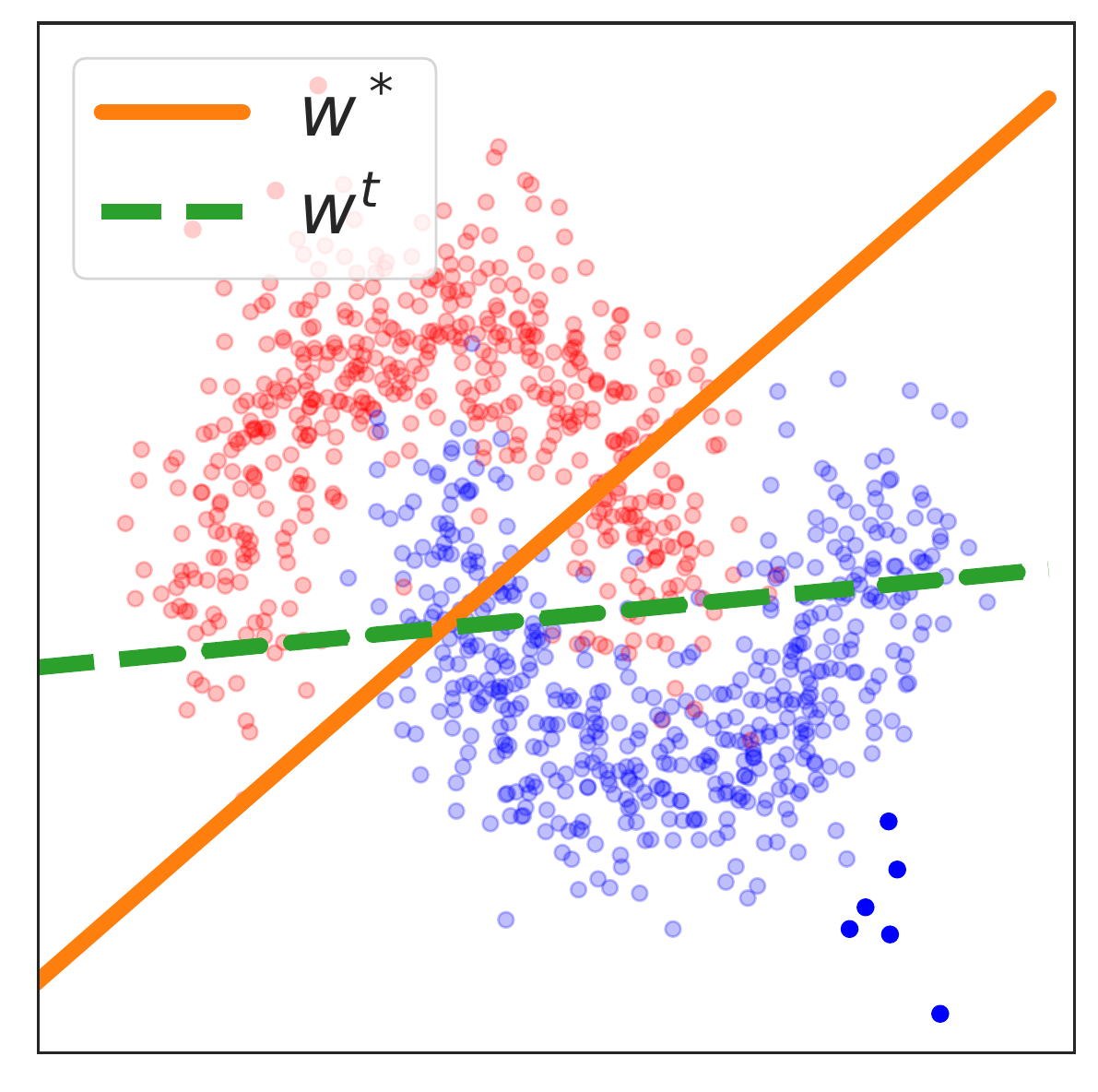}
    \end{subfigure}%
    \begin{subfigure}{.33\linewidth}
        \centering
        \includegraphics[width=0.99\linewidth]{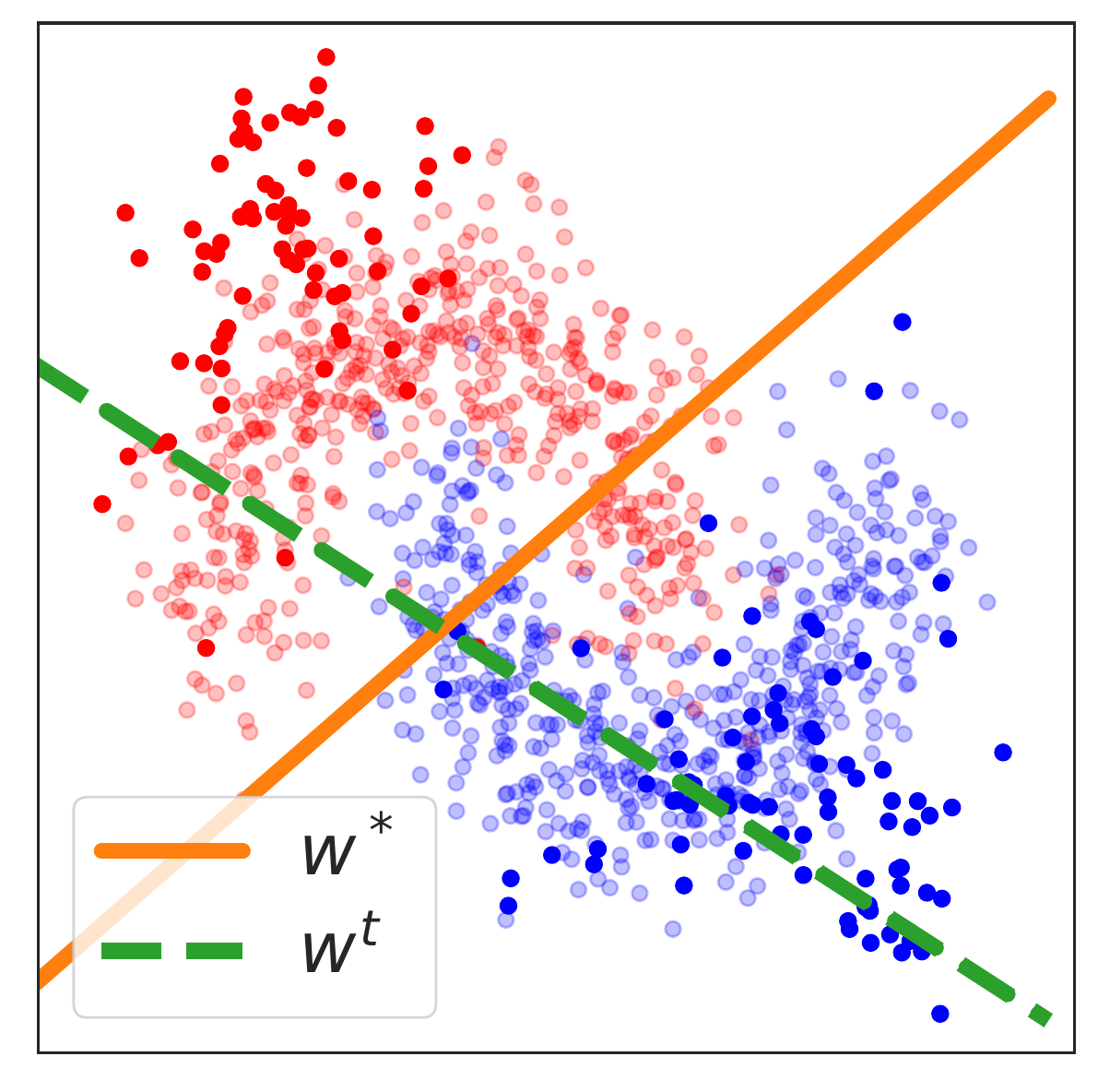}
    \end{subfigure}%
    \begin{subfigure}{.33\linewidth}
        \centering
        \includegraphics[width=0.99\linewidth]{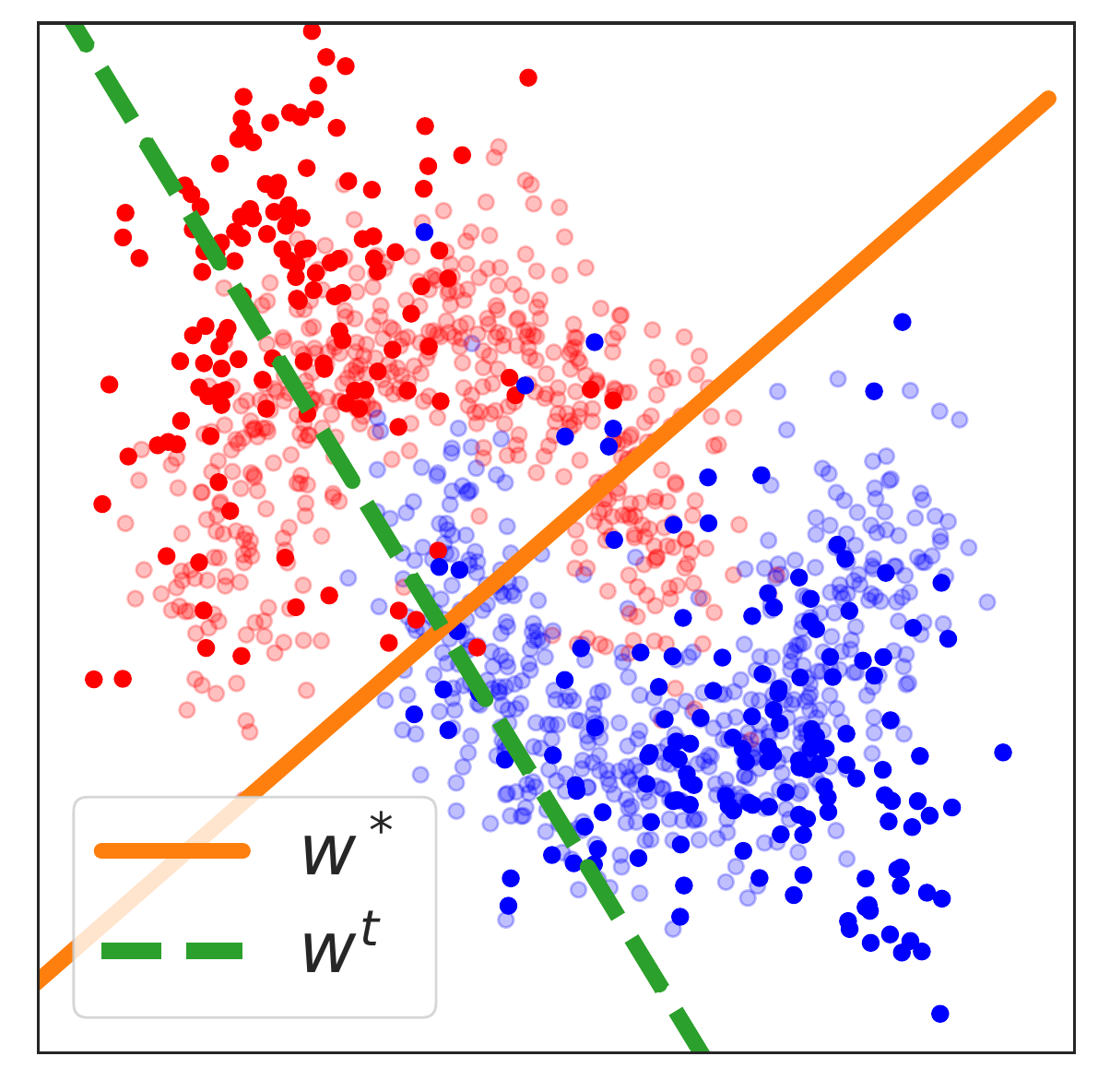}
    \end{subfigure}
    \vspace{-2mm}
    \caption{\footnotesize Visualization of the data synthesized by a VAE-based teacher after iterations 10, 150, and 290. The orange line indicates the target classifier $w^*$; the green dashed line indicates the student classifier. Different colors indicate different classes; points with lower opacity represent the ground truth data distribution.}
    \label{fig:vae_samples_moon}
    \vspace{-0.1mm}
\end{figure}

\begin{figure}[t!]
    \centering
    \includegraphics[width=0.99\linewidth]{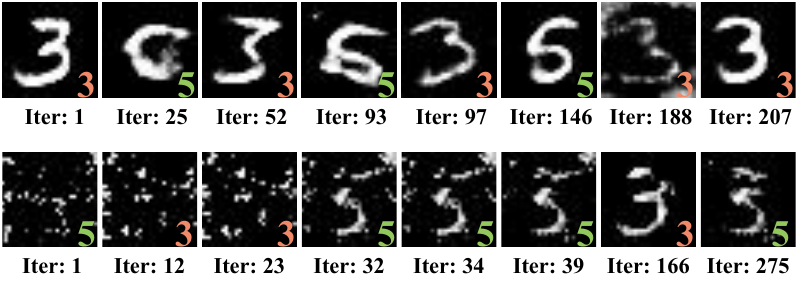}
    \vspace{-2mm}
    \caption{\footnotesize MNIST samples synthesized by the DHT teacher conditioned on GT labels (3/5) during the training process. Upper row: samples synthesized by the VAE-based teacher. Lower row: samples synthesized by the GAN-based teacher.}
    \vspace{-0.2mm}
    \label{fig:samples_mnist}
\end{figure}

\vspace{-2mm}
\subsection{Black-box Teaching}
\vspace{-1.7mm}

\begin{figure}[t!]
    \centering
    \begin{subfigure}{.48\linewidth}
        \centering
        \includegraphics[width=0.99\linewidth]{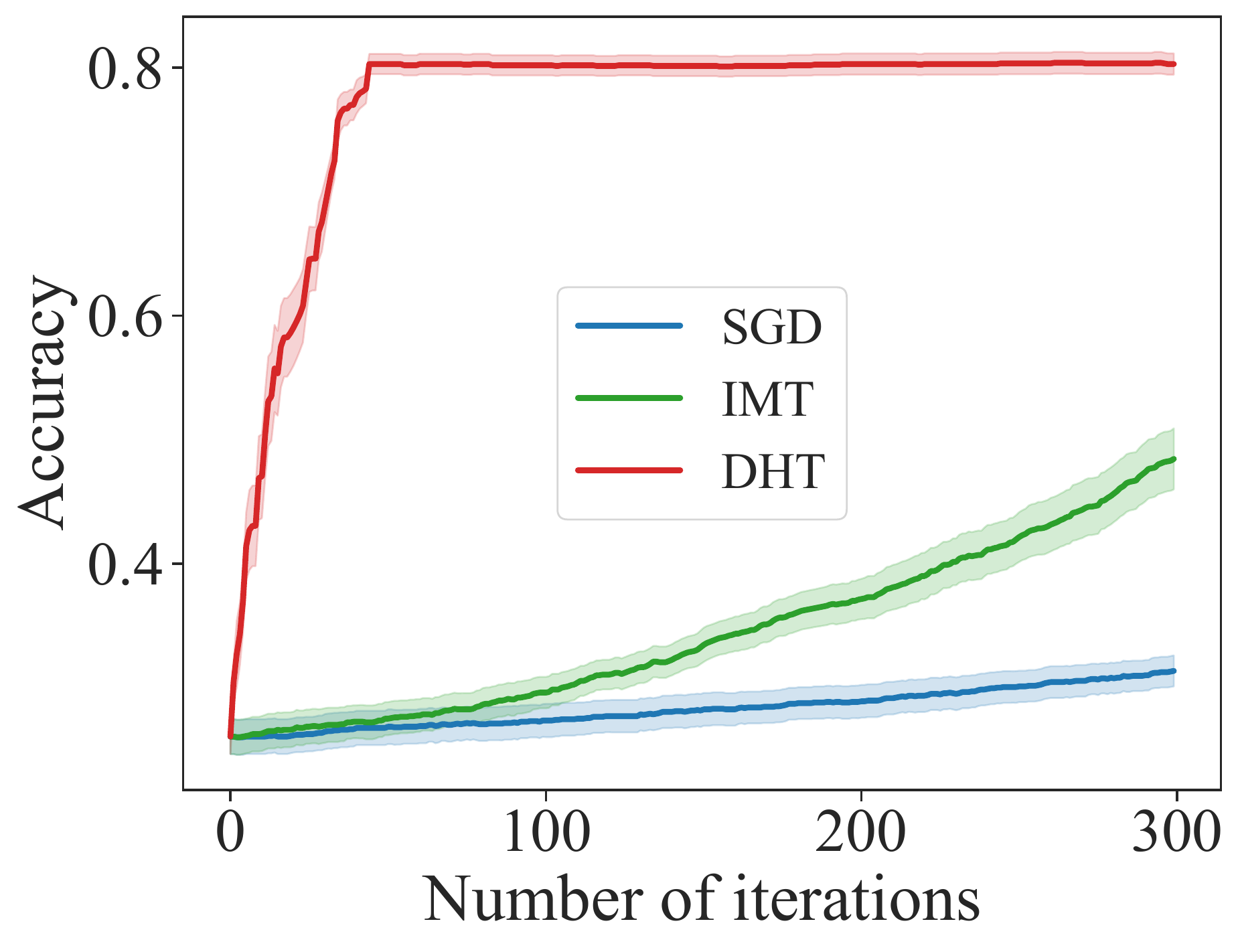}
    \end{subfigure}%
    \begin{subfigure}{.48\linewidth}
        \centering
        \includegraphics[width=0.99\linewidth]{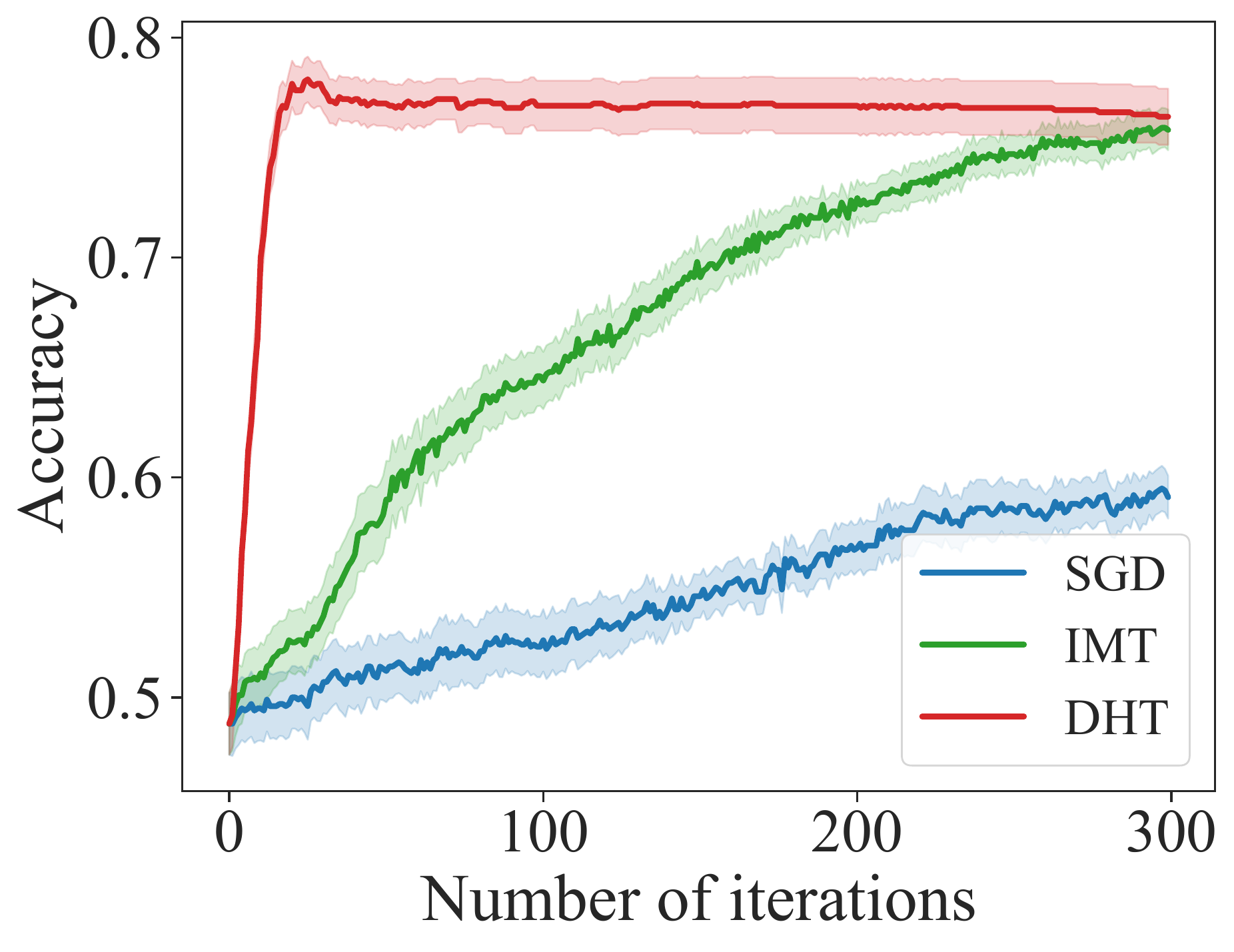}
    \end{subfigure}
    \vspace{-2mm}
    \caption{\footnotesize Convergence of black-box parameterized teaching policy with the full teaching space. Left: half-moon. Right: MNIST.}
    \vspace{-2.5mm}
    \label{fig:blackbox_unrolled}
\end{figure}

\textbf{Parameterized teaching}. We start with the black-box version of parameterized teaching. We use the same teaching objective as the parameterized teaching in the omniscient setting and remove any information about the target classifier $w^*$, \ie, by relying solely on the empirical classification loss on the validation set. Surprisingly, when teaching a linear logistic regression learner, the knowledge about the target classifier is not indispensable to achieve fast convergence in terms of test accuracy (see Figure~\ref{fig:blackbox_unrolled}).

\setlength{\columnsep}{9pt}
\begin{wraptable}{r}[0cm]{0pt}
\footnotesize
 \centering
 \setlength{\tabcolsep}{3pt}
 \renewcommand{\arraystretch}{1.2}
 \hspace{-1mm}
 \scalebox{0.93}{
 \begin{tabular}{ccc}
 \specialrule{0em}{-14pt}{0pt}
 Method & Accuracy~(\%)\\
  \shline
    ERM         & $61.75$ \\
    cMixup      & $66.27$ \\
    dMixup      & $65.80$ \\
    \cellcolor{Gray}{Unrolling} & \cellcolor{Gray}{$65.18$} \\
    \cellcolor{Gray}{Policy gradient}         & \cellcolor{Gray}{$\bm{67.64}$} \\
  \specialrule{0em}{-6pt}{0pt}
 \end{tabular}}
 \caption{\footnotesize Results on CIFAR-10.} \label{tab:mixup_teaching}
 \vspace{-2mm}
\end{wraptable}

\textbf{Mixup-based teaching}.\! We empirically evaluate our mixup-based teaching with unrolling and policy gradient in Table~\ref{tab:mixup_teaching}. The learner is a simple CNN model optimized by a standard Adam optimizer on CIFAR-10 without data augmentation for 50 epochs. We compare our Mixup-based teaching policy with standard empirical risk minimization (ERM), ERM with continuous mixup data augmentation (\ie, interpolation coefficient sampled from a beta distribution, denoted as {cMixup}), and ERM with discrete mixup data augmentation (\ie, mixing coefficient discretized to $\{0, 0.5, 1\}$, denoted as {dMixup}). With policy gradient, we observe that our DHT teacher is able to find a policy that greatly facilitates the convergence of the student model and outperforms the other baselines. Note that we characterize the model states with model features that are obtained through querying the student model, similar to \cite{liu2021iterative}. We also notice that unrolling does not perform as well as the baselines, and we suspect that this is because the space of Mixup coefficients is highly non-smooth. There are many poor local minima that prevent the unrolling approach from finding a good solution to the bi-level optimization.

\begin{table}[t]
\scriptsize
 \centering
 \setlength{\tabcolsep}{3.5pt}
 \renewcommand{\arraystretch}{1.2}
 \begin{tabular}{c|cccc}
  {Dataset} & {Learner} & {SGD} & Random Policy & \cellcolor{Gray}{DHT}\\
  \shline
  \multirow{1}{*}{MNIST}    & MLP  & $92.45 \pm 0.07$ & $92.47 \pm 0.06$ & \cellcolor{Gray}{$\bm{95.02 \pm 0.04}$}\\
  \hline
  \multirow{4}{*}{CIFAR-10} & CNN-3  & $87.30 \pm 0.28$ & $87.17 \pm 0.17$ & \cellcolor{Gray}{$\bm{88.77 \pm 0.35}$}\\
                            & CNN-6  & $90.34 \pm 0.10$ & $90.20 \pm 0.09$ & \cellcolor{Gray}{$\bm{91.61 \pm 0.23}$}\\
                            & CNN-9  & $91.10 \pm 0.26$ & $91.12 \pm 0.12$ & \cellcolor{Gray}{$\bm{92.30 \pm 0.13}$}\\
                            & CNN-15 & $91.85 \pm 0.28$ & $91.67 \pm 0.13$ & \cellcolor{Gray}{$\bm{92.44 \pm 0.15}$}\\
  \hline
  \multirow{4}{*}{CIFAR-100}& CNN-3  & $62.10 \pm 0.29$ & $62.04 \pm 0.11$ & \cellcolor{Gray}{$\bm{62.69 \pm 0.37}$}\\
                            & CNN-6  & $65.02 \pm 0.24$ & $64.96 \pm 0.17$ & \cellcolor{Gray}{$\bm{66.81 \pm 0.17}$}\\
                            & CNN-9  & $67.05 \pm 0.29$ & $67.19 \pm 0.23$ & \cellcolor{Gray}{$\bm{69.23 \pm 0.34}$}\\
                            & CNN-15 & $68.39 \pm 0.39$ & $68.49 \pm 0.17$ & \cellcolor{Gray}{$\bm{68.96 \pm 0.36}$}\\
  \specialrule{0em}{-7pt}{0pt}
 \end{tabular}
\caption{\footnotesize Testing accuracy (\%) of performative teaching. Multiple types of neural learners (\eg, MLP and CNN) are considered.} \label{tab:performative_teaching} 
 \vspace{-3.5mm}
\end{table}

\vspace{-0.35mm}

\textbf{Performative teaching for neural learners}. We comprehensively evaluate the performance of the performative teaching by conducting image classification experiments with similar architectures and settings as \cite{liu2018learning}. For CIFAR-10 and CIFAR-100, we start the training with the learning rate of $0.1$ and divide it by $10$ at iteration 20k, 30k and 37.5k. The training stops at iteration 42.5k. For MNIST, we start the training with a learning rate of $0.001$ and train for 39k iterations. We use a standard SGD optimizer with weight decay. The batch size is set to 128 and only basic data augmentation is performed. Multiple network architectures are used to serve as the learner and the specific architectures are given in Appendix. Results are given in Table~\ref{tab:performative_teaching}.

\vspace{-0.35mm}

We compare our performative teaching with two other baselines. The first one is vanilla SGD optimization, where no teaching takes place during the training. This is to demonstrate the clean performance gain obtained by performative teaching. From Table~\ref{tab:performative_teaching}, we can clearly see that DHT consistently outperforms vanilla SGD by a considerable margin. To further verify whether DHT indeed teaches useful information or not, we construct a random policy in the exact same action space of the performative teacher, \ie, we omit the teaching process, but instead uniformly sample a new point on the same $\epsilon$-neighborhood on the hypersphere of the representation space. The only difference between random policy and performative teaching is how we generate $\tilde{\bm{x}}$, and we note that the space to generate $\tilde{\bm{x}}$ is the same for both. This comparison shows that the performance gain does not result from implicit data augmentation in the representation space, as can be seen from Table~\ref{tab:performative_teaching} that a random policy does not have a noticeable effect on the final performance. All the results in Table~\ref{tab:performative_teaching} are averaged over 5 runs and the standard deviations of accuracy are also given to make sure that the performance gain is not due to randomness. The performance gain of performative teaching is evident across all datasets and all different neural learners. We emphasize that we do not have $\bm{w}^*$ for the neural networks, so performative teaching can be used to teach any neural network on any dataset. We only consider a simple greedy DHT in performative teaching, and the teaching performance could be further improved with an advanced teaching algorithm.

\vspace{-2mm}
\subsection{Privacy-preserving Teaching by constrained DHT}
\vspace{-1.8mm}

\begin{figure}[t!]
    \centering
    \vspace{-2mm}
    \begin{subfigure}{.48\linewidth}
      \centering
      \includegraphics[width=.99\linewidth]{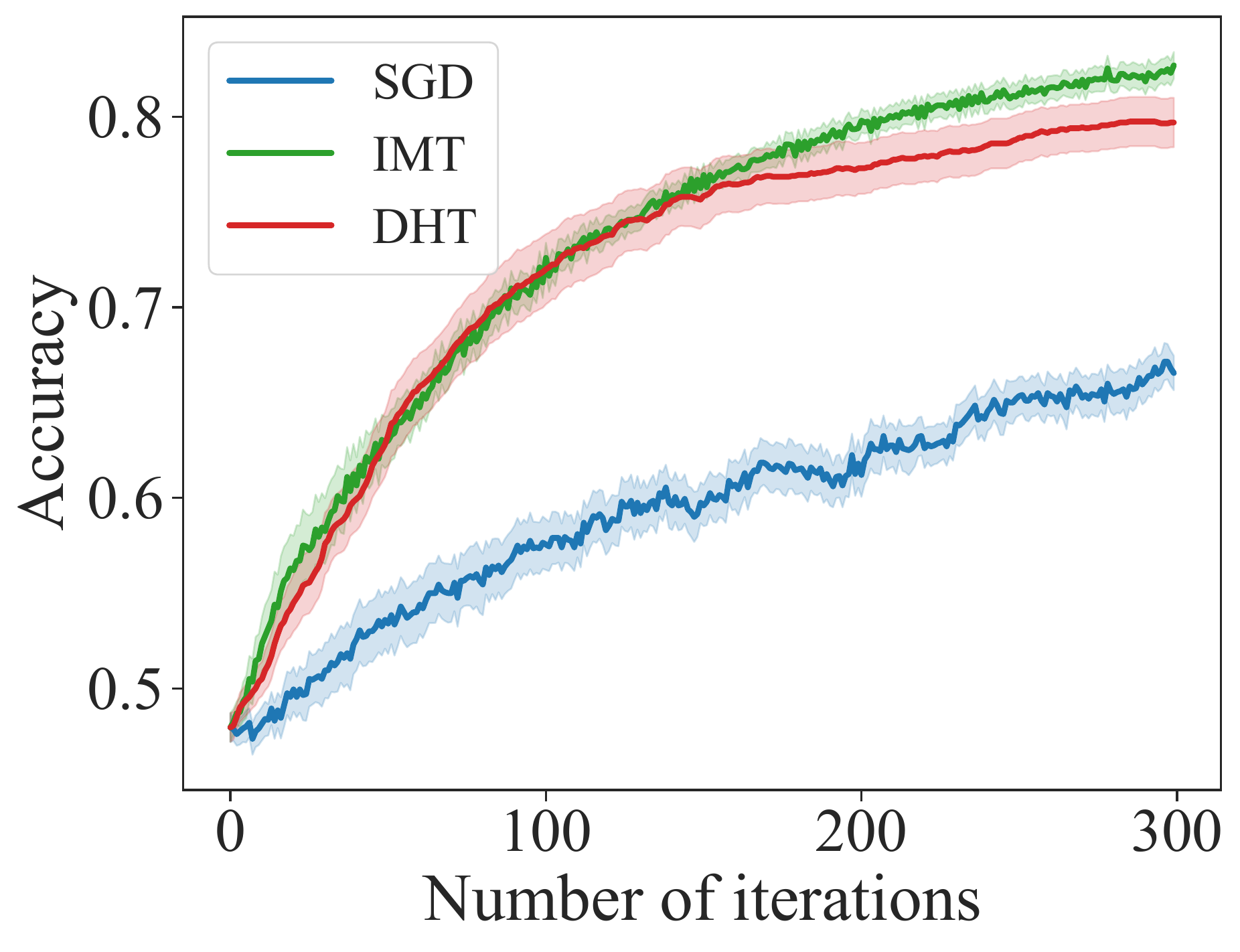}
    \end{subfigure}%
    \begin{subfigure}{.48\linewidth}
      \centering
      \includegraphics[width=.99\linewidth]{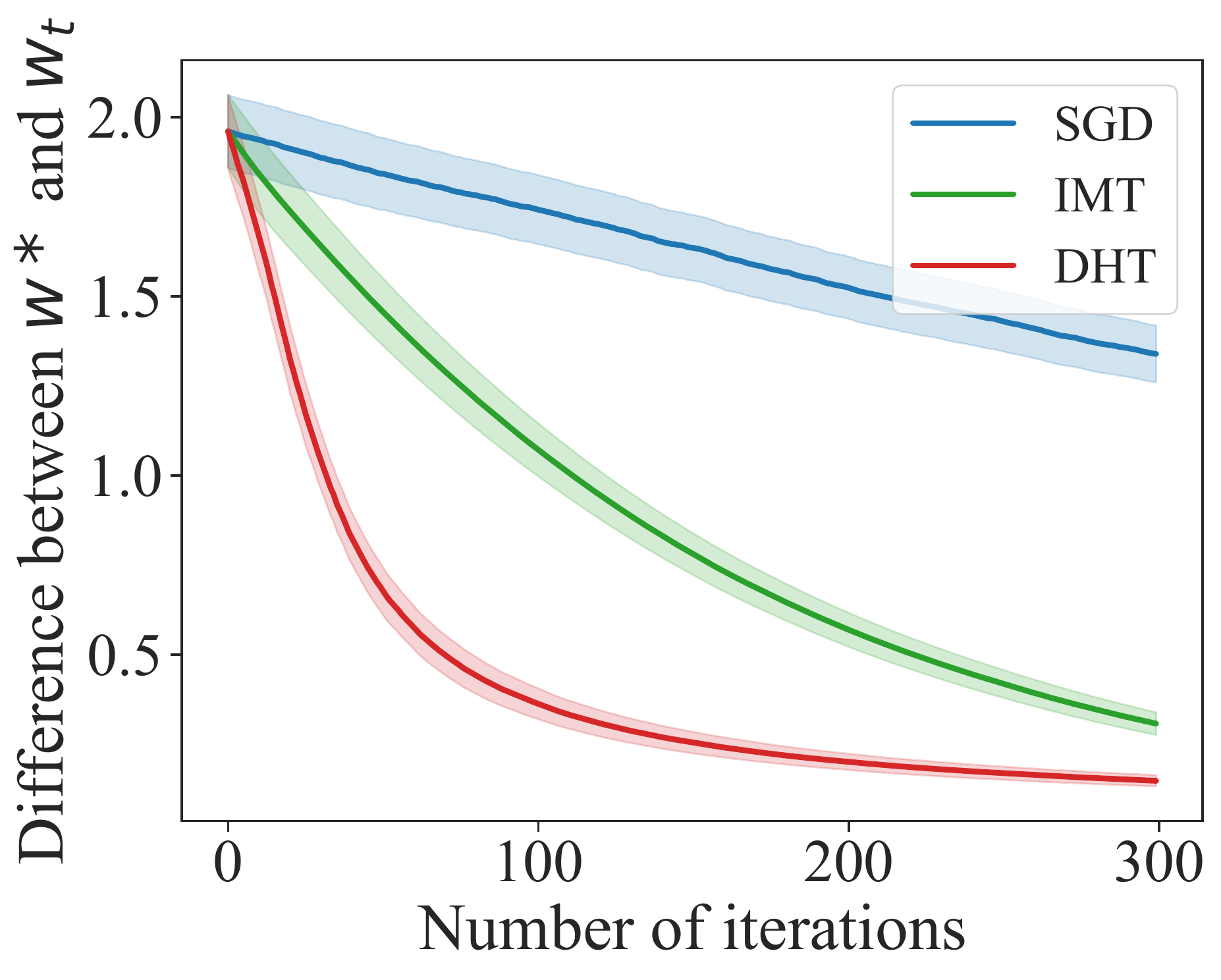}
    \end{subfigure}
    \vspace{-2mm}
    \caption{\footnotesize Convergence of privacy-preserving teaching on MNIST.}
    \label{fig:privacy_results}
    \vspace{-3mm}
\end{figure}

\setlength{\columnsep}{12pt}
\begin{wrapfigure}{r}{0.48\linewidth}
    \centering
    \vspace{-4.3mm}
    \includegraphics[width=0.98\linewidth]{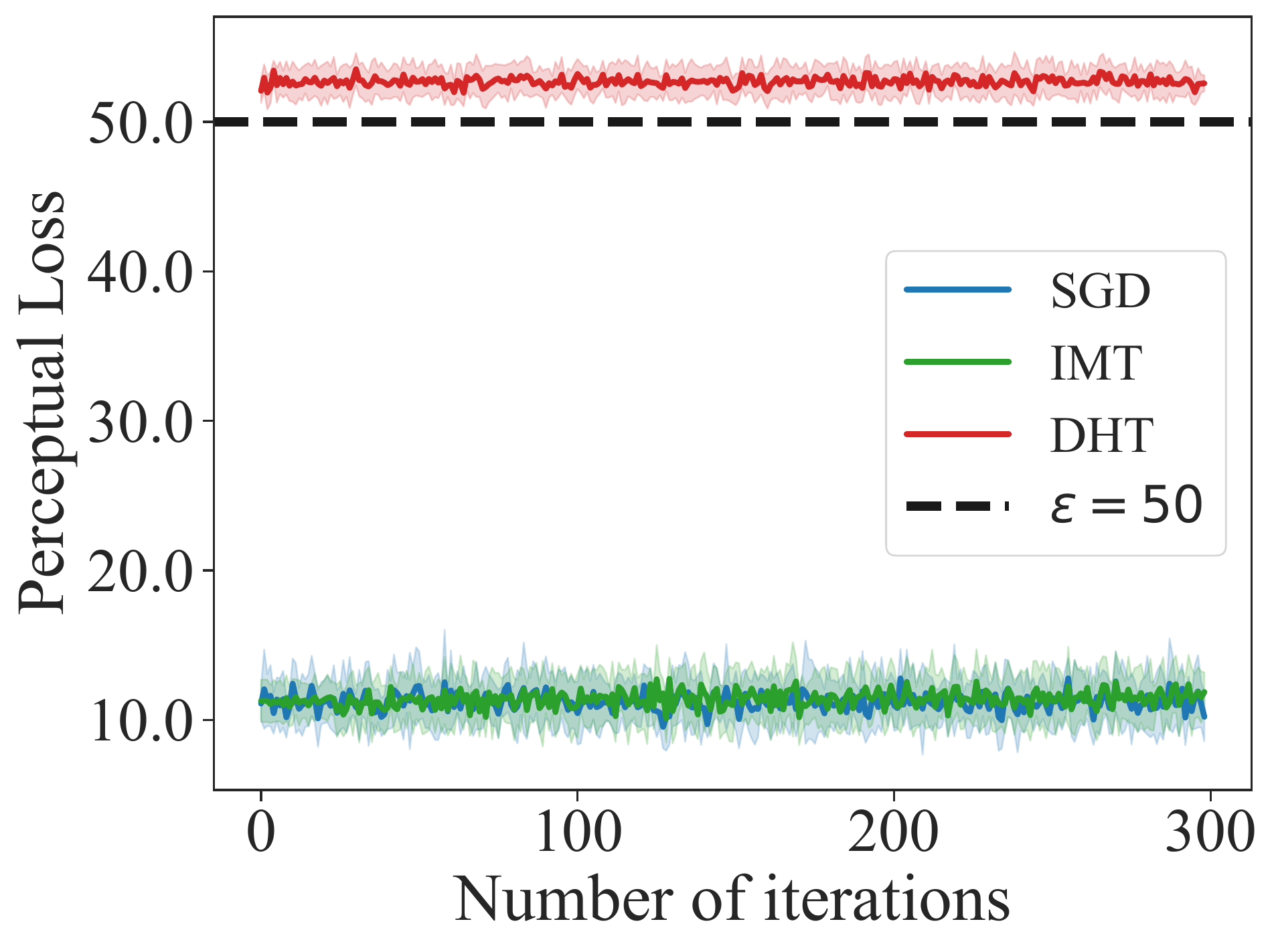}
    \vspace{-6.5mm}
    \caption{\footnotesize Private perceptual distance of the synthesized samples during teaching. $\epsilon$ is a prescribed distance threshold.}
    \vspace{-2.2mm}
    \label{fig:privacy}
\end{wrapfigure}

In practical applications, generating samples that are semantically distinct from the original data distribution could be beneficial. For example, in the medical domain, we wish not to reveal sensitive information that might contain in the original dataset. As a proof of concept, we regard privacy preservation as some distance constraints on the feature space, \ie, the generated samples are at least $\epsilon$-away (in a semantic latent space) from a pre-defined privacy set. The teaching objective is 
\vspace{-1.8mm}
\begin{equation}
\footnotesize
\begin{aligned}
& \min_{\bm{\theta}}\norm{ \bm{w}^v(\bm{\theta}) - \bm{w}^* }^2_2 + \sum_{t=1}^{v} \ell (\bm{\pi}_{\bm{\theta}}, \bm{y}) | \bm{w}^t) \\[-0.35mm]
&~~~~~~~~~~~~~~~~~~+ \max\{0, \epsilon - \| \phi(\bm{\pi}_{\bm{\theta}}) - \phi(\bm{x})\|^2_2 \} \\
& ~~\text{s.t.}~ \norm{  \phi(\bm{\pi}_{\bm{\theta}}) - \phi(\bm{x})}^2_2 \geq\epsilon
\end{aligned}
\end{equation}
 where $\phi(\cdot)$ is a pre-trained neural network for computing perceptual distance. Here, we demonstrate that it is possible to achieve similar teaching performance by only synthesizing samples that satisfy the privacy constraints (see Figure~\ref{fig:privacy_results}). We also show the private perceptual distance during the teaching in Figure~\ref{fig:privacy} in which the private perceptual distance is defined as the minimal distance between the samples in the privacy set and the synthesized sample.                                       

\vspace{-2.8mm}
\section{Concluding Remarks}
\vspace{-2mm}

In this paper, we introduce a novel data hallucination teaching framework and demonstrate, both theoretically and empirically, that DHT achieves promising teaching performance in both omniscient and black-box settings. We also highlight that a novel performative teaching formulation is proposed for teaching black-box neural learners. Experiments show that DHT is able to achieve significant performance gains when teaching black-box neural learners.

\newpage

\section*{Acknowledgements}
\vspace{-1.8mm}

Zeju Qiu and Weiyang Liu contributed equally to this work. This work was done when Zeju Qiu did a master thesis with Weiyang Liu at MPI. This work is supported by the German Federal Ministry of Education and Research (BMBF): Tübingen AI Center, FKZ: 01IS18039A, 01IS18039B; and by the Machine Learning Cluster of Excellence, EXC number 2064/1 – Project number 390727645. AW acknowledges support from a Turing AI Fellowship under EPSRC grant EP/V025279/1, The Alan Turing Institute, and the Leverhulme Trust via CFI.

\bibliography{ref} 

\begin{thebibliography}{10}

\bibitem{zhu2015machine}
X.~Zhu, ``Machine teaching: An inverse problem to machine learning and an
  approach toward optimal education.,'' in {\em AAAI}, 2015.

\bibitem{zhu2018overview}
X.~Zhu, A.~Singla, S.~Zilles, and A.~N. Rafferty, ``An overview of machine
  teaching,'' {\em arXiv preprint arXiv:1801.05927}, 2018.

\bibitem{singla2014near}
A.~Singla, I.~Bogunovic, G.~Bartok, A.~Karbasi, and A.~Krause, ``Near-optimally
  teaching the crowd to classify.,'' in {\em ICML}, 2014.

\bibitem{singla2013actively}
A.~Singla, I.~Bogunovic, G.~Bart{\'o}k, A.~Karbasi, and A.~Krause, ``On
  actively teaching the crowd to classify,'' in {\em NeurIPS Workshop on Data
  Driven Education}, 2013.

\bibitem{zhou2018unlearn}
Y.~Zhou, A.~R. Nelakurthi, and J.~He, ``Unlearn what you have learned: Adaptive
  crowd teaching with exponentially decayed memory learners,'' in {\em KDD},
  2018.

\bibitem{zhou2020crowd}
Y.~Zhou, A.~R. Nelakurthi, R.~Maciejewski, W.~Fan, and J.~He, ``Crowd teaching
  with imperfect labels,'' in {\em WWW}, 2020.

\bibitem{alfeld2016data}
S.~Alfeld, X.~Zhu, and P.~Barford, ``Data poisoning attacks against
  autoregressive models.,'' in {\em AAAI}, 2016.

\bibitem{alfeld2017explicit}
S.~Alfeld, X.~Zhu, and P.~Barford, ``Explicit defense actions against test-set
  attacks,'' in {\em AAAI}, 2017.

\bibitem{rakhsha2020policy}
A.~Rakhsha, G.~Radanovic, R.~Devidze, X.~Zhu, and A.~Singla, ``Policy teaching
  via environment poisoning: Training-time adversarial attacks against
  reinforcement learning,'' in {\em ICML}, 2020.

\bibitem{ma2019policy}
Y.~Ma, X.~Zhang, W.~Sun, and J.~Zhu, ``Policy poisoning in batch reinforcement
  learning and control,'' in {\em NeurIPS}, 2019.

\bibitem{bengio2009curriculum}
Y.~Bengio, J.~Louradour, R.~Collobert, and J.~Weston, ``Curriculum learning,''
  in {\em ICML}, 2009.

\bibitem{wang2018dataset}
T.~Wang, J.-Y. Zhu, A.~Torralba, and A.~A. Efros, ``Dataset distillation,''
  {\em arXiv preprint arXiv:1811.10959}, 2018.

\bibitem{liu2017iterative}
W.~Liu, B.~Dai, A.~Humayun, C.~Tay, C.~Yu, L.~B. Smith, J.~M. Rehg, and
  L.~Song, ``Iterative machine teaching,'' in {\em ICML}, 2017.

\bibitem{liu2021iterative}
W.~Liu, Z.~Liu, H.~Wang, L.~Paull, B.~Sch{\"o}lkopf, and A.~Weller, ``Iterative
  teaching by label synthesis,'' in {\em NeurIPS}, 2021.

\bibitem{cazenavette2022dataset}
G.~Cazenavette, T.~Wang, A.~Torralba, A.~A. Efros, and J.-Y. Zhu, ``Dataset
  distillation by matching training trajectories,'' in {\em CVPR Workshops},
  2022.

\bibitem{zhao2021dataset}
B.~Zhao, K.~R. Mopuri, and H.~Bilen, ``Dataset condensation with gradient
  matching.,'' in {\em ICLR}, 2021.

\bibitem{zhang2017mixup}
H.~Zhang, M.~Cisse, Y.~N. Dauphin, and D.~Lopez-Paz, ``mixup: Beyond empirical
  risk minimization,'' {\em arXiv preprint arXiv:1710.09412}, 2017.

\bibitem{verma2019manifold}
V.~Verma, A.~Lamb, C.~Beckham, A.~Najafi, I.~Mitliagkas, D.~Lopez-Paz, and
  Y.~Bengio, ``Manifold mixup: Better representations by interpolating hidden
  states,'' in {\em ICML}, 2019.

\bibitem{krizhevsky2017imagenet}
A.~Krizhevsky, I.~Sutskever, and G.~E. Hinton, ``Imagenet classification with
  deep convolutional neural networks,'' {\em Communications of the ACM},
  vol.~60, no.~6, pp.~84--90, 2017.

\bibitem{perez2017effectiveness}
L.~Perez and J.~Wang, ``The effectiveness of data augmentation in image
  classification using deep learning,'' {\em arXiv preprint arXiv:1712.04621},
  2017.

\bibitem{dao2019kernel}
T.~Dao, A.~Gu, A.~Ratner, V.~Smith, C.~De~Sa, and C.~R{\'e}, ``A kernel theory
  of modern data augmentation,'' in {\em ICML}, 2019.

\bibitem{chen2020simple}
T.~Chen, S.~Kornblith, M.~Norouzi, and G.~Hinton, ``A simple framework for
  contrastive learning of visual representations,'' in {\em ICML}, 2020.

\bibitem{perdomo2020performative}
J.~Perdomo, T.~Zrnic, C.~Mendler-D{\"u}nner, and M.~Hardt, ``Performative
  prediction,'' in {\em ICML}, 2020.

\bibitem{zhu2013machine}
X.~Zhu, ``Machine teaching for bayesian learners in the exponential family,''
  in {\em NeurIPS}, 2013.

\bibitem{liu2016teaching}
J.~Liu, X.~Zhu, and H.~G. Ohannessian, ``The teaching dimension of linear
  learners,'' in {\em ICML}, 2016.

\bibitem{mansouri2019preference}
F.~Mansouri, Y.~Chen, A.~Vartanian, J.~Zhu, and A.~Singla, ``Preference-based
  batch and sequential teaching: Towards a unified view of models,'' in {\em
  NeurIPS}, 2019.

\bibitem{chen2018understanding}
Y.~Chen, A.~Singla, O.~Mac~Aodha, P.~Perona, and Y.~Yue, ``Understanding the
  role of adaptivity in machine teaching: The case of version space learners,''
  in {\em NeurIPS}, 2018.

\bibitem{tabibian2019enhancing}
B.~Tabibian, U.~Upadhyay, A.~De, A.~Zarezade, B.~Sch{\"o}lkopf, and
  M.~Gomez-Rodriguez, ``Enhancing human learning via spaced repetition
  optimization,'' {\em Proceedings of the National Academy of Sciences}, 2019.

\bibitem{kumar2020teaching}
A.~Kumar, H.~Zhang, A.~Singla, and Y.~Chen, ``The teaching dimension of kernel
  perceptron,'' {\em arXiv preprint arXiv:2010.14043}, 2020.

\bibitem{zhang2020teaching}
X.~Zhang, S.~K. Bharti, Y.~Ma, A.~Singla, and X.~Zhu, ``The teaching dimension
  of q-learning,'' {\em arXiv preprint arXiv:2006.09324}, 2020.

\bibitem{wang2021teaching}
C.~Wang, A.~Singla, and Y.~Chen, ``Teaching an active learner with contrastive
  examples,'' in {\em NeurIPS}, 2021.

\bibitem{peltola2019machine}
T.~Peltola, M.~M. {\c{C}}elikok, P.~Daee, and S.~Kaski, ``Machine teaching of
  active sequential learners,'' in {\em NeurIPS}, 2019.

\bibitem{yuan2021iterative}
L.~Yuan, D.~Zhou, J.~Shen, J.~Gao, J.~L. Chen, Q.~Gu, Y.~N. Wu, and S.-C. Zhu,
  ``Iterative teacher-aware learning,'' in {\em NeurIPS}, 2021.

\bibitem{hunziker2019teaching}
A.~Hunziker, Y.~Chen, O.~Mac~Aodha, M.~G. Rodriguez, A.~Krause, P.~Perona,
  Y.~Yue, and A.~Singla, ``Teaching multiple concepts to a forgetful learner,''
  in {\em NeurIPS}, 2019.

\bibitem{liu2018blackbox}
W.~Liu, B.~Dai, X.~Li, Z.~Liu, J.~Rehg, and L.~Song, ``Towards black-box
  iterative machine teaching,'' in {\em ICML}, 2018.

\bibitem{lessard2019optimal}
L.~Lessard, X.~Zhang, and X.~Zhu, ``An optimal control approach to sequential
  machine teaching,'' in {\em AISTATS}, 2019.

\bibitem{Xu2021LST}
Z.~Xu, B.~Chen, C.~Li, W.~Liu, L.~Song, Y.~Lin, and A.~Shrivastava, ``Locality
  sensitive teaching,'' in {\em NeurIPS}, 2021.

\bibitem{tschiatschek2019learner}
S.~Tschiatschek, A.~Ghosh, L.~Haug, R.~Devidze, and A.~Singla, ``Learner-aware
  teaching: Inverse reinforcement learning with preferences and constraints,''
  in {\em NeurIPS}, 2019.

\bibitem{kamalaruban2019interactive}
P.~Kamalaruban, R.~Devidze, V.~Cevher, and A.~Singla, ``Interactive teaching
  algorithms for inverse reinforcement learning,'' {\em arXiv preprint
  arXiv:1905.11867}, 2019.

\bibitem{haug2018teaching}
L.~Haug, S.~Tschiatschek, and A.~Singla, ``Teaching inverse reinforcement
  learners via features and demonstrations,'' in {\em NeurIPS}, 2018.

\bibitem{johns2015becoming}
E.~Johns, O.~Mac~Aodha, and G.~J. Brostow, ``Becoming the expert-interactive
  multi-class machine teaching,'' in {\em CVPR}, 2015.

\bibitem{chen2018near}
Y.~Chen, O.~Mac~Aodha, S.~Su, P.~Perona, and Y.~Yue, ``Near-optimal machine
  teaching via explanatory teaching sets,'' in {\em AISTATS}, 2018.

\bibitem{mac2018teaching}
O.~Mac~Aodha, S.~Su, Y.~Chen, P.~Perona, and Y.~Yue, ``Teaching categories to
  human learners with visual explanations,'' in {\em CVPR}, 2018.

\bibitem{mei2015using}
S.~Mei and X.~Zhu, ``Using machine teaching to identify optimal training-set
  attacks on machine learners.,'' in {\em AAAI}, 2015.

\bibitem{zhang2018training}
X.~Zhang, X.~Zhu, and S.~Wright, ``Training set debugging using trusted
  items,'' in {\em AAAI}, 2018.

\bibitem{zhang2020adaptive}
X.~Zhang, Y.~Ma, A.~Singla, and X.~Zhu, ``Adaptive reward-poisoning attacks
  against reinforcement learning,'' in {\em ICML}, 2020.

\bibitem{zhang2020online}
X.~Zhang, X.~Zhu, and L.~Lessard, ``Online data poisoning attacks,'' in {\em
  L4DC}, 2020.

\bibitem{wang2020mathematical}
P.~Wang, J.~Wang, P.~Paranamana, and P.~Shafto, ``A mathematical theory of
  cooperative communication,'' in {\em NeurIPS}, 2020.

\bibitem{shafto2021cooperative}
P.~Shafto, J.~Wang, and P.~Wang, ``Cooperative communication as belief
  transport,'' {\em Trends in Cognitive Sciences}, 2021.

\bibitem{wang2020sequential}
J.~Wang, P.~Wang, and P.~Shafto, ``Sequential cooperative bayesian inference,''
  in {\em ICML}, 2020.

\bibitem{wei2019eda}
J.~Wei and K.~Zou, ``Eda: Easy data augmentation techniques for boosting
  performance on text classification tasks,'' in {\em EMNLP}, 2019.

\bibitem{park2019specaugment}
D.~S. Park, W.~Chan, Y.~Zhang, C.-C. Chiu, B.~Zoph, E.~D. Cubuk, and Q.~V. Le,
  ``Specaugment: A simple data augmentation method for automatic speech
  recognition,'' in {\em Interspeech}, 2019.

\bibitem{zhang2021understanding}
C.~Zhang, S.~Bengio, M.~Hardt, B.~Recht, and O.~Vinyals, ``Understanding deep
  learning (still) requires rethinking generalization,'' {\em Communications of
  the ACM}, vol.~64, no.~3, pp.~107--115, 2021.

\bibitem{simonyan2014very}
K.~Simonyan and A.~Zisserman, ``Very deep convolutional networks for
  large-scale image recognition,'' {\em arXiv preprint arXiv:1409.1556}, 2014.

\bibitem{he2015deep}
K.~He, X.~Zhang, S.~Ren, and J.~Sun, ``Deep residual learning for image
  recognition,'' {\em arXiv preprint arXiv:1512.03385}, 2015.

\bibitem{he2020momentum}
K.~He, H.~Fan, Y.~Wu, S.~Xie, and R.~Girshick, ``Momentum contrast for
  unsupervised visual representation learning,'' in {\em CVPR}, 2020.

\bibitem{tsang2005core}
I.~W. Tsang, J.~T. Kwok, P.-M. Cheung, and N.~Cristianini, ``Core vector
  machines: Fast svm training on very large data sets.,'' {\em Journal of
  Machine Learning Research}, vol.~6, no.~4, 2005.

\bibitem{campbell2018bayesian}
T.~Campbell and T.~Broderick, ``Bayesian coreset construction via greedy
  iterative geodesic ascent,'' in {\em ICML}, 2018.

\bibitem{sener2018active}
O.~Sener and S.~Savarese, ``Active learning for convolutional neural networks:
  A core-set approach,'' in {\em ICLR}, 2018.

\bibitem{mirzasoleiman2020coresets}
B.~Mirzasoleiman, J.~Bilmes, and J.~Leskovec, ``Coresets for data-efficient
  training of machine learning models,'' in {\em ICML}, 2020.

\bibitem{angelova2005pruning}
A.~Angelova, Y.~Abu-Mostafam, and P.~Perona, ``Pruning training sets for
  learning of object categories,'' in {\em CVPR}, 2005.

\bibitem{lapedriza2013all}
A.~Lapedriza, H.~Pirsiavash, Z.~Bylinskii, and A.~Torralba, ``Are all training
  examples equally valuable?,'' {\em arXiv preprint arXiv:1311.6510}, 2013.

\bibitem{liu2021orthogonal}
W.~Liu, R.~Lin, Z.~Liu, J.~M. Rehg, L.~Paull, L.~Xiong, L.~Song, and A.~Weller,
  ``Orthogonal over-parameterized training,'' in {\em CVPR}, 2021.

\bibitem{finn2017model}
C.~Finn, P.~Abbeel, and S.~Levine, ``Model-agnostic meta-learning for fast
  adaptation of deep networks,'' in {\em ICML}, 2017.

\bibitem{werbos1988generalization}
P.~J. Werbos, ``Generalization of backpropagation with application to a
  recurrent gas market model,'' {\em Neural networks}, 1988.

\bibitem{goodfellow2020generative}
I.~Goodfellow, J.~Pouget-Abadie, M.~Mirza, B.~Xu, D.~Warde-Farley, S.~Ozair,
  A.~Courville, and Y.~Bengio, ``Generative adversarial networks,'' {\em
  Communications of the ACM}, vol.~63, no.~11, pp.~139--144, 2020.

\bibitem{kingma2013auto}
D.~P. Kingma and M.~Welling, ``Auto-encoding variational bayes,'' {\em arXiv
  preprint arXiv:1312.6114}, 2013.

\bibitem{hinton2015distilling}
G.~Hinton, O.~Vinyals, and J.~Dean, ``Distilling the knowledge in a neural
  network,'' {\em arXiv preprint arXiv:1503.02531}, 2015.

\bibitem{zoph2016neural}
B.~Zoph and Q.~V. Le, ``Neural architecture search with reinforcement
  learning,'' {\em arXiv preprint arXiv:1611.01578}, 2016.

\bibitem{liu2018progressive}
C.~Liu, B.~Zoph, M.~Neumann, J.~Shlens, W.~Hua, L.-J. Li, L.~Fei-Fei,
  A.~Yuille, J.~Huang, and K.~Murphy, ``Progressive neural architecture
  search,'' in {\em ECCV}, 2018.

\bibitem{liu2018darts}
H.~Liu, K.~Simonyan, and Y.~Yang, ``Darts: Differentiable architecture
  search,'' {\em arXiv preprint arXiv:1806.09055}, 2018.

\bibitem{andrychowicz2016learning}
M.~Andrychowicz, M.~Denil, S.~G. Colmenarejo, M.~W. Hoffman, D.~Pfau,
  T.~Schaul, B.~Shillingford, and N.~de~Freitas, ``Learning to learn by
  gradient descent by gradient descent,'' in {\em NeurIPS}, 2016.

\bibitem{cubuk2019autoaugment}
E.~D. Cubuk, B.~Zoph, D.~Mane, V.~Vasudevan, and Q.~V. Le, ``Autoaugment:
  Learning augmentation strategies from data,'' in {\em CVPR}, 2019.

\bibitem{williams1992simple}
R.~J. Williams, ``Simple statistical gradient-following algorithms for
  connectionist reinforcement learning,'' {\em Machine learning}, 1992.

\bibitem{healy2015performativity}
K.~Healy, ``The performativity of networks,'' {\em European Journal of
  Sociology/Archives Europ{\'e}ennes de Sociologie}, vol.~56, no.~2,
  pp.~175--205, 2015.

\bibitem{mackenzie2007economists}
D.~A. MacKenzie, F.~Muniesa, L.~Siu, {\em et~al.}, {\em Do economists make
  markets?: on the performativity of economics}.
\newblock Princeton University Press, 2007.

\bibitem{hardt2022performative}
M.~Hardt, M.~Jagadeesan, and C.~Mendler-D{\"u}nner, ``Performative power,''
  {\em arXiv preprint arXiv:2203.17232}, 2022.

\bibitem{liu2017sphereface}
W.~Liu, Y.~Wen, Z.~Yu, M.~Li, B.~Raj, and L.~Song, ``Sphereface: Deep
  hypersphere embedding for face recognition,'' in {\em CVPR}, 2017.

\bibitem{wang2018cosface}
H.~Wang, Y.~Wang, Z.~Zhou, X.~Ji, D.~Gong, J.~Zhou, Z.~Li, and W.~Liu,
  ``Cosface: Large margin cosine loss for deep face recognition,'' in {\em
  CVPR}, 2018.

\bibitem{deng2019arcface}
J.~Deng, J.~Guo, N.~Xue, and S.~Zafeiriou, ``Arcface: Additive angular margin
  loss for deep face recognition,'' in {\em CVPR}, 2019.

\bibitem{liu2018learning}
W.~Liu, R.~Lin, Z.~Liu, L.~Liu, Z.~Yu, B.~Dai, and L.~Song, ``Learning towards
  minimum hyperspherical energy,'' in {\em NeurIPS}, 2018.

\bibitem{liu2018decoupled}
W.~Liu, Z.~Liu, Z.~Yu, B.~Dai, R.~Lin, Y.~Wang, J.~M. Rehg, and L.~Song,
  ``Decoupled networks,'' in {\em CVPR}, 2018.

\bibitem{chen2020angular}
B.~Chen, W.~Liu, Z.~Yu, J.~Kautz, A.~Shrivastava, A.~Garg, and A.~Anandkumar,
  ``Angular visual hardness,'' in {\em ICML}, 2020.

\bibitem{zhang2019lookahead}
M.~Zhang, J.~Lucas, J.~Ba, and G.~E. Hinton, ``Lookahead optimizer: k steps
  forward, 1 step back,'' in {\em NeurIPS}, 2019.

\bibitem{lin2003some}
G.-H. Lin and M.~Fukushima, ``Some exact penalty results for nonlinear programs
  and mathematical programs with equilibrium constraints,'' {\em Journal of
  Optimization Theory and Applications}, 2003.

\bibitem{vaswani2019painless}
S.~Vaswani, A.~Mishkin, I.~Laradji, M.~Schmidt, G.~Gidel, and
  S.~Lacoste-Julien, ``Painless stochastic gradient: Interpolation,
  line-search, and convergence rates,'' in {\em NeurIPS}, 2019.

\end{thebibliography}
\bibliographystyle{ieeetr}

\newpage
\onecolumn
\appendix
\addcontentsline{toc}{section}{Appendix} 
\renewcommand \thepart{} 
\renewcommand \partname{}
\part{\Large\centerline{Appendix}}

\parttoc 

\newpage

\section{Proof of Theorem~\ref{thm:ET_DHT}}

From the $(t+1)$-th gradient update with the greedy DHT teacher, we have that
\begin{equation}
\begin{aligned}
\norm{\bm{w}^{t+1}-\bm{w}^*}^2&=\norm{\bm{w}^t-\eta_t\nabla_{\bm{w}^t}\ell(\tilde{\bm{x}}_{i_t},y_{i_t}|\bm{w}^t)-\bm{w}^*}^2\\
&=\norm{\bm{w}^t-\eta_t   g(\tilde{\bm{x}}_{i_t})\cdot\mathcal{T}_{\bm{x}\rightarrow\tilde{\bm{x}}}\circ\nabla_{\bm{w}^t}\ell(\bm{x}_{i_t},{y}_{i_t}|\bm{w}^t) -\bm{w}^*}^2\\
&=\bigg{\|}{\bm{w}^t-\eta_t   \underbrace{g(\tilde{\bm{x}}_{i_t})\cdot\beta}_{\stackrel{\text{def}}{=}g_s(\tilde{\bm{x}}_{i_t})}\cdot\nabla_{\bm{w}^t}\ell(\bm{x}_{i_t},{y}_{i_t}|\bm{w}^t) -\bm{w}^*}\bigg{\|}^2
\end{aligned}
\end{equation}
where $\tilde{\bm{x}}$ is the data generated by DHT and $i_t$ denotes a randomly sampled index from the pool in the $t$-th iteration. Because $\mathcal{T}_{\bm{x}\rightarrow\tilde{\bm{x}}}$ is a scaling mapping, we have that $g_s(\tilde{\bm{x}})$ is generally defined as
\begin{equation}
    g_s(\tilde{\bm{x}})=\beta\cdot g(\tilde{\bm{x}})=\beta \cdot\frac{\norm{\nabla_{\bm{w}}\ell(\bm{x},{y}|\bm{w})}}{\norm{\nabla_{\bm{w}}\ell(\tilde{\bm{x}},{y}|\bm{w})}}\cdot\frac{\norm{\tilde{\bm{x}}}}{\norm{{\bm{x}}}},
\end{equation}
which, for different linear learners, can be instantiated as
\begin{equation}
\begin{aligned}
    \text{LSR learner:}~~g_s(\tilde{\bm{x}})&=\frac{\beta\langle\bm{w},\bm{x}\rangle -\beta\cdot y}{\beta\langle\bm{w},\bm{x}\rangle -y}\\
    \text{LR learner:}~~g_s(\tilde{\bm{x}})&=\frac{\beta+\beta\cdot\exp(\beta\cdot{y}\langle\bm{w},{\bm{x}}\rangle)}{1+\exp(y\langle\bm{w},\bm{x}\rangle)}
\end{aligned}
\end{equation}
which can be controlled by adjusting the value of $\beta$. $\beta$ can be dependent on $\bm{w}$, so it can be different in different iterations. Intuitively, since we can adjust $\beta$ to equivalently adjust the learning rate, we can therefore provably have a better convergence rate. Concretely, we have that
\begin{equation}
\begin{aligned}
\norm{\bm{w}^{t+1}-\bm{w}^*}^2&=\norm{\bm{w}^t-\bm{w}^*}^2-2\eta_t g_s(\tilde{\bm{x}}_{i_t})\langle \nabla_{\bm{w}^t}\ell(\bm{x}_{i_t},{y}_{i_t}|\bm{w}^t), \bm{w}^t-\bm{w}^* \rangle \\
&\ \ \ \ \ \ \ \ \ + \eta_t^2 (g_s(\tilde{\bm{x}}_{i_t}))^2\norm{\nabla_{\bm{w}^t}\ell(\bm{x}_{i_t},{y}_{i_t}|\bm{w}^t)}^2
\end{aligned}
\end{equation}
which can be simplified as (by denoting $\nabla_{\bm{w}^t}\ell(\bm{x}_{i_t},{y}_{i_t}|\bm{w}^t)$ as $\nabla\ell_{i_t}(\bm{w}^t)$):
\begin{equation}\label{thm1_ineq1}
\begin{aligned}
\norm{\bm{w}^{t+1}-\bm{w}^*}^2=\norm{\bm{w}^t-\bm{w}^*}^2-2\eta_t g_s(\tilde{\bm{x}}_{i_t})\langle \nabla\ell_{i_t}(\bm{w}^t), \bm{w}^t-\bm{w}^* \rangle + \eta_t^2 (g_s(\tilde{\bm{x}}_{i_t}))^2\norm{\nabla\ell_{i_t}(\bm{w}^t)}^2.
\end{aligned}
\end{equation}
Because we know that the synthesized data $\bm{x}_{i_t}$ generated by the greedy DHT policy is the solution to the following minimization:
\begin{equation}
\tilde{\bm{x}}_{i_t}=\arg\min_{\bm{x}'_{i_t}}\bigg{\{} \eta_t^2 (g_s(\bm{x}'_{i_t}))^2\norm{\nabla_{\bm{w}^t}\ell(\bm{x}_{i_t},{y}_{i_t}|\bm{w}^t)}^2-2\eta_t g_s(\bm{x}'_{i_t})\langle \nabla_{\bm{w}^t}\ell(\bm{x}_{i_t},{y}_{i_t}|\bm{w}^t), \bm{w}^t-\bm{w}^* \rangle\bigg{\}},
\end{equation}
then we plug a new $\tilde{\bm{x}}''_{i_t}$ which satisfies $g_s(\tilde{\bm{x}}''_{i_t})=c_1\norm{\bm{w}^{t}-\bm{w}^*}$ to Eq.~\eqref{thm1_ineq1} and have the following inequality:
\begin{equation}
\begin{aligned}
\norm{\bm{w}^{t+1}-\bm{w}^*}^2&\leq\norm{\bm{w}^t-\bm{w}^*}^2-2\eta_t g_s(\tilde{\bm{x}}''_{i_t})\langle \nabla\ell_{i_t}(\bm{w}^t), \bm{w}^t-\bm{w}^* \rangle + \eta_t^2 (g_s(\tilde{\bm{x}}''_{i_t}))^2\norm{\nabla\ell_{i_t}(\bm{w}^t)}^2\\
&=\norm{\bm{w}^t-\bm{w}^*}^2-2\eta_t c_1\norm{\bm{w}^{t}-\bm{w}^*}\langle \nabla\ell_{i_t}(\bm{w}^t), \bm{w}^t-\bm{w}^* \rangle \\
&\ \ \ \ \ \ \ \ \ \ \ + \eta_t^2 c_1^2\norm{\nabla\ell_{i_t}(\bm{w}^t)}^2\norm{\bm{w}^{t}-\bm{w}^*}^2
\end{aligned}
\end{equation}
which holds because $\tilde{\bm{x}}_{i_t}$ leads to the minimal $\norm{\bm{w}^{t+1}-\bm{w}^*}^2$ and $\tilde{\bm{x}}''_{i_t}$ has to result in a larger or equal $\norm{\bm{w}^{t+1}-\bm{w}^*}^2$.

Next, we apply the convexity of $f(\cdot)$ and the order-1 strong convexity~\cite{lin2003some} of $\ell_{i_t}(\cdot)$, and therefore have that (let $\mu_{i_t}=0$ when $\ell_{i_t}$ is not order-1 strongly convex):
\begin{equation}
\begin{aligned}
-\langle \nabla\ell_{i_t}(\bm{w}^t), \bm{w}^t-\bm{w}^* \rangle\leq \ell_{i_t}(\bm{w}^*)-\ell_{i_t}(\bm{w}^t)-\frac{\mu_{i_t}}{2}\norm{\bm{w}^t-\bm{w}^*}
\end{aligned}
\end{equation}
which results in
\begin{equation*}
\begin{aligned}
\norm{\bm{w}^{t+1}-\bm{w}^*}^2&\leq\norm{\bm{w}^t-\bm{w}^*}^2 + 2\eta_t c_1\norm{\bm{w}^{t}-\bm{w}^*}\big(\ell_{i_t}(\bm{w}^*)-\ell_{i_t}(\bm{w}^t)-\frac{\mu_{i_t}}{2}\norm{\bm{w}^t-\bm{w}^*}\big) \\
&\ \ \ \ \ \ \ \ \ + c_1^2\norm{\nabla\ell_{i_t}(\bm{w}^t)}^2\norm{\bm{w}^{t}-\bm{w}^*}^2\\
&=\norm{\bm{w}^t-\bm{w}^*}^2 + 2\eta_t c_1\norm{\bm{w}^{t}-\bm{w}^*}(\ell_{i_t}(\bm{w}^*)-\ell_{i_t}(\bm{w}^t))-\eta_t c_1\mu_{i_t}\norm{\bm{w}^t-\bm{w}^*}^2\\
&\ \ \ \ \ \ \ \ \ + c_1^2\norm{\nabla\ell_{i_t}(\bm{w}^t)}^2\norm{\bm{w}^{t}-\bm{w}^*}^2.
\end{aligned}
\end{equation*}
Considering the condition that $\ell_i$ is $L_i$-Lipschitz continuous and denoting $L_{\max}=\max_i L_i$, we then have
\begin{equation*}
\begin{aligned}
\norm{\bm{w}^{t+1}-\bm{w}^*}^2 & \leq\norm{\bm{w}^t-\bm{w}^*}^2 + 2\eta_t c_1\norm{\bm{w}^{t}-\bm{w}^*}(\ell_{i_t}(\bm{w}^*)\\
&\ \ \ \ \ \ \ \ \ -\ell_{i_t}(\bm{w}^t))-\eta_t c_1\mu_{i_t}\norm{\bm{w}^t-\bm{w}^*}^2 + \eta_t^2 c_1^2L_{\max}^2\norm{\bm{w}^t-\bm{w}^*}^2.
\end{aligned}
\end{equation*}
The interpolation condition~\cite{vaswani2019painless} indicates that $\bm{w}^*$ is the minimum for all functions $\ell_i$, which is equivalent to $\ell_i(\bm{w}^*)\leq\ell_i(\bm{w}^{t})$ for all $i$. Therefore, we have that $(\ell_{i_t}(\bm{w}^*)-\ell_{i_t}(\bm{w}^t))\leq0$. Finally we arrive at
\begin{equation}
\begin{aligned}
\norm{\bm{w}^{t+1}-\bm{w}^*}^2&\leq\norm{\bm{w}^t-\bm{w}^*}^2 -\eta_t c_1\mu_{i_t}\norm{\bm{w}^t-\bm{w}^*}^2 + \eta_t^2 c_1^2L_{\max}^2\norm{\bm{w}^t-\bm{w}^*}^2\\
&=(1-\mu_{i_t}\eta_t c_1+\eta_t^2L_{\max}^2c_1^2)\norm{\bm{w}^t-\bm{w}^*}^2.
\end{aligned}
\end{equation}
Taking expectation \emph{w.r.t.} $i_t$, we obtain that
\begin{equation}
\begin{aligned}
\mathbb{E}\{\norm{\bm{w}^{t+1}-\bm{w}^*}^2\}&\leq\mathbb{E}_{i_t}\{(1-\mu_{i_t}\eta_t c_1+\eta_t^2L_{\max}^2c_1^2)\norm{\bm{w}^t-\bm{w}^*}^2\}\\
&=(1-\mathbb{E}_{i_t}\{\mu_{i_t}\}\eta_t c_1+\eta_t^2L_{\max}^2c_1^2)\norm{\bm{w}^t-\bm{w}^*}^2\\
&=(1-\bar{\mu}\eta_t c_1+\eta_t^2L_{\max}^2c_1^2)\norm{\bm{w}^t-\bm{w}^*}^2.
\end{aligned}
\end{equation}
Using recursion, we have that
\begin{equation}
\begin{aligned}
\mathbb{E}\{\norm{\bm{w}^{T}-\bm{w}^*}^2\}\leq(1-\bar{\mu}\eta_t c_1+\eta_t^2c_1^2L_{\max}^2)^T\norm{\bm{w}^0-\bm{w}^*}^2
\end{aligned}
\end{equation}
where we usually make $\eta_tc_1$ a constant such that $(1-\bar{\mu}\eta_t c_1+\eta_t^2c_1^2L_{\max}^2)$  also becomes a constant between $0$ and $1$. This is equivalent to the statement in the theorem that at most $\lceil (\log\frac{1}{1-c_1\eta_t\bar{\mu}+\eta_t^2c_1^2 L_{\max}})^{-1}\log(\frac{1}{\epsilon}\|{\bm{w}^0-\bm{w}^*}\|^2) \rceil$ iterations are needed to achieve the $\epsilon$-approximation, namely $\mathbb{E}\{\|\bm{w}^T-\bm{w}^*\|^2\}\leq \epsilon$. The proof is concluded.  \hfill $\square$

\newpage
\section{Experimental Details}

\textbf{Experiments on MNIST.} For teaching logistic regression learners, we do not use the original MNIST dataset but use a fixed projection matrix $\bm{P}$ ($\bm{P} \in \mathbb{R}^{784 \times 24}$) to downscale the flattened 784-dimensional MNIST image data to a 24-dimensional feature vector. Experiments are performed on the 24D feature vectors. For visualization, we un-project the 24D feature vectors to the original image shape using the pseudo-inverse matrix $\bm{P^{+}}$ ($\bm{P^{+}} \in \mathbb{R}^{24 \times 784}$). For teaching a logistic regression learner, we use 1100 (1000/100) images from class 3 and 5. For teaching neural learners with a performative teaching policy, we use the full MNIST dataset without any data augmentation.

\textbf{Experiments on Half-moon.} For half-moon, we use the built-in function from scikit-learn to generate 1000 (800/200) sample points with a Gaussian noise of 0.2.

\textbf{Performative Teaching.} The training schedule has been elaborated on in the main paper. The employed network architectures are described in Table~\ref{tab:architecture}. The MLP used for MNIST training has two layers (input dimension - 128 - output dimension). There are in total three hyperparameters: we denote the number of update on $\bm{w}$ to obtain $\bm{w}^*$ as $n_w$, the number of feature update to obtain $\bm{\tilde{x}}^i_j$ as $m$ and the $\epsilon$-neighbourhood. In our experiments, we use a $m$ of 15, $n_w$ of 5 and $\epsilon$ of 0.1.

\begin{table}[h!]
    \centering
    \begin{tabular}{|c|c|c|c|c|} 
      \hline
      Layer           & CNN-3                   & CNN-6                   & CNN-9                   & CNN-15 \\ 
      \hline
      Conv1.x         & $[3\times3,64]\times1$  & $[3\times3,64]\times2$  & $[3\times3,64]\times3$  & $[3\times3,64]\times5$ \\ 
      \hline
      Pool1           & \multicolumn{4}{c|}{$2\times2$, Max Pooling, Stride $2$} \\ 
      \hline
      Conv2.x         & $[3\times3,128]\times1$ & $[3\times3,128]\times2$ & $[3\times3,128]\times3$ & $[3\times3,128]\times5$ \\ 
      \hline
      Pool2           & \multicolumn{4}{c|}{$2\times2$, Max Pooling, Stride $2$} \\ 
      \hline
      Conv3.x         & $[3\times3,256]\times1$ & $[3\times3,256]\times2$ & $[3\times3,256]\times3$ & $[3\times3,256]\times5$ \\  
      \hline
      Pool3           & \multicolumn{4}{c|}{$2\times2$, Max Pooling, Stride $2$} \\  
      \hline
      Fully Connected & 256                    & 256                    & 256                    & 256 \\ 
      \hline
    \end{tabular}
    \caption{Network architecture specification of the used CNN models. Note, we use the same CNN network architecture definition as~\cite{liu2018learning} with one additional CNN-3 following the same principle.  $[3\times3,64]\times2$ denotes 2 cascaded 2D convolution layers with 64 3$\times$3 filters.}
    \label{tab:architecture}
\end{table}

\textbf{Mixup-based Teaching (Unrolling).} The training schedule has been elaborated on in the main paper. The teaching objective of mixup-based teaching is inspired by~\cite{liu2018darts}: 
\begin{align*}
& \underset{\bm{\theta}}{\text{min}} \mathcal{L}_{val}(\bm{w}^*(\bm{\theta}), \bm{\theta}) \\
& \text{s.t.} \quad \bm{w}^*(\bm{\theta}) = \argmin_{\bm{w}} \mathcal{L}_{train}(\bm{w}, \bm{\theta})
\end{align*}
with $\bm{w}$ as the student weight and $\bm{\theta}$ as the teacher weight. Following~\cite{liu2018darts}, we also perform first-order and second-order optimization. The result reported in the main paper is obtained by using second-order optimization.

\vspace{-5mm}
\begin{minipage}[t]{0.495\textwidth}
\begin{align*}
& \nabla_{\bm{\theta}} \mathcal{L}_{val}(\bm{w}, \bm{\theta}) \\
& \nabla_{\bm{w}} \mathcal{L}_{train} (\bm{w}, \bm{\theta})
\end{align*}
\end{minipage}%
\begin{minipage}[t]{0.495\textwidth}
\begin{align*}
& \nabla_{\bm{\theta}} \mathcal{L}_{val}(\bm{w} - \xi \nabla_{\bm{w}} \mathcal{L}_{train} (\bm{w}, \bm{\theta}), \bm{\theta}) \\
& \nabla_{\bm{w}} \mathcal{L}_{train} (\bm{w}, \bm{\theta})
\end{align*}
\end{minipage}
\vspace{5mm}

\textbf{Mixup-based Teaching (Policy Gradient).} We use a simple MLP with two layers (input dimension (3) - 128 - output dimension (3)) and dropout as policy network. We perform one step of optimization after running two epochs of training.
\begin{itemize}
    \item \textbf{Reward Signal}: $\ell (\bm{x}_{val}, \bm{y}_{val}|\bm{w}^v)$
    \item \textbf{Reward}: $R = r_T (s_{1...T}, a_{1...T})$
    \item \textbf{Objective Function}: $J(\bm{\theta}) = \mathbb{E}_{\pi_{\bm{\theta}}} \{ R \}$
    \item \textbf{Teacher Update Function}: $\nabla_{\bm{\theta}} J(\bm{\theta}) = \sum_{t}^{} \nabla_{\bm{\theta}} \text{log} \pi_{\bm{\theta}} (a_t | r_t) R$
    \item \textbf{State}: model features (obtained through query student model) consists of current iteration, average training loss and best validation loss. The best validation loss is updated every 100 iterations.
    \item \textbf{Action}: discretise the $\lambda$ into discrete action space $[0, 0.5, 1.0]$
\end{itemize}

\newpage
\section{Additional Experimental Results}

The objective of DHT is to reduce the distance between the student model $\bm{w}^t$ with some desired $\bm{w}^*$. Therefore, we only show the convergence of the weight difference in the main paper. $\bm{w}^*$ is obtained by using a model with the same initial weight as the student model and training it until convergence. We use a learning rate of $0.001$ and a standard SGD optimizer with no momentum and weight decay to optimize the logistic regression learner. Since the $\bm{w}^*$ is associated with good classification accuracy on the test dataset, we further show the convergence of the test accuracy.

\subsection{Teaching Logistic Regression on Half-moon Data with Greedy DHT}

We observe similar behavior between the weight convergence and the accuracy convergence of the examined methods using a greedy teaching policy on half-moon data. We use an Adam optimizer with a step size of 0.02, factor for average gradient of 0.8 ($\beta_1$), factor for average squared gradient of 0.999 ($\beta_2$), and update each model for 300 iterations. We perform early stopping when the loss converges. We constrain the sample value to be within the maximum and minimum values of the original dataset. We use the same optimizer (expect a lower learning rate of 0.001) to optimize the label. We also tested different magnitudes, but the teaching effectiveness degenerates with a smaller magnitude. Note, the original LAST framework~\cite{liu2021iterative} does not impose any constraint to achieve good teaching results.

\begin{figure}[h!]
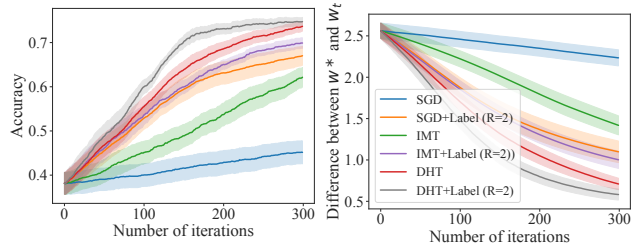

    \centering
    \begin{subfigure}{.245\linewidth}
      \centering
      \includegraphics[width=.99\linewidth]{imgs/paper_results_acc_omniscient_optimized_moon.pdf}
    \end{subfigure}%
    \begin{subfigure}{.245\linewidth}
      \centering
      \includegraphics[width=.99\linewidth]{imgs/paper_results_w_diff_omniscient_optimized_moon.pdf}
    \end{subfigure}
    \vspace{-1.7mm}
    \caption{Omniscient teaching with or without label synthesis. Convergence comparison between our greedy teaching policy with several other baseline methods on half-moon.}
    \vspace{-2mm}
\end{figure}

\subsection{Teaching Logistic Regression on MNIST with Greedy DHT}

We observe similar behavior between the weight convergence and the accuracy convergence of the examined methods using a greedy teaching policy on MNIST data. We use an AMSGrad optimizer with a step size of 0.02, a factor for average gradient of 0.8 ($\beta_1$), a factor for average squared gradient of 0.999 ($\beta_2$) and update each model for 300 iterations. We use AMSGrad because we found it converges faster for our 24D data. We perform early stopping when the loss converges. We constrain the sample value to be within the maximum and minimum values of the original dataset. We use the same optimizer (expect a lower learning rate of 0.001) to optimize the label. We constrain the magnitude of the label to be 2.

\begin{figure}[h!]
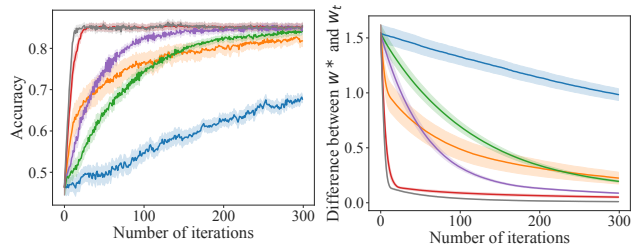

    \centering
    \begin{subfigure}{.245\linewidth}
      \centering
      \includegraphics[width=.99\linewidth]{imgs/paper_results_acc_omniscient_optimized_mnist.pdf}
    \end{subfigure}%
    \begin{subfigure}{.245\linewidth}
      \centering
      \includegraphics[width=.99\linewidth]{imgs/paper_results_w_diff_omniscient_optimized_mnist.pdf}
    \end{subfigure}
    \vspace{-1.7mm}
    \caption{Omniscient teaching with or without label synthesis. Convergence comparison between our greedy teaching policy with several other baseline methods on MNIST.}
    \vspace{-2mm}
\end{figure}

\subsection{Teaching Logistic Regression on Half-moon Data with Data Transformation}

We observe similar behavior between the weight convergence and the accuracy convergence of the examined methods using the data transformation policy on half-moon data. Generally, we notice a much steeper convergence compared to the greedy teaching policy. We optimize the teacher using an Adam optimizer with a learning rate of 0.002, a factor for average gradient of 0.9 ($\beta_1$), a factor for average squared gradient of 0.999 ($\beta_2$) and train for 1000 iterations. For each iteration, we perform 40 steps of unrolling.

\textbf{Model input.} Model input is the current student weight $\bm{w}^t$, the difference to the target model weight $\bm{w}^t - \bm{w}^*$, one random sample/label pair from the original dataset $(\bm{x}, \bm{y})$. The synthesized sample $\bm{\tilde{x}}$ is conditioned on label $\bm{y}$.

\textbf{Teacher architecture.} The teacher is a simple MLP with three layers (input dimension (8) - 32 - 16 - output dimension (2)) and ReLU activation.

\begin{figure}[h!]
    \centering
    \begin{subfigure}{0.245\linewidth}
        \centering
        \includegraphics[width=0.99\linewidth]{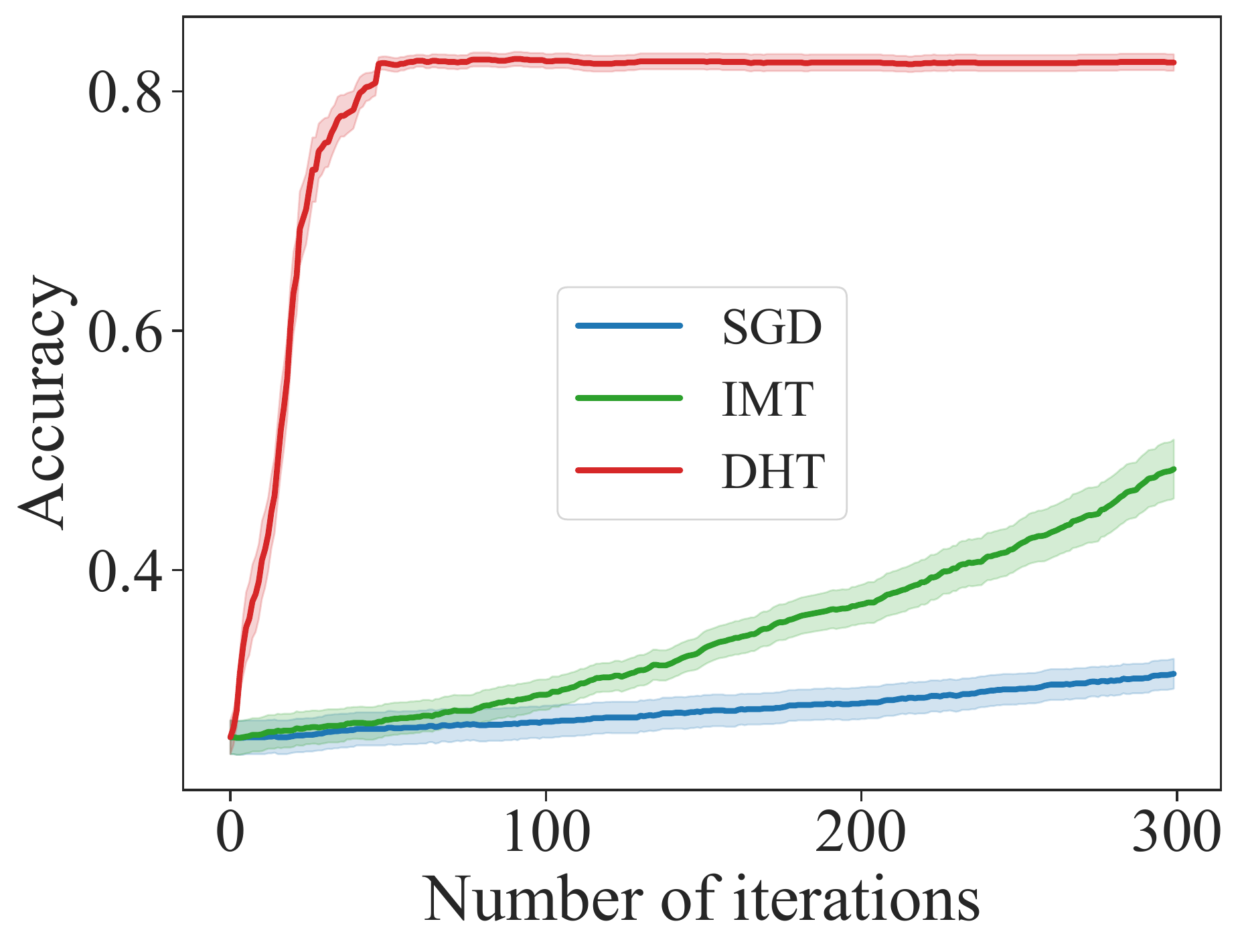}
    \end{subfigure}%
    \begin{subfigure}{0.245\linewidth}
        \centering
        \includegraphics[width=0.99\linewidth]{imgs/paper_results_w_diff_omniscient_unrolled_moon.pdf}
    \end{subfigure}
    \caption{Accuracy and weight convergence using data transformation policy in binary classification on half-moon.}
\end{figure}

\subsection{Teaching Logistic Regression on MNIST with Data Transformation}

We observe similar behavior between the weight convergence and the accuracy convergence of the examined methods using a data transformation policy on MNIST data. Generally, we notice a much steeper convergence compared to the greedy teaching policy. We optimize the teacher using an Adam optimizer with a learning rate of 0.002, a factor for average gradient of 0.9 ($\beta_1$), a factor for average squared gradient of 0.999 ($\beta_2$) and train for 1000 iterations. For each iteration, we perform 40 steps of unrolling.

\textbf{Model input.} Model input is the current student weight $\bm{w}^t$, the difference to the target model weight $\bm{w}^t - \bm{w}^*$, one random sample/label pair from the original dataset $(\bm{x}, \bm{y})$. The synthesized sample $\bm{\tilde{x}}$ is conditioned on label $\bm{y}$.

\textbf{Teacher architecture.} The teacher is a MLP with five layers (input dimension (82) - 128 - 256 - 512 - 512 - 1024 - output dimension (24)), ReLU activation and 1D batch normalization.

\begin{figure}[h!]
    \centering
    \begin{subfigure}{0.245\linewidth}
        \centering
        \includegraphics[width=0.99\linewidth]{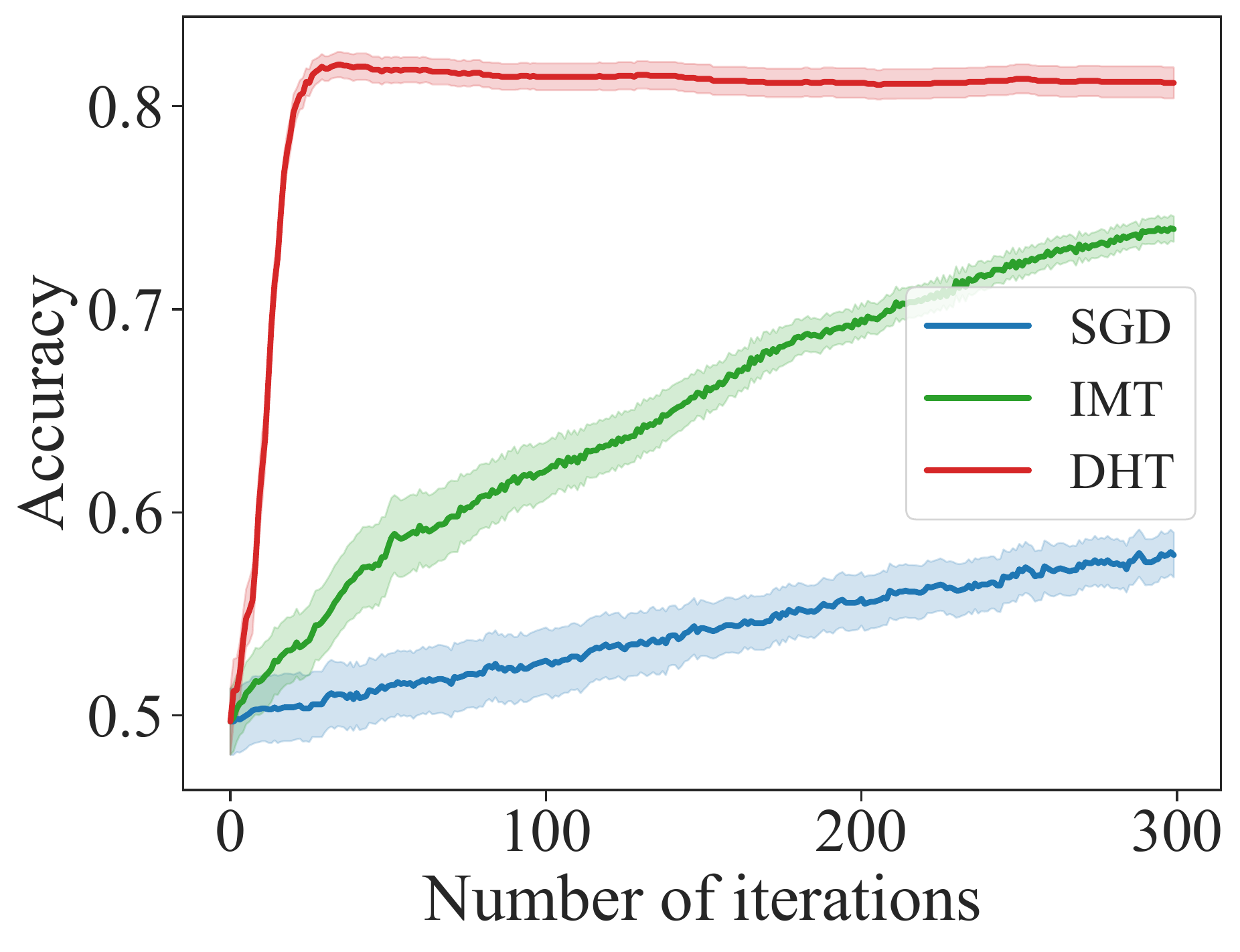}
    \end{subfigure}%
    \begin{subfigure}{0.245\linewidth}
        \centering
        \includegraphics[width=0.99\linewidth]{imgs/paper_results_w_diff_omniscient_unrolled_mnist.pdf}
    \end{subfigure}
    \caption{Accuracy and weight convergence using data transformation policy in 3/5 classification on MNIST.}
\end{figure}

\subsection{Teaching Logistic Regression on Half-moon Data with Generative Modeling} 

We observe similar behavior between the weight convergence and the accuracy convergence of the examined methods using a VAE-based generative teacher on half-moon data. In general, the performance is slightly worse than the IMT baseline but still significantly outperforms optimizing using random samples (SGD).

\textbf{Model input.} Model input for the teacher is the current student weight $\bm{w}^t$, the difference to the target model weight $\bm{w}^t - \bm{w}^*$, one random sample/label pair from the original dataset $(\bm{x}, \bm{y})$. The synthesized sample $\bm{\tilde{x}}$ is conditioned on label $\bm{y}$. The pre-trained VAE model takes one random sample/label pair $(\bm{x}, \bm{y})$ as input.

\textbf{Teacher architecture.} The VAE-based teacher utilizes a pre-trained VAE model, parametrized as a simple MLP with three layers (input dimension (4) - 128 - 256 - 128 - output dimension) as encoder and another MLP with three layers (input dimension - 128 - 256 - 128 - output dimension) as the decoder.

\begin{figure}[h!]
    \centering
    \begin{subfigure}{0.245\linewidth}
        \centering
        \includegraphics[width=0.99\linewidth]{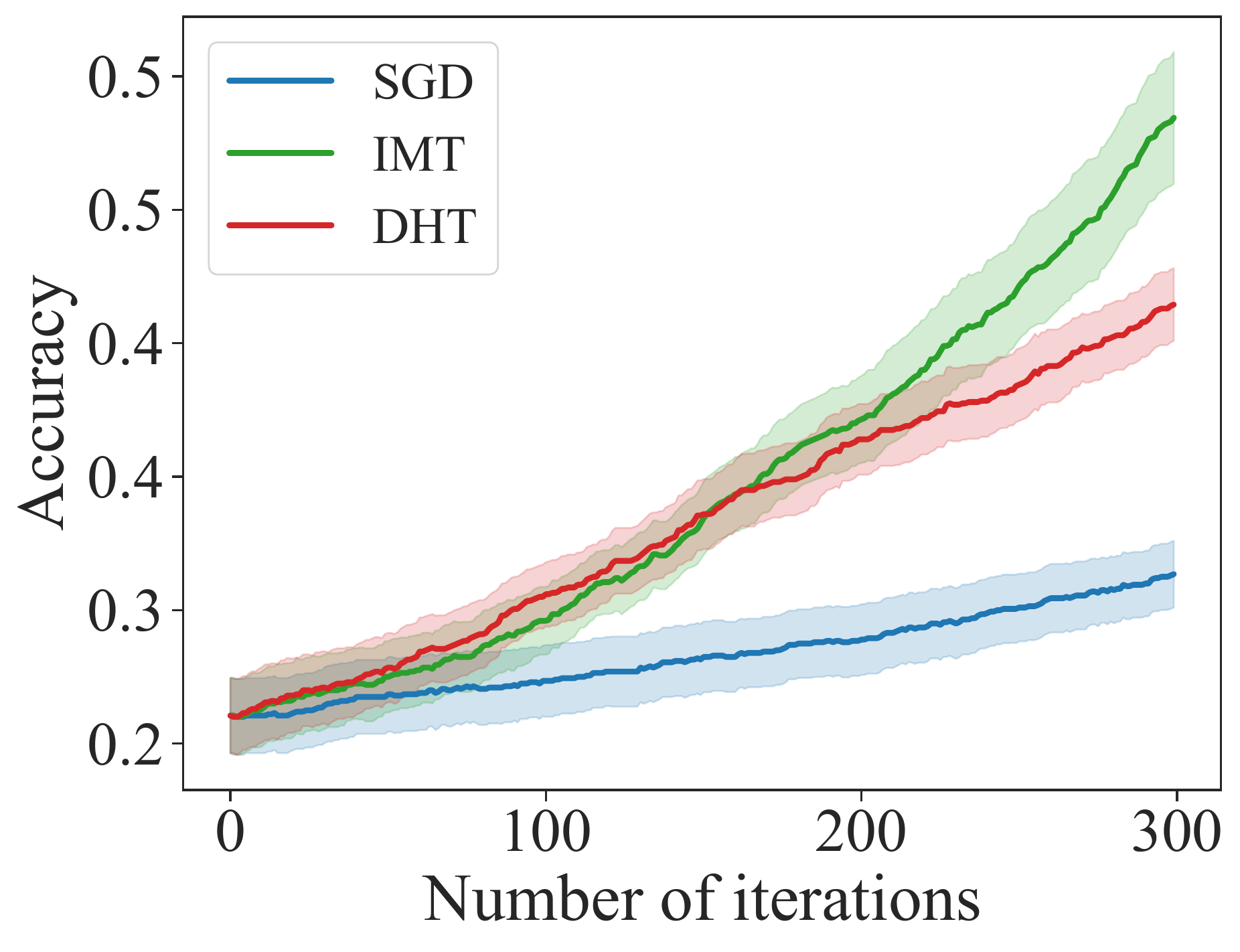}
    \end{subfigure}%
    \begin{subfigure}{0.245\linewidth}
        \centering
        \includegraphics[width=0.99\linewidth]{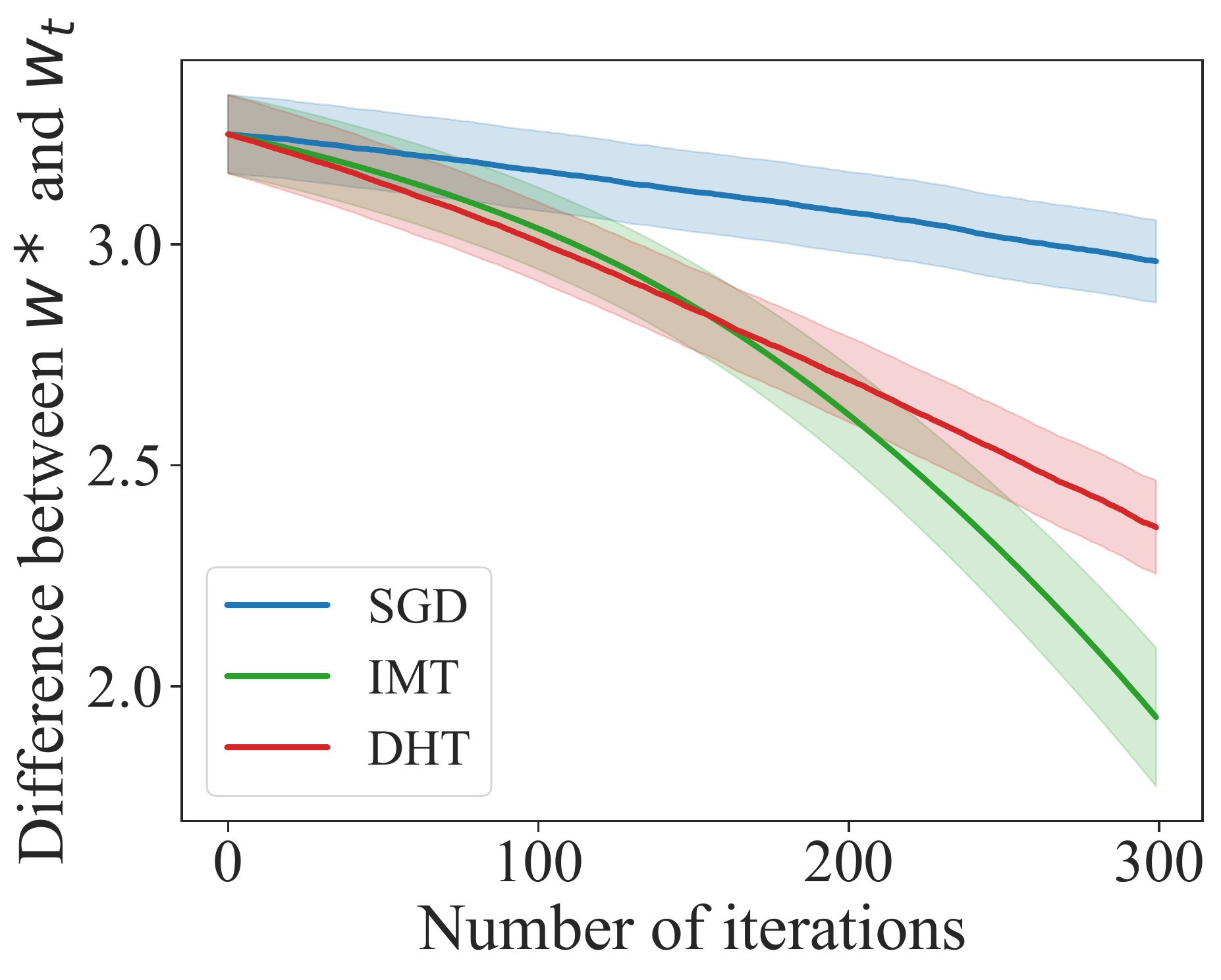}
    \end{subfigure}
    \caption{Accuracy and weight convergence using VAE-based teacher in binary classification on half-moon.}
\end{figure}

We observe similar behavior between the weight convergence and the accuracy convergence of the examined methods using a GAN-based generative teacher on half-moon data. In general, the performance is comparable with that of the IMT baseline and significantly outperforms optimizing using random samples (SGD).

\textbf{Model input.} Model input is the current student weight $\bm{w}^t$, the difference to the target model weight $\bm{w}^t - \bm{w}^*$ and one random label from the original dataset $\bm{y}$. The synthesized sample $\bm{\tilde{x}}$ is conditioned on label $\bm{y}$.

\textbf{Teacher architecture.} The GAN-based teacher is a simple MLP with three layers (input dimension (8) - 32 - 16 - output dimension (2)) and ReLU activation. The additional discriminator is a simple MLP with two layers (input dimension (4) - 8 - output dimension'), leaky ReLU (0.2), and a drop-out layer (0.3).

\begin{figure}[h!]
    \centering
    \begin{subfigure}{0.245\linewidth}
        \centering
        \includegraphics[width=0.99\linewidth]{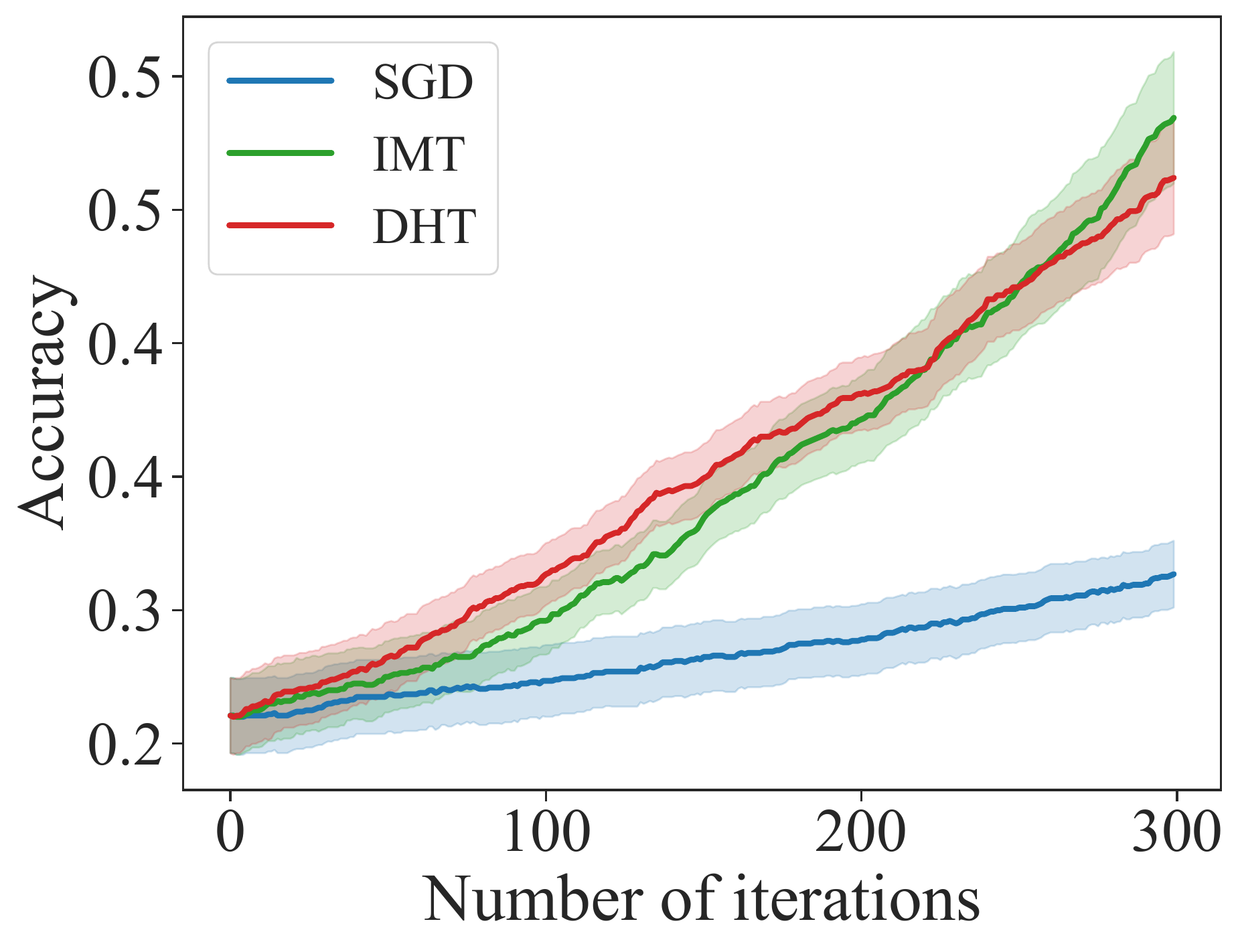}
    \end{subfigure}%
    \begin{subfigure}{0.245\linewidth}
        \centering
        \includegraphics[width=0.99\linewidth]{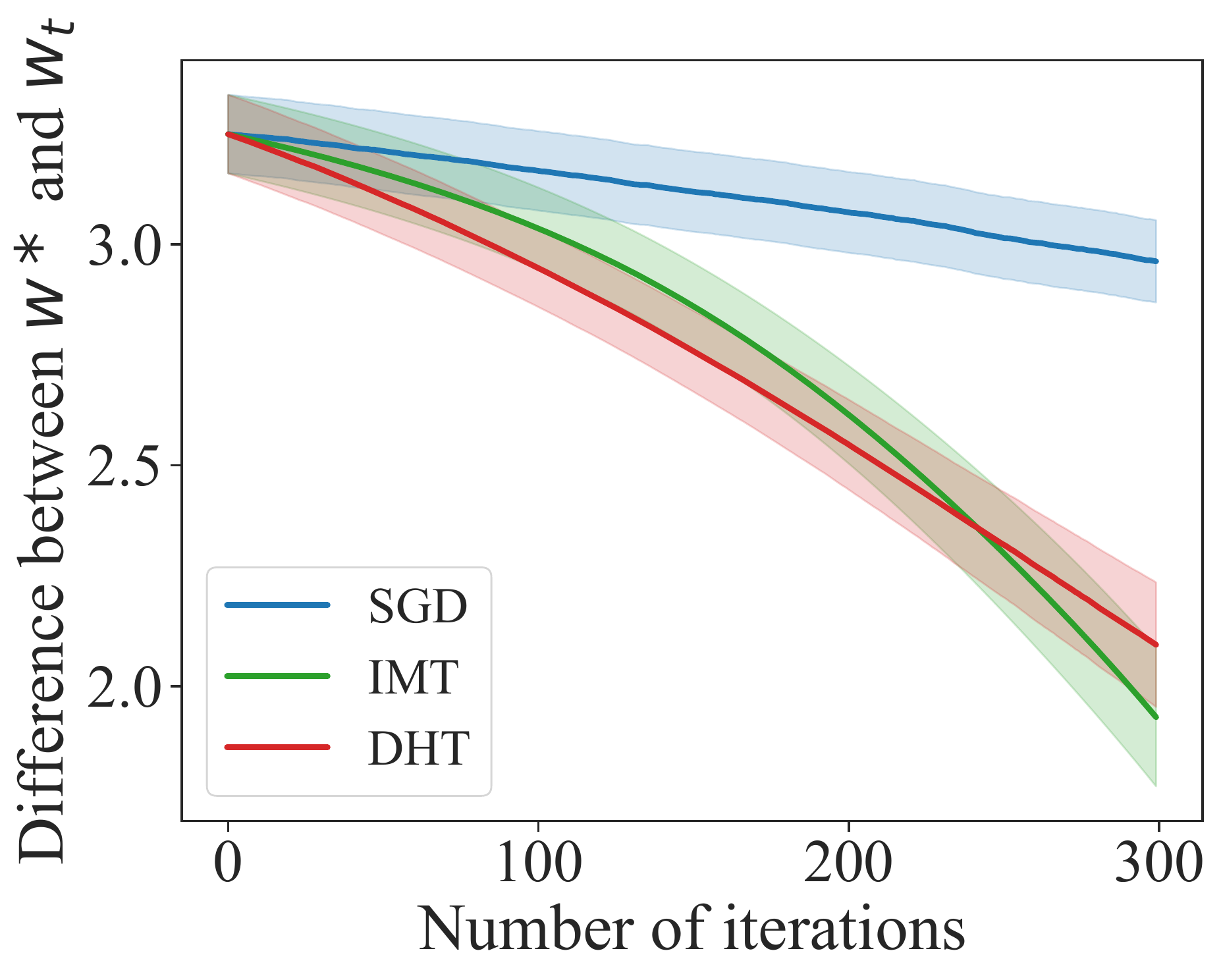}
    \end{subfigure}
    \caption{Accuracy and weight convergence using GAN-based teacher in binary classification on half-moon.}
\end{figure}

\subsection{More Qualitative Results of Generative Modeling Policy on Half-moon}

\begin{figure}[h!]
    \centering
    \includegraphics[width=0.5\linewidth]{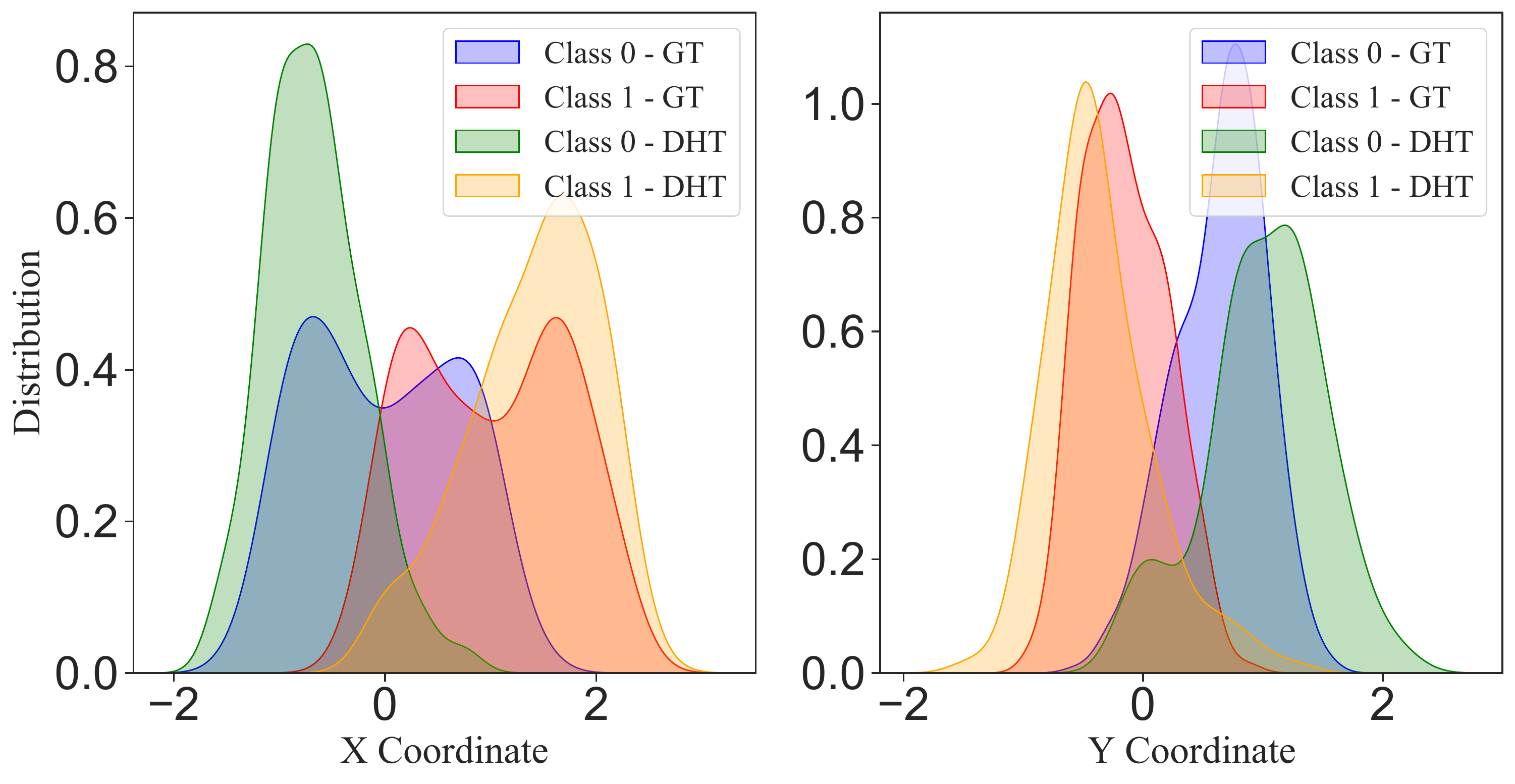}
    \caption{Visualization of the data distribution of the synthesized data by a VAE-based teacher after 300 iterations. GT means the ground truth.}
    \label{fig:qualitative_results_main0}
\end{figure}

We visualize the data distribution of the synthesized data by a VAE-based teacher after we finish teaching in Figure~\ref{fig:qualitative_results_main0}. We observe that by using a VAE-based teacher, we can generate data samples with similar distribution in the $x$- and $y$-coordinates of both classes as the original data distribution. The generated samples are more widespread and match the original data distribution better; however, sometimes, samples are synthesized outside of the original data distribution.
 
\begin{figure}[h!]
    \centering
    \includegraphics[width=0.5\linewidth]{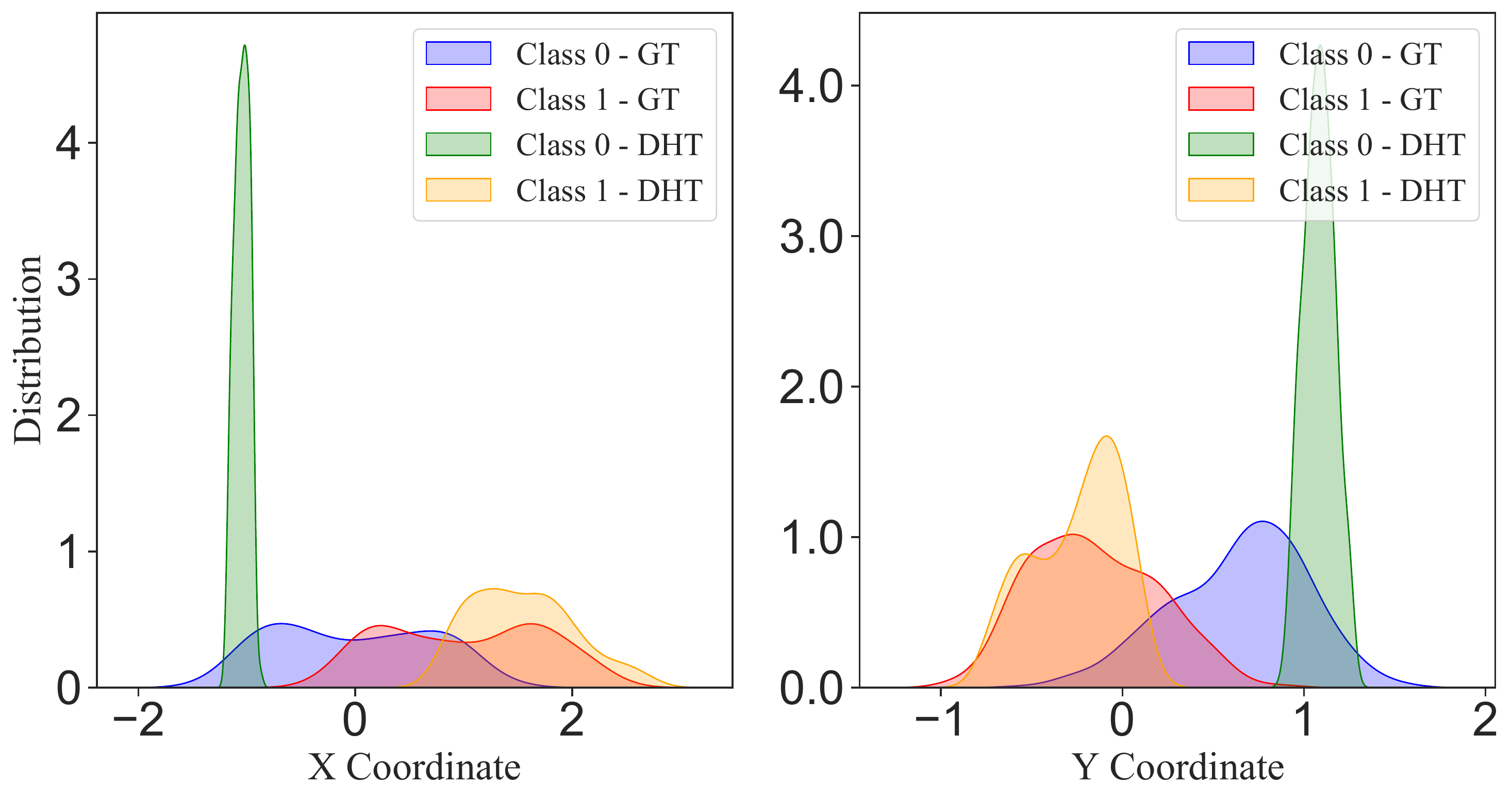}
    \caption{Visualization of the data distribution of the synthesized data by a GAN-based teacher after 300 iterations. GT denotes the ground truth.}
    \label{fig:qualitative_results_main1}
\end{figure}

We also visualize the data distribution of the synthesized data by a GAN-based teacher after we finish teaching in Figure~\ref{fig:qualitative_results_main1}. We observe that by using a GAN-based teacher, we can generate data samples with similar distribution in the $x$- and $y$-coordinates of class 1 as the original data distribution. Samples generated for class 0 are more clustered but still lie completely within the original data distribution. We do not interpret these clustered samples as an indication of mode collapse but rather view these samples as the most informative ones that can cause the model to converge faster to the desired $\bm{w}^*$. Since we also observe this kind of clustered generation when using a greedy teaching policy.

Here we visualize the synthesized samples together with the ground truth data distribution, the target classifier $\bm{w}^*$, and the student classifier $\bm{w}^t$ at different epochs. We can clearly notice the difference between the generated data samples and also its effect on the student classifier. By synthesizing more clustered data samples using a GAN-based teacher, the student classifier is able to converge faster to $\bm{w}^*$. When training a GAN-based teacher, there exists a trade-off between generating samples that will lead to faster convergence (clustered samples) and samples that are more similar to the original data distribution (spread samples). The advantage of using a VAE-based teacher is that the training is relatively stable, and the generated samples are much closer to the original dataset because we only teach in the feature space.

\begin{figure*}[h!]
    \centering
    \begin{subfigure}{0.198\linewidth}
        \centering
        \includegraphics[width=0.99\linewidth]{imgs/paper_generated_samples_moon_1_95873_10.pdf}
    \end{subfigure}%
    \begin{subfigure}{0.198\linewidth}
        \centering
        \includegraphics[width=0.99\linewidth]{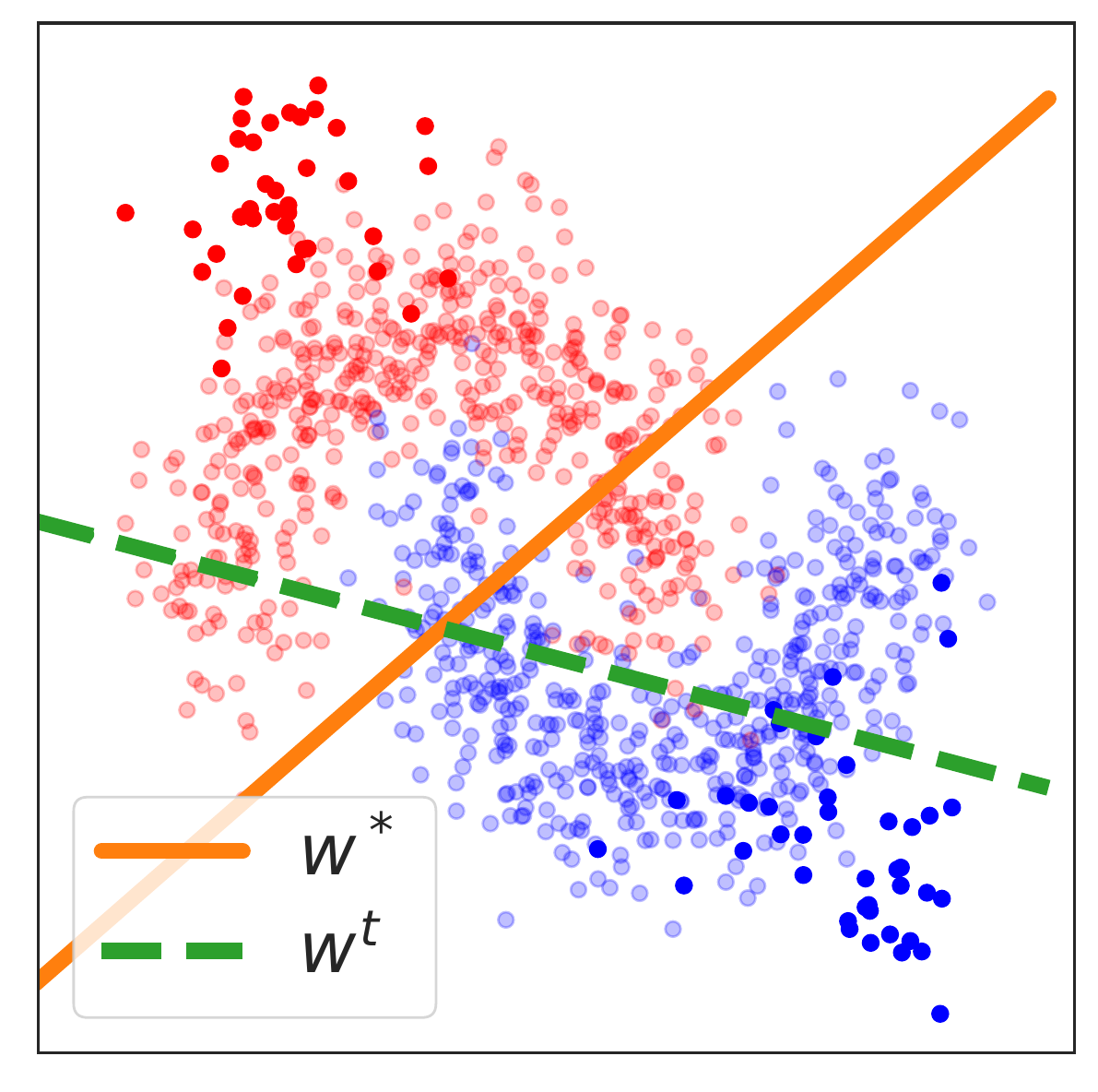}
    \end{subfigure}%
    \begin{subfigure}{0.198\linewidth}
        \centering
        \includegraphics[width=0.99\linewidth]{imgs/paper_generated_samples_moon_1_95873_150.pdf}
    \end{subfigure}%
    \begin{subfigure}{0.198\linewidth}
        \centering
        \includegraphics[width=0.99\linewidth]{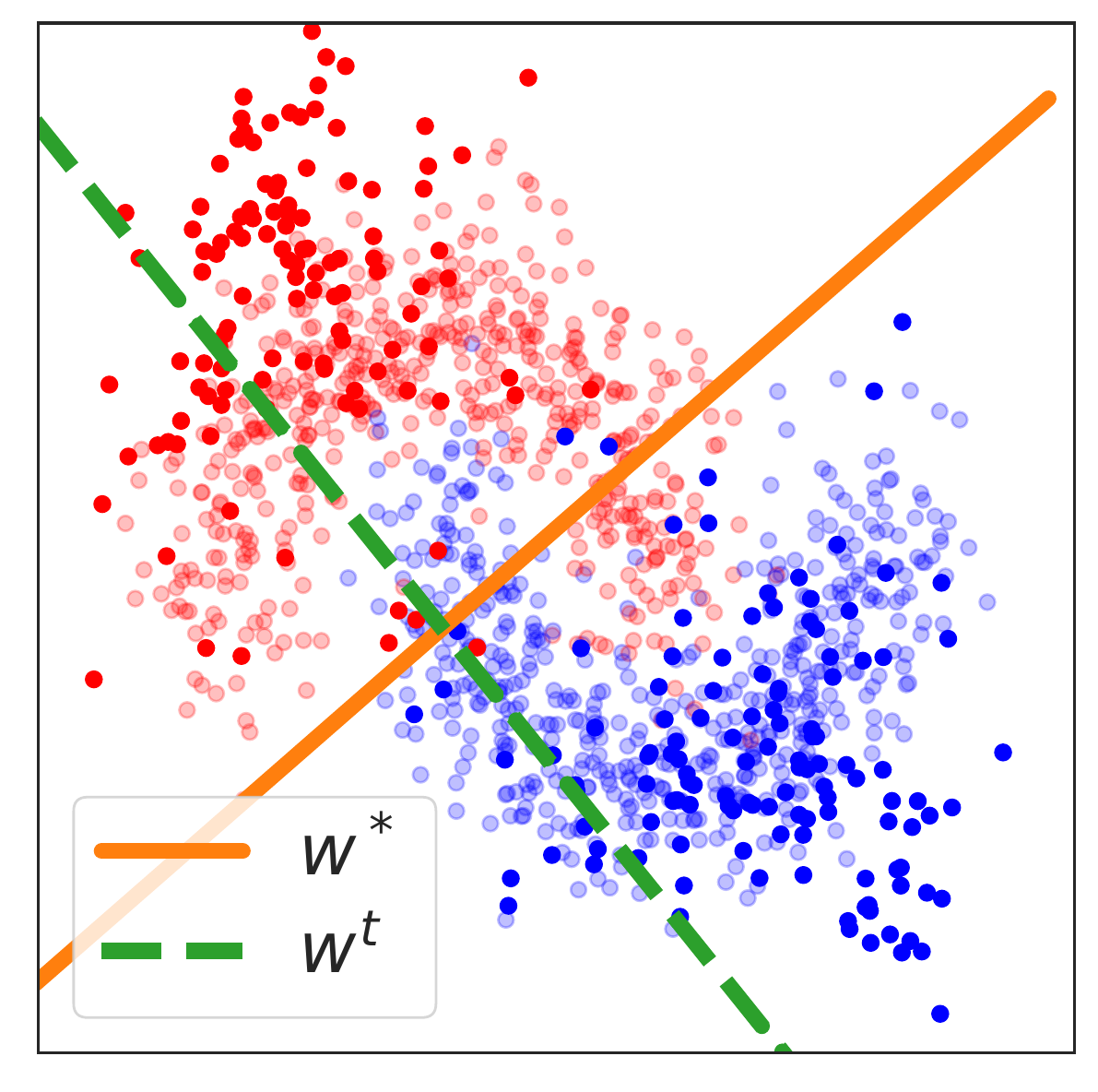}
    \end{subfigure}%
    \begin{subfigure}{0.198\linewidth}
        \centering
        \includegraphics[width=0.99\linewidth]{imgs/paper_generated_samples_moon_1_95873_290.pdf}
    \end{subfigure}
    \caption{Visualization of the data synthesized by a VAE-based teacher after iteration 10, 80, 150, 240, and 290. The orange line indicates the target classifier $w^*$; the green dashed line indicates the student classifier. Different colors indicate different classes; points with lower opacity represent the ground truth data distribution.}
    \label{fig:qualitative_results_main2}
\end{figure*}

\begin{figure}[h!]
    \centering
    \begin{subfigure}{0.198\linewidth}
        \centering
        \includegraphics[width=0.99\linewidth]{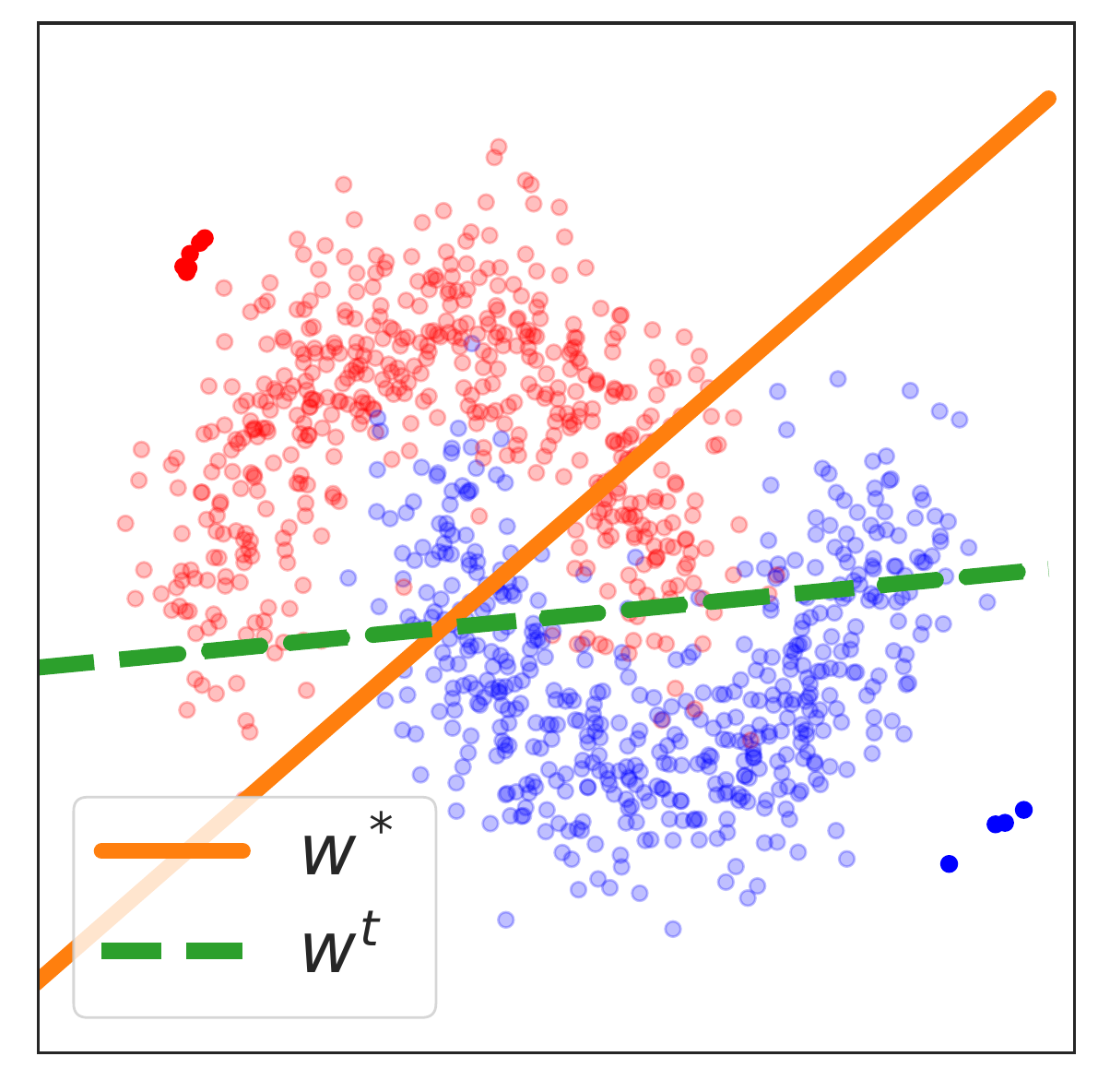}
    \end{subfigure}%
    \begin{subfigure}{0.198\linewidth}
        \centering
        \includegraphics[width=0.99\linewidth]{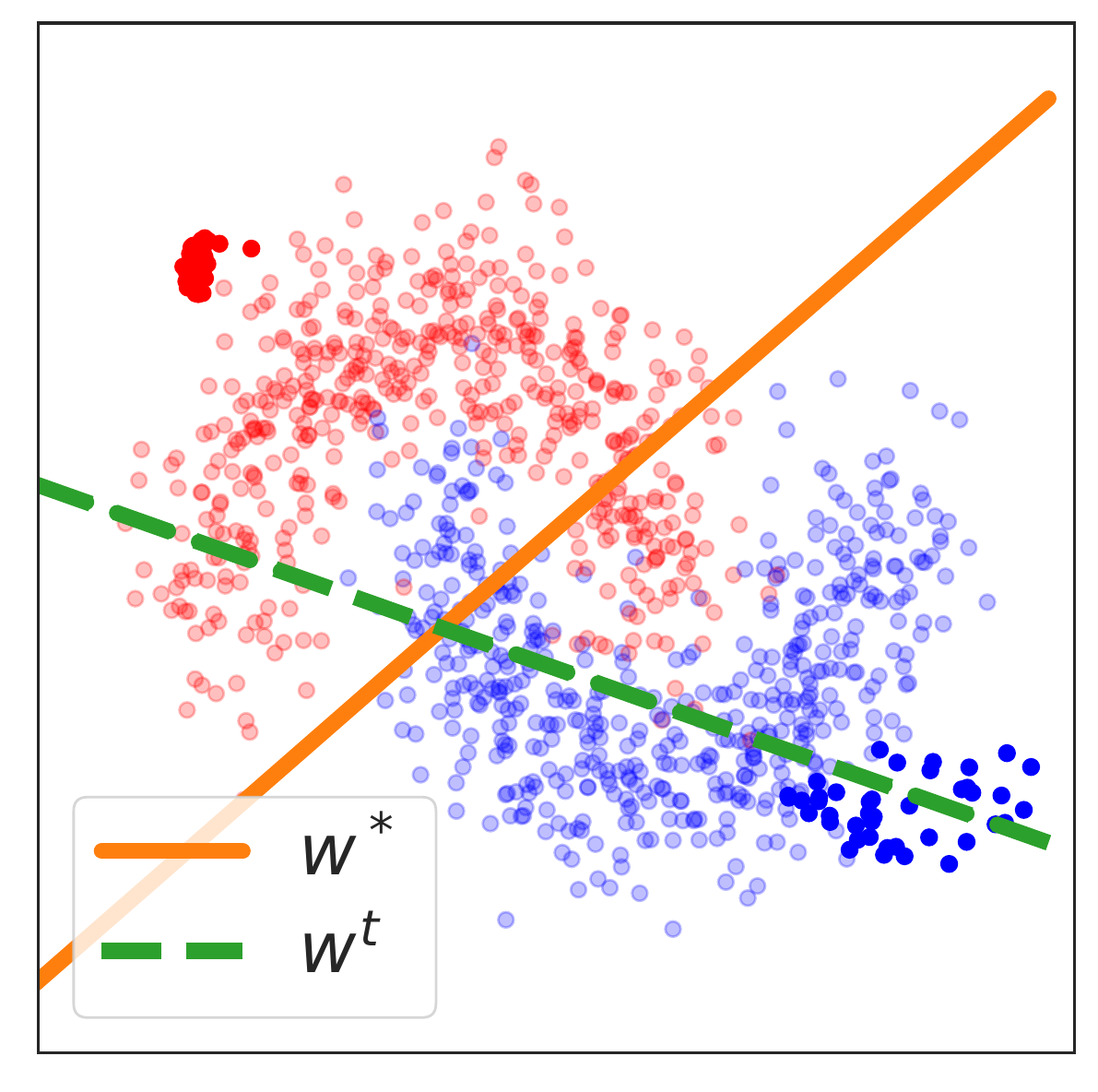}
    \end{subfigure}%
    \begin{subfigure}{0.198\linewidth}
        \centering
        \includegraphics[width=0.99\linewidth]{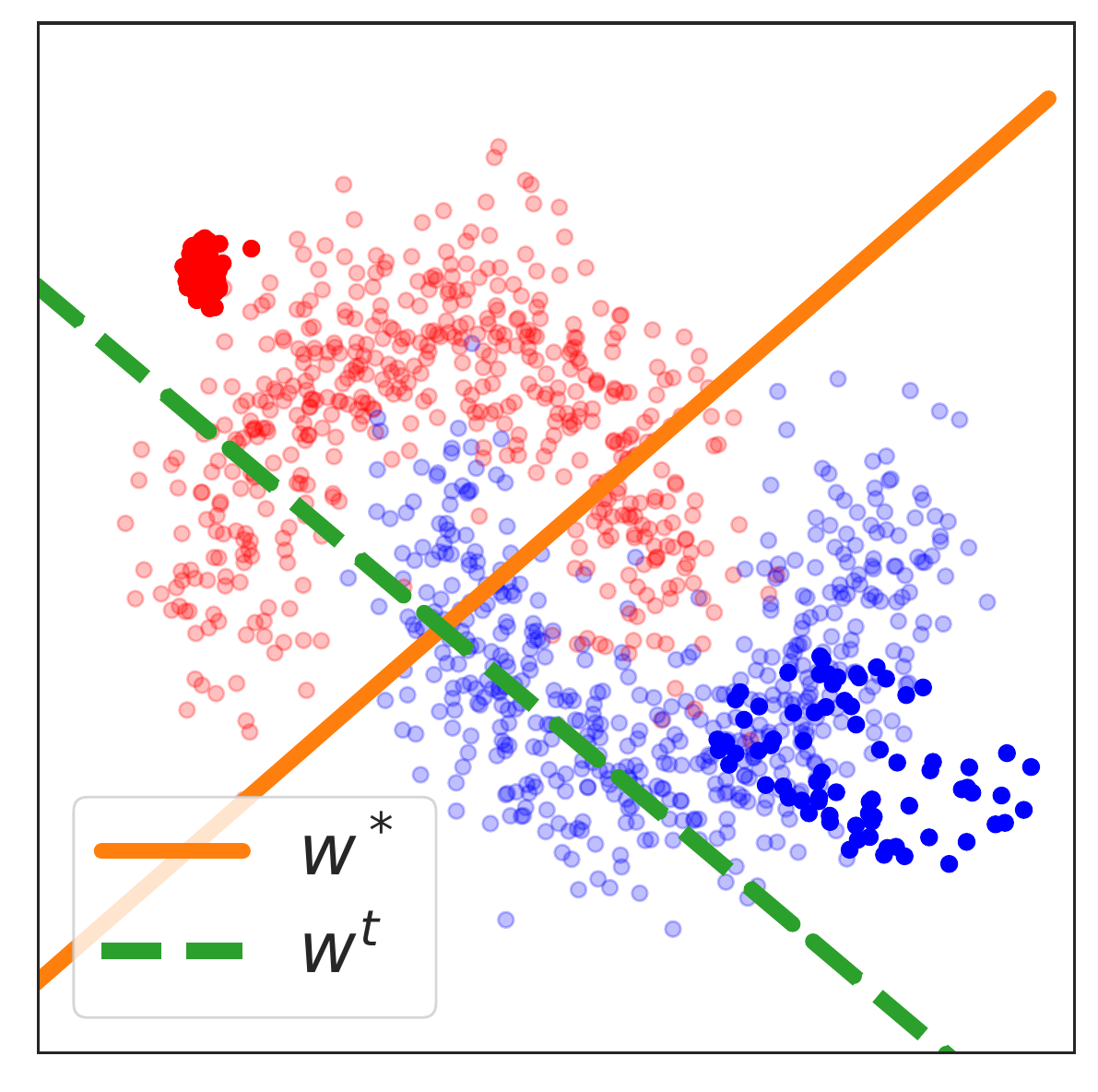}
    \end{subfigure}%
    \begin{subfigure}{0.198\linewidth}
        \centering
        \includegraphics[width=0.99\linewidth]{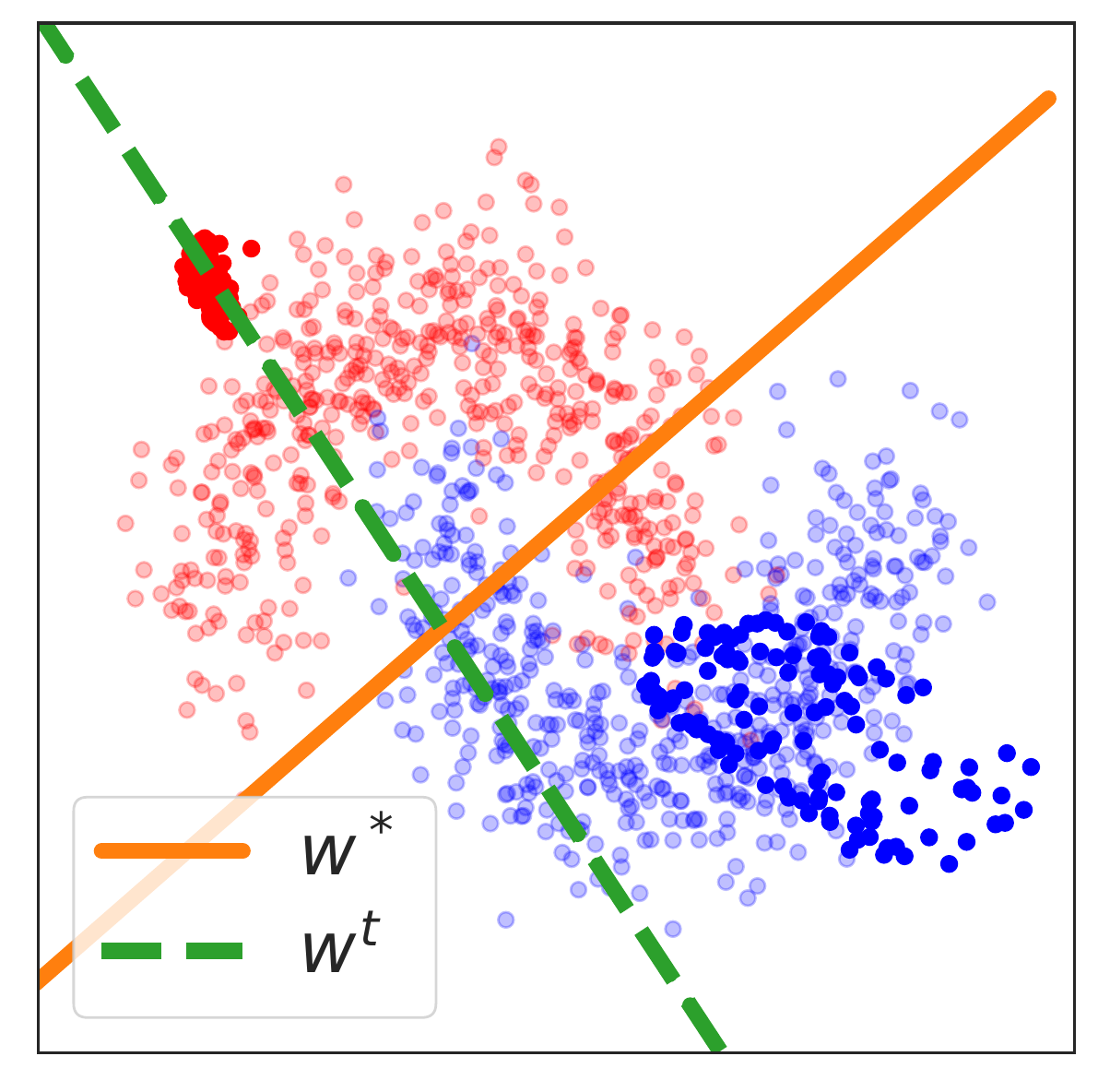}
    \end{subfigure}%
    \begin{subfigure}{0.198\linewidth}
        \centering
        \includegraphics[width=0.99\linewidth]{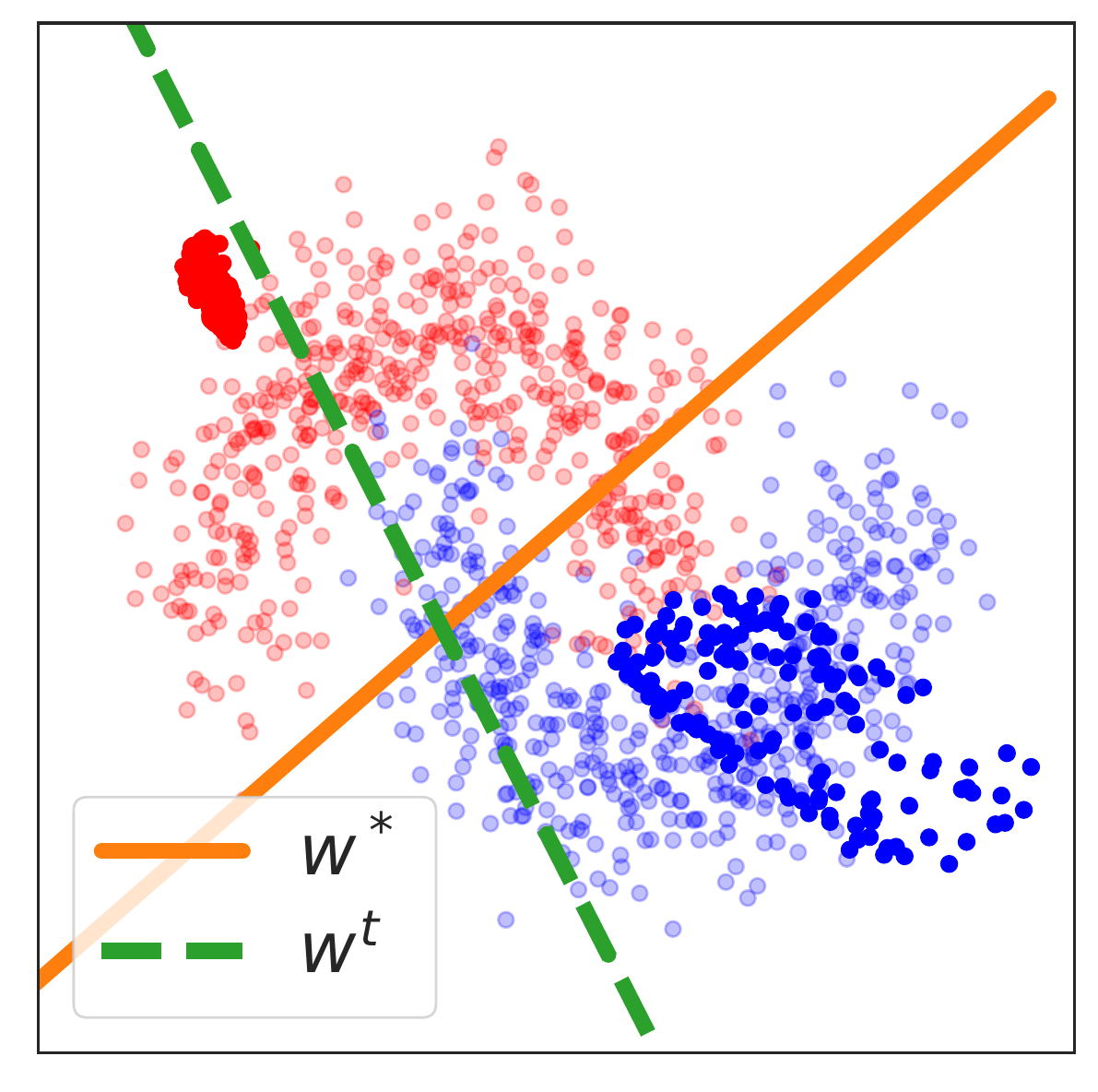}
    \end{subfigure}
    \caption{Visualization of the data synthesized by a GAN-based teacher after iteration 10, 80, 150, 240, and 290. The orange line indicates the target classifier $\bm{w}^*$; the green dashed line indicates the student classifier. Different colors indicate different classes; points with lower opacity represent the ground truth data distribution.}
\end{figure}

\subsection{Teaching Logistic Regression on MNIST with Generative Modeling}

We observe similar behavior between the weight convergence and the accuracy convergence of the examined methods using a VAE-based generative teacher on MNIST data. In general, the performance is worse than the IMT baseline but still outperforms optimizing using random samples (SGD).

\textbf{Model input.} Model input for the teacher is the current student weight $\bm{w}^t$, the difference to the target model weight $\bm{w}^t - \bm{w}^*$, one random sample/label pair from the original dataset $(\bm{x}, \bm{y})$. The synthesized sample $\bm{\tilde{x}}$ is conditioned on label $\bm{y}$. The pre-trained VAE model takes one random sample/label pair $(\bm{x}, \bm{y})$ as input.

\textbf{Teacher architecture.} The VAE-based teacher utilizes a pre-trained VAE model, with a simple CNN with two 2D convolutional layers as the encoder and another CNN with two 2D transposed convolution layers as the decoder.

\begin{figure}[h!]
    \centering
    \begin{subfigure}{0.245\linewidth}
        \centering
        \includegraphics[width=0.99\linewidth]{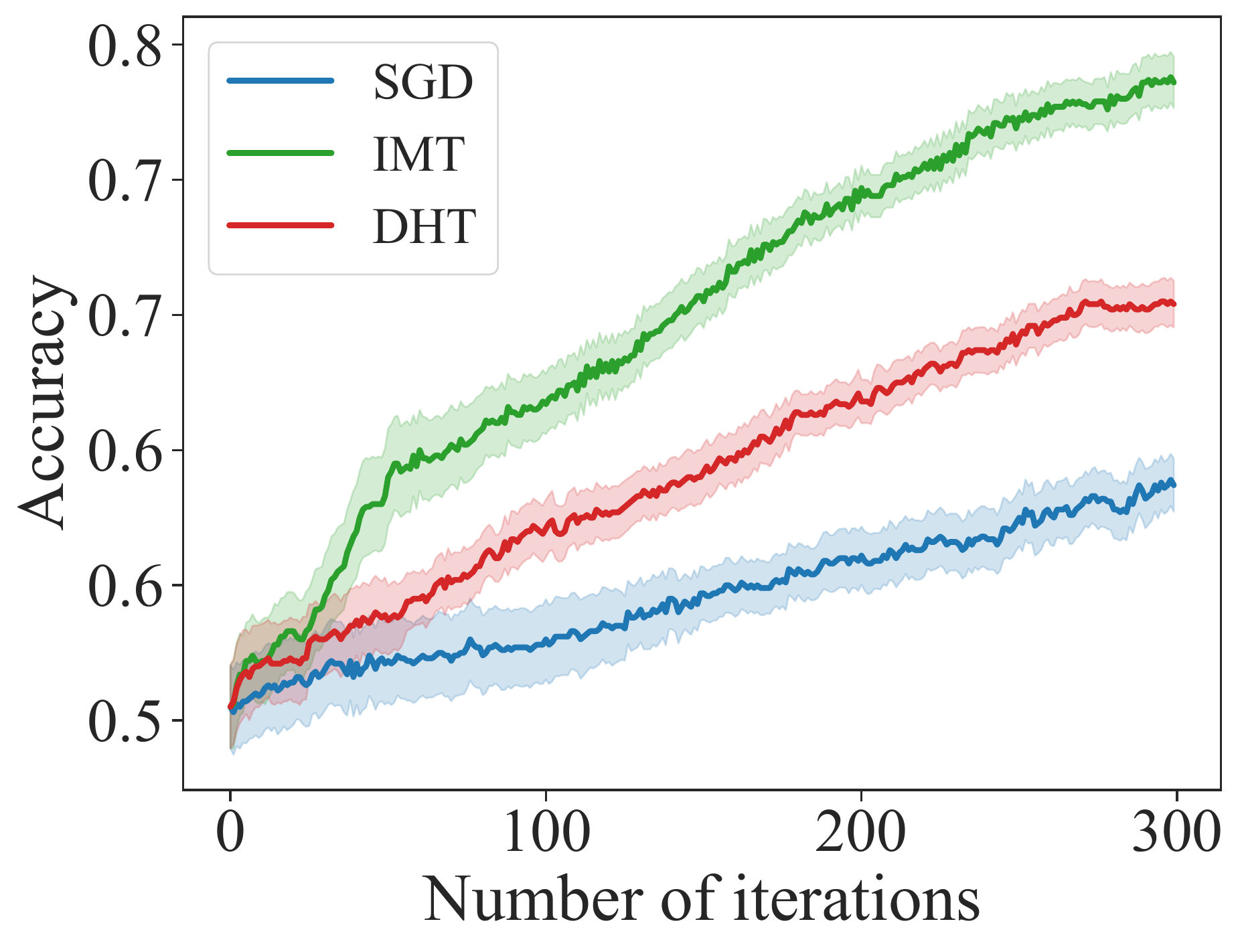}
    \end{subfigure}%
    \begin{subfigure}{0.245\linewidth}
        \centering
        \includegraphics[width=0.99\linewidth]{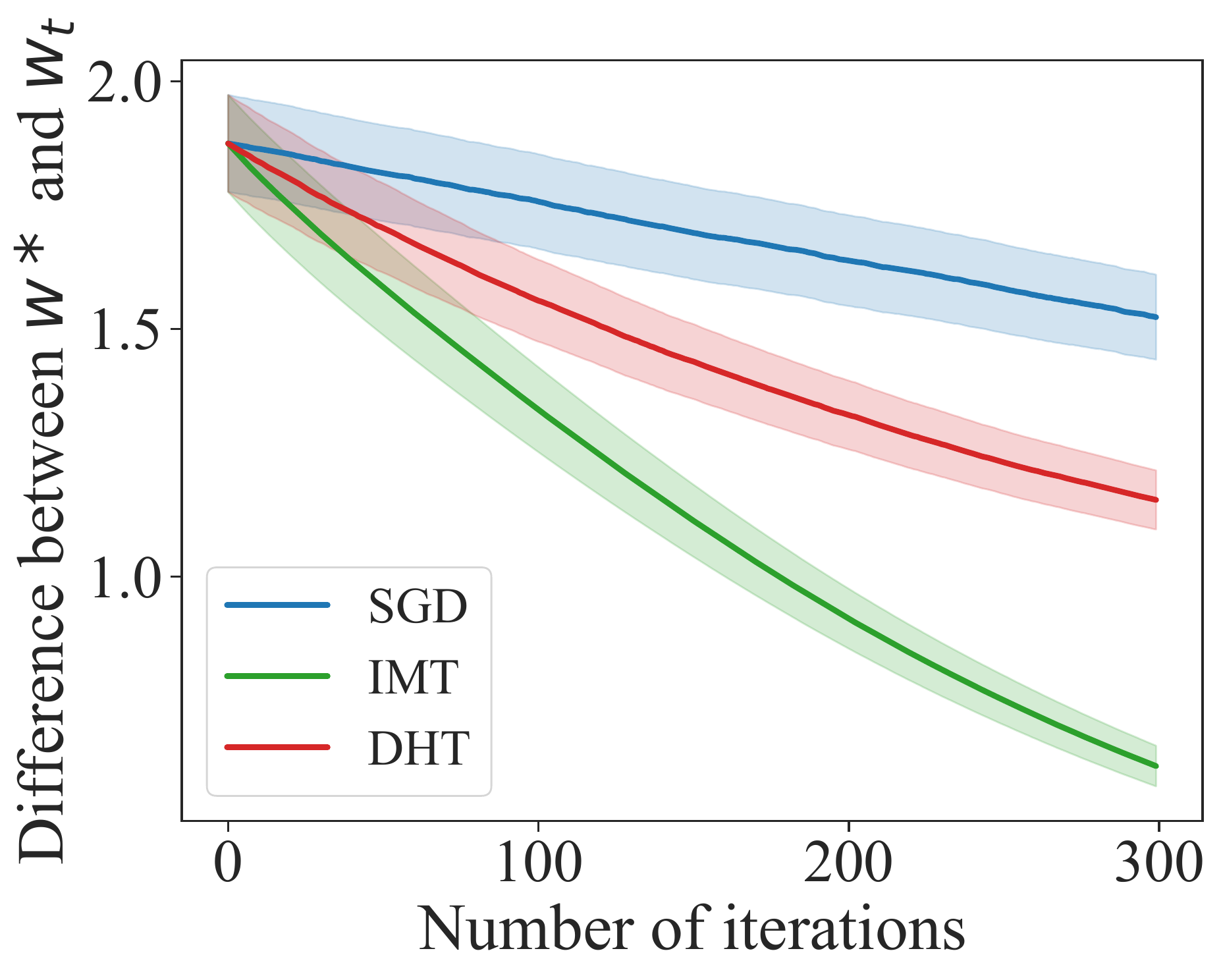}
    \end{subfigure}
    \caption{Accuracy and weight convergence using VAE-based teacher in 3/5 classification on MNIST.}
\end{figure}

We observe similar behavior between the weight convergence and the accuracy convergence of the examined methods using a GAN-based generative teacher on MNIST data. In general, the performance is comparable with that of the IMT baseline and significantly outperforms optimizing using random samples (SGD).

\textbf{Model input.} Model input for the teacher is the current student weight $\bm{w}^t$, the difference to the target model weight $\bm{w}^t - \bm{w}^*$, one random sample/label pair from the original dataset $(\bm{x}, \bm{y})$. The synthesized sample $\bm{\tilde{x}}$ is conditioned on label $\bm{y}$.

\textbf{Teacher architecture.} The GAN-based teacher utilizes a conditional deep convolutional GAN (DCGAN) to directly generate MNIST images with the original size ($28 \times 28$). The synthesized images are then downscaled using a projection matrix to teach the logistic regression learner. The generator consists of three blocks: each block consists of 2D transposed convolution layers, 2D batch normalization, and ReLU activation. Block $x$ upscales the input feature ($\mathbb{R}^{N \times D}$) to a sample feature map ($\mathbb{R}^{N \times 128 \times 3 \times 3}$); block $y$ upscales the label embedding ($\mathbb{R}^{N \times n_{cls}}$) to a label feature map ($\mathbb{R}^{N \times 128 \times 3 \times 3}$). Both feature maps are concatenated and inserted into the third block $xy$ and upscaled to generate MNIST-like samples (($\mathbb{R}^{N \times 1 \times 28 \times 28}$)). The discriminator operates in a similar fashion: block $x$ downscales the input image ($\mathbb{R}^{N \times 1 \times 28 \times 28}$) to a sample feature map ($\mathbb{R}^{N \times 32 \times 14 \times 14}$); block $y$ downscales the one-hot label embedding ($\mathbb{R}^{N \times 10 \times 28 \times 28}$) to a label feature map ($\mathbb{R}^{N \times 32 \times 14 \times 14}$). Both feature maps are concatenated and inserted into the third block $xy$ to predict if the sample is real or fake.

\begin{figure}[h!]
    \centering
    \begin{subfigure}{0.245\linewidth}
        \centering
        \includegraphics[width=0.99\linewidth]{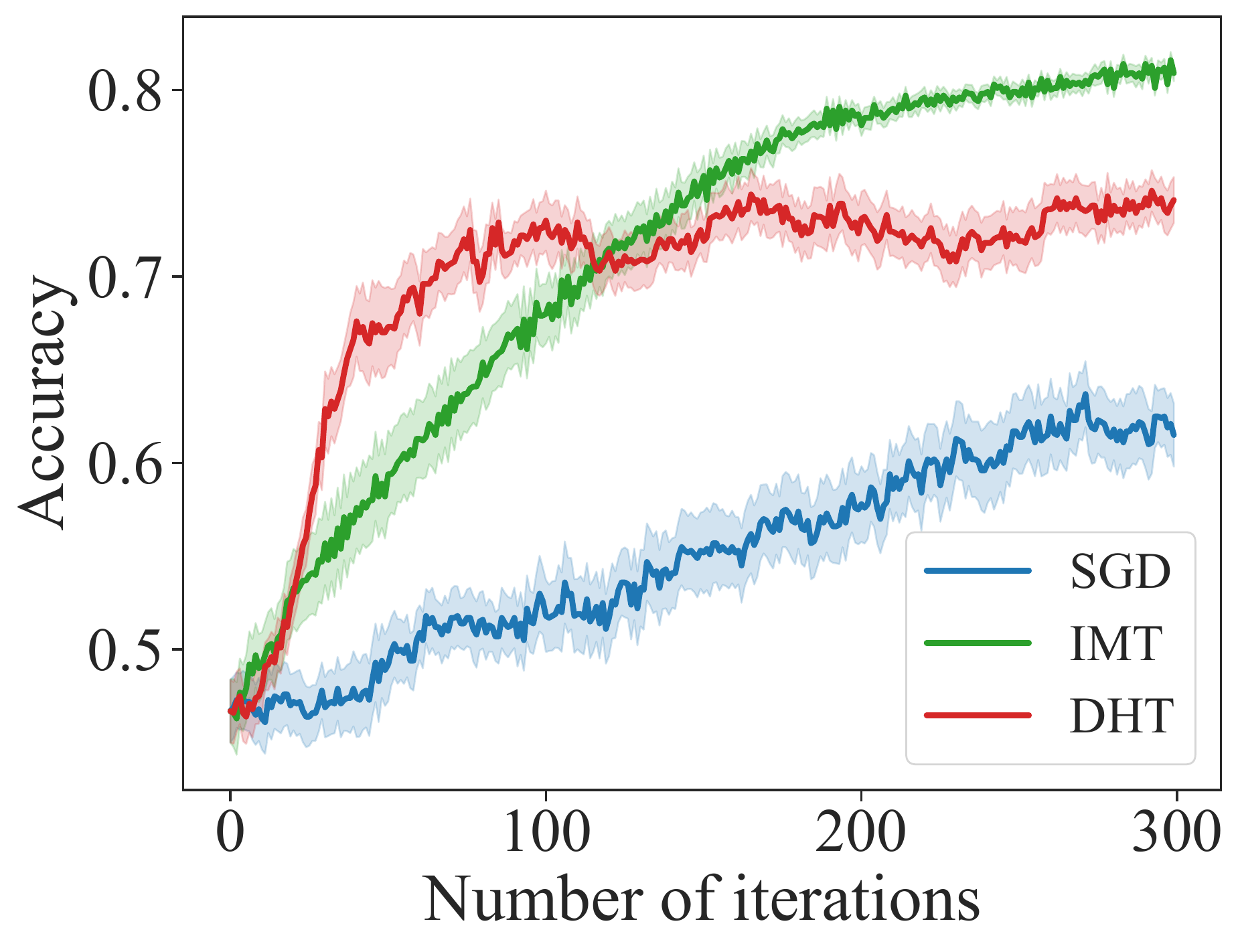}
    \end{subfigure}%
    \begin{subfigure}{0.245\linewidth}
        \centering
        \includegraphics[width=0.99\linewidth]{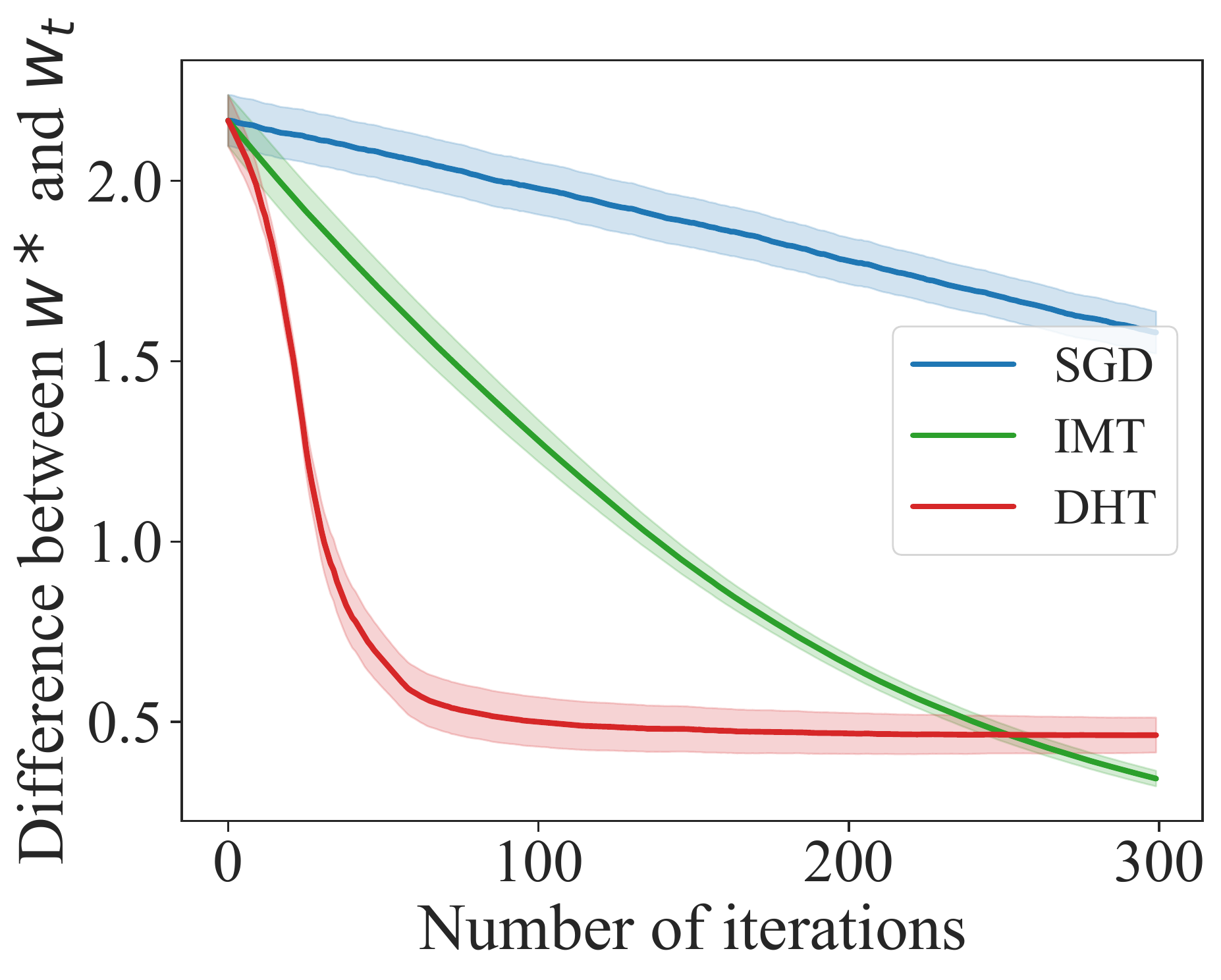}
    \end{subfigure}
    \caption{Accuracy and weight convergence using GAN-based teacher in 3/5 classification on MNIST.}
\end{figure}

\subsection{Teaching Logistic Regression on Half-moon Data with Parameterized Black-box Teaching}

We observe similar behavior between the weight convergence and the accuracy convergence of the examined methods using a parametrized black-box teaching policy on half-moon data. Even without the knowledge about $\bm{w}^*$, the DHT is able to significantly outperform the IMT baseline. It is worth mentioning that we use a surrogate $\bm{w}^*$ to calculate the \textbf{distance between $\bm{w}^t$ and $\bm{w}^*$}, as the $\bm{w}^*$ is not present. Here, we just use a general $\bm{w}^*$ with good classification performance.

\textbf{Model input.} Model input for the teacher is the current student weight $\bm{w}^t$ and one random sample/label pair from the original dataset $(\bm{x}, \bm{y})$. The synthesized sample $\bm{\tilde{x}}$ is conditioned on label $\bm{y}$.

\textbf{Teacher architecture.} The teacher is a simple MLP with three layers (input dimension (6) - 32 - 16 - output dimension (2)) and ReLU activation.

\begin{figure}[h!]
    \centering
    \begin{subfigure}{0.245\linewidth}
        \centering
        \includegraphics[width=0.99\linewidth]{imgs/paper_results_acc_blackbox_unrolled_moon.pdf}
    \end{subfigure}%
    \begin{subfigure}{0.245\linewidth}
        \centering
        \includegraphics[width=0.99\linewidth]{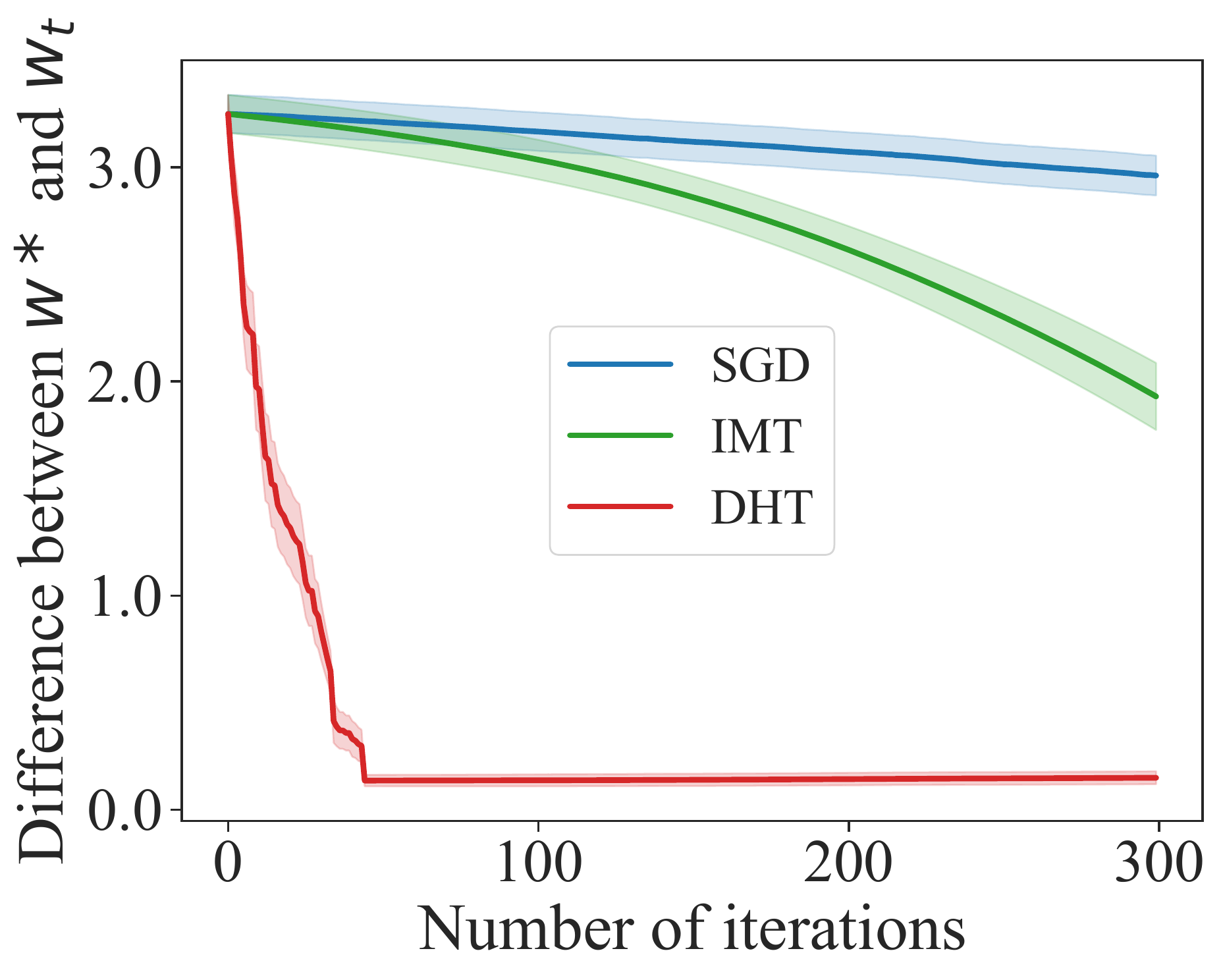}
    \end{subfigure}
    \caption{Accuracy and weight convergence using parametrized black-box teaching policy in binary classification on half-moon.}
\end{figure}

\subsection{Teaching Logistic Regression on MNIST with Parameterized Black-box Teaching}

We observe similar behavior between the weight convergence and the accuracy convergence of the examined methods using a parametrized black-box teaching policy on MNIST data. Even without the knowledge about $\bm{w}^*$, the DHT is able to significantly outperform the IMT baseline.

\textbf{Model input.} Model input for the teacher is the current student weight $\bm{w}^t$ and one random sample/label pair from the original dataset $(\bm{x}, \bm{y})$. The synthesized sample $\bm{\tilde{x}}$ is conditioned on label $\bm{y}$.

\textbf{Teacher architecture.} The teacher is an MLP with five layers (input dimension 58) - 128 - 256 - 512 - 512 - 1024 - output dimension (24)), ReLU activation and 1D batch normalization.

\begin{figure}[h!]
    \centering
    \begin{subfigure}{0.245\linewidth}
        \centering
        \includegraphics[width=0.99\linewidth]{imgs/paper_results_acc_blackbox_unrolled_mnist.pdf}
    \end{subfigure}%
    \begin{subfigure}{0.245\linewidth}
        \centering
        \includegraphics[width=0.99\linewidth]{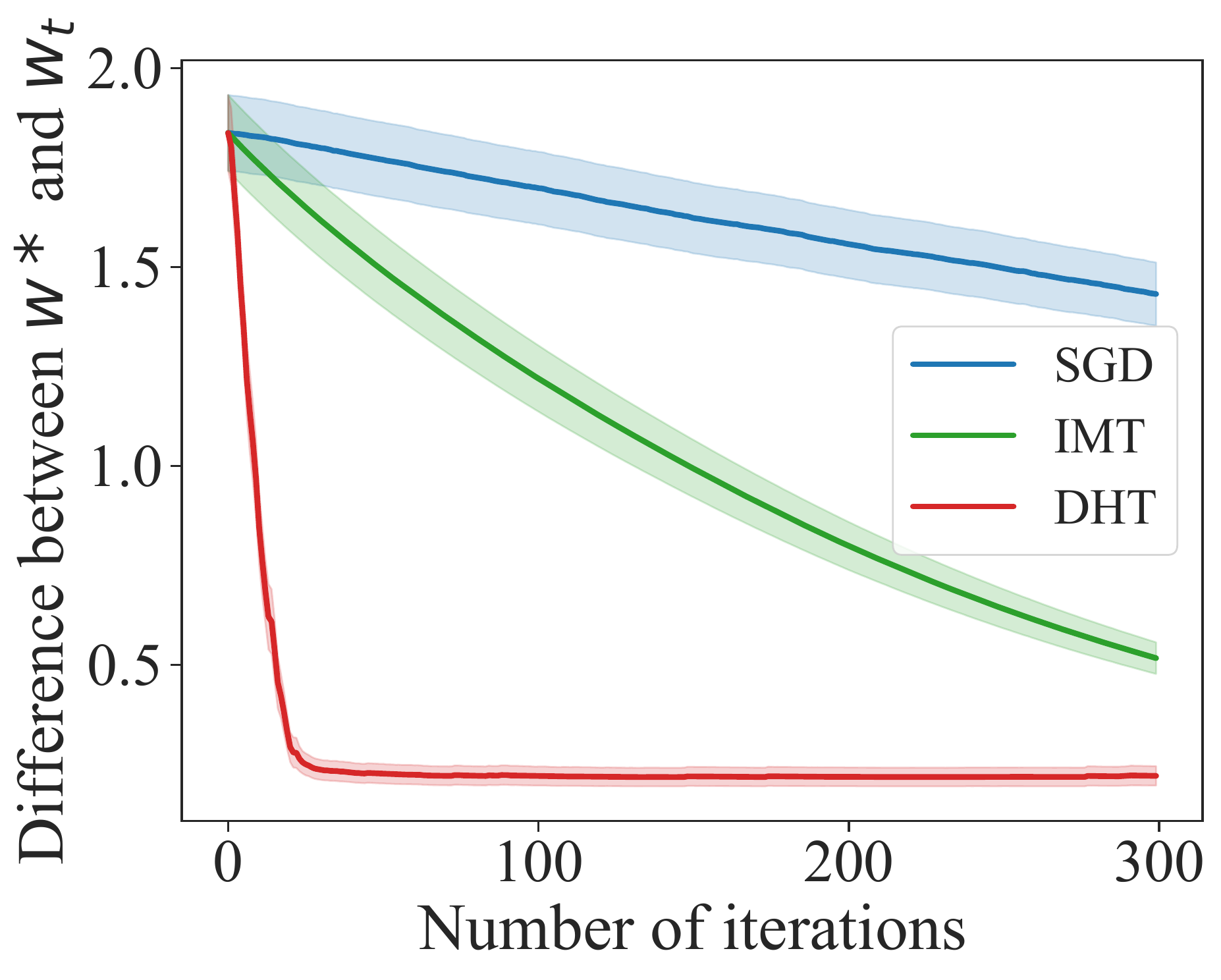}
    \end{subfigure}
    \caption{Accuracy and weight convergence using parametrized black-box teaching policy in 3/5 classification on MNIST.}
\end{figure}

\end{document}